\documentclass[sigconf, nonacm, pdfa]{acmart}

\newcommand\vldbdoi{10.14778/3665844.3665863}
\newcommand\vldbpages{2363 - 2377}
\newcommand\vldbvolume{17}
\newcommand\vldbissue{9}
\newcommand\vldbyear{2024}

\newcommand\vldbtitle{\shorttitle} 
\newcommand\vldbavailabilityurl{https://github.com/decisionintelligence/TFB}
\newcommand\vldbpagestyle{empty} 
\usepackage{makecell}
\usepackage{array}
\usepackage{algorithmicx,algorithm}
\usepackage{multirow}
\usepackage[normalem]{ulem}
\usepackage{subcaption}
\usepackage{graphicx}
\usepackage{tabularx}
\usepackage{diagbox}
\usepackage{balance}
\usepackage[flushleft]{threeparttable}
\usepackage{booktabs}
\usepackage{pifont}
\usepackage{amssymb}
\usepackage[misc]{ifsym} 
\usepackage{fontawesome}
\usepackage{color}
\usepackage[utf8]{inputenc}
\usepackage[T1]{fontenc}
\usepackage{CJKutf8}
\usepackage{enumitem}

\newtheorem{definition}{Definition}

\begin{document}

\title{TFB: Towards Comprehensive and Fair Benchmarking of Time Series Forecasting Methods}

\settopmatter{authorsperrow=4} 
 
\author{Xiangfei Qiu}
\affiliation{%
  \institution{East China Normal University, China}
}

\author{Jilin Hu}
\affiliation{%
  \institution{East China Normal University, China \Letter}
}


\author{Lekui Zhou}
\affiliation{%
  \institution{Huawei Cloud Algorithm Innovation Lab, China}
}

\author{Xingjian Wu}
\affiliation{%
  \institution{East China Normal University, China}
}

\author{Junyang Du}
\affiliation{%
  \institution{East China Normal University, China}
}

\author{Buang Zhang}
\affiliation{%
  \institution{East China Normal University, China}
}

\author{Chenjuan Guo}
\affiliation{%
  \institution{East China Normal University, China}
}

\author{Aoying Zhou}
\affiliation{%
  \institution{East China Normal University, China}
}

\author{Christian S. Jensen}
\affiliation{%
  \institution{Aalborg University, Denmark}
}

\author{Zhenli Sheng}
\affiliation{%
  \institution{Huawei Cloud Algorithm Innovation Lab, China}
}

\author{Bin Yang}
\affiliation{%
  \institution{East China Normal University, China}
}

\begin{abstract}
Time series are generated in diverse domains such as economic, traffic, health, and energy, where forecasting of future values has numerous important applications. Not surprisingly, many forecasting methods are being proposed. To ensure progress, it is essential to be able to study and compare such methods empirically in a comprehensive and reliable manner.  To achieve this, we propose \textbf{TFB}, an automated benchmark for Time Series Forecasting~(TSF) methods. TFB advances the state-of-the-art by addressing shortcomings related to datasets, comparison methods, and evaluation pipelines: 1)~insufficient coverage of data domains, 2)~stereotype bias against traditional methods, and 3)~inconsistent and inflexible pipelines.
To achieve better domain coverage, we include datasets from 10 different domains : traffic, electricity, energy, the environment, nature, economic, stock markets, banking, health, and the web. 
We also provide a time series characterization to ensure that the selected datasets are comprehensive. To remove biases against some methods, we include a diverse range of methods, including statistical learning, machine learning, and deep learning methods, and we also support a variety of evaluation strategies and metrics to ensure a more comprehensive evaluations of different methods. To support the integration of different methods into the benchmark and enable fair comparisons, TFB features a flexible and scalable pipeline that eliminates biases. 
Next, we employ TFB to perform a thorough evaluation of 21 Univariate Time Series Forecasting~(UTSF) methods on 8,068 univariate time series and 14 Multivariate Time Series Forecasting~(MTSF) methods on 25 datasets. The results offer a deeper understanding of the forecasting methods, allowing us to better select the ones that are most suitable for particular datasets and settings. Overall, TFB and this evaluation provide researchers with improved means of designing new TSF methods. 
\end{abstract}

\maketitle
\pagestyle{\vldbpagestyle}

\begingroup
\renewcommand\thefootnote{}\footnote{\noindent
This work is licensed under the Creative Commons BY-NC-ND 4.0 International License. Visit \url{https://creativecommons.org/licenses/by-nc-nd/4.0/} to view a copy of this license. For any use beyond those covered by this license, obtain permission by emailing \href{mailto:info@vldb.org}{info@vldb.org}. Copyright is held by the owner/author(s). Publication rights licensed to the VLDB Endowment. \\
\raggedright Proceedings of the VLDB Endowment, Vol. \vldbvolume, No. \vldbissue\ %
ISSN 2150-8097. \\
\href{https://doi.org/\vldbdoi}{doi:\vldbdoi} \\
}\addtocounter{footnote}{-1}\endgroup

\begingroup\small\noindent\raggedright\textbf{PVLDB Reference Format:}\\
Xiangfei Qiu, Jilin Hu, Lekui Zhou, Xingjian Wu, Junyang Du, Buang Zhang, Chenjuan Guo, Aoying Zhou, Christian S. Jensen, Zhenli Sheng and Bin Yang. \vldbtitle. PVLDB, \vldbvolume(\vldbissue): \vldbpages, \vldbyear.\\
\href{https://doi.org/\vldbdoi}{doi:\vldbdoi}
\endgroup

\ifdefempty{\vldbavailabilityurl}{}{
\vspace{.3cm}
\begingroup\small\noindent\raggedright\textbf{PVLDB Artifact Availability:}\\
The source code, data, and/or other artifacts have been made available at \url{https://github.com/decisionintelligence/TFB}.
\endgroup
}

\section{Introduction}
\label{sec:intro}
As part of the ongoing digitalization, time series are generated in a variety of domains, such as economic~\cite{sezer2020financial,huang2022dgraph}, traffic~\cite{ xu2023tme,lin2023origin, hu2018risk,mo2022ths, lin2024genstl, sun2023graph,lin2021pre,wan2022mttpre,yang2023lightpath,hu2019stochastic,DBLP:journals/vldb/GuoYHJC20,DBLP:journals/geoinformatica/HuYJM17}, health~\cite{ wei2022cancer, miao2021generative, lee2017big,tran2019representation}, energy~\cite{alvarez2010energy, guo2015ecomark}, and AIOps~\cite{ davidpvldb,wang2023real,qi2023high,campos2024qcore,DBLP:conf/icde/KieuYGJZHZ22,zhao2022outlier}. 
Time Series Forecasting~(TSF) is essential in key applications in these domains~\cite{pan2023magicscaler,DBLP:journals/sigmod/GuoJ014,yu2023cgf,DBLP:conf/ijcai/YangGHT021}. Given historical observations, it is valuable if we can know the future values ahead of time.
Correspondingly, TSF has been firmly established as an active research field, witnessing the proposal of numerous methods.

\begin{figure*}[t]
    \centering
    \includegraphics[width=0.95\linewidth]{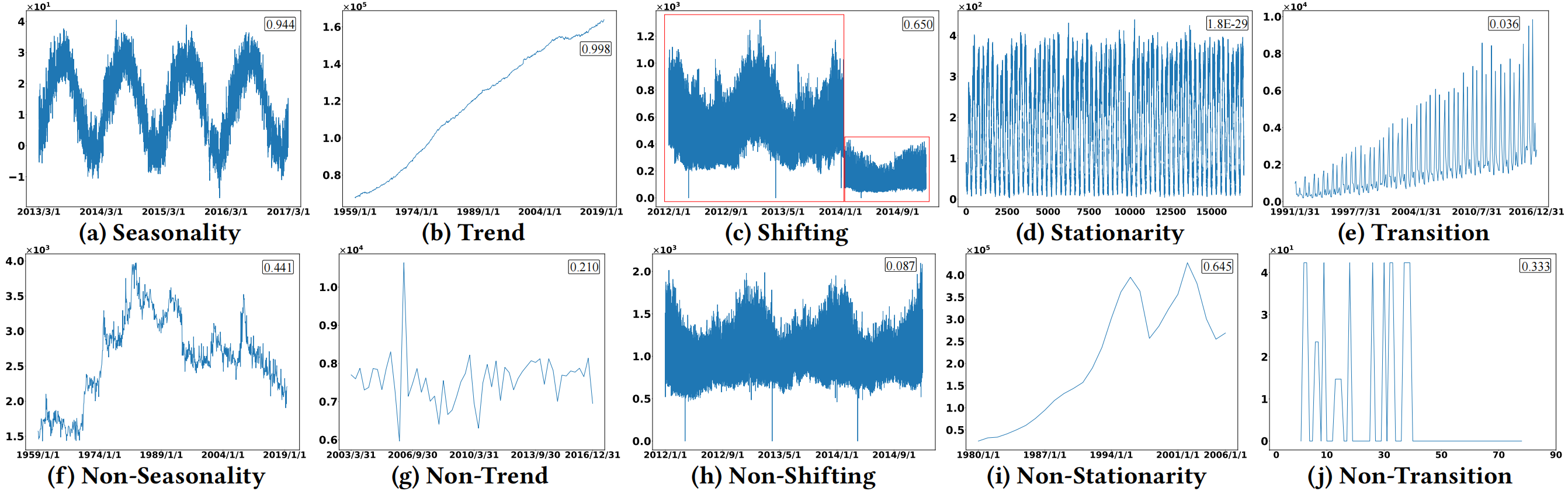}
    \caption{Visualization of data with different characteristics.}
\label{fig:pattern_eg}
\end{figure*}

Time series organize data points chronologically and are either univariate or multivariate depending on the number of variables in each data point. Accordingly, TSF methods can be classified as either Univariate Time Series Forecasting~(UTSF) or Multivariate Time Series Forecasting~(MTSF) methods. 
Among early methods, Autoregressive Integrated Moving Average~(ARIMA)~\cite{box1970distribution} and Vector Autoregression~(VAR)~\cite{toda1994vector} are arguably the most popular univariate and multivariate forecasting methods, respectively. Subsequent methods that exploit machine learning, e.g., XGBoost~\cite{chen2016xgboost,zhang2021time} and Random Forest~\cite{breiman2001random, mei2014random} offer better performance than the early methods. Most recently, methods based on deep learning have demonstrated state-of-the-art~(SOTA) forecasting performance on a variety of datasets~\cite{zhou2021informer, nie2022time, wu2022timesnet, zhang2022crossformer, chen2024pathformer, yao2023simplets, wu2021autocts, wu2023autocts+,zhao2023multiple,DBLP:journals/pvldb/PedersenYJ20,cheng2023weakly,DBLP:conf/icde/CirsteaYGKP22,MileTS,razvanicde2021,lin2023pre,haoicde24}.

As more and more methods are being proposed for different datasets and settings, there is an increasing need for fair and comprehensive empirical evaluations. To achieve this, we identify and address three issues in existing evaluation frameworks, thereby advancing our evaluation capabilities.

\begin{figure}[t]
    \centering
    \includegraphics[width=0.8\linewidth]{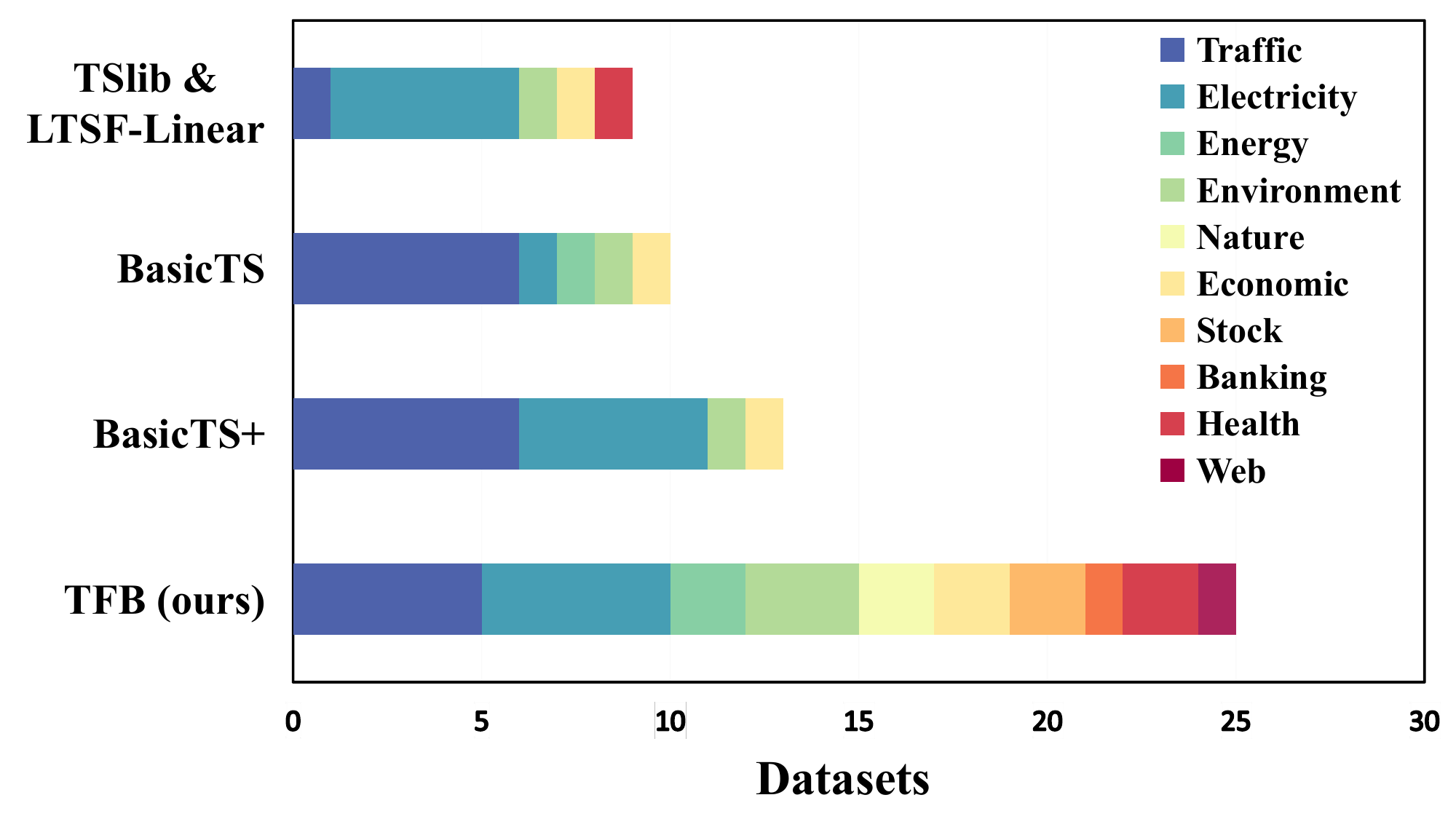}
    \caption{Statistics of data domains covered by existing multivariate time series benchmarks.}
    \label{fig:Stacked_bar_chart}
\end{figure}

\begin{figure}[t]
    \centering
    \includegraphics[width=0.75\linewidth]{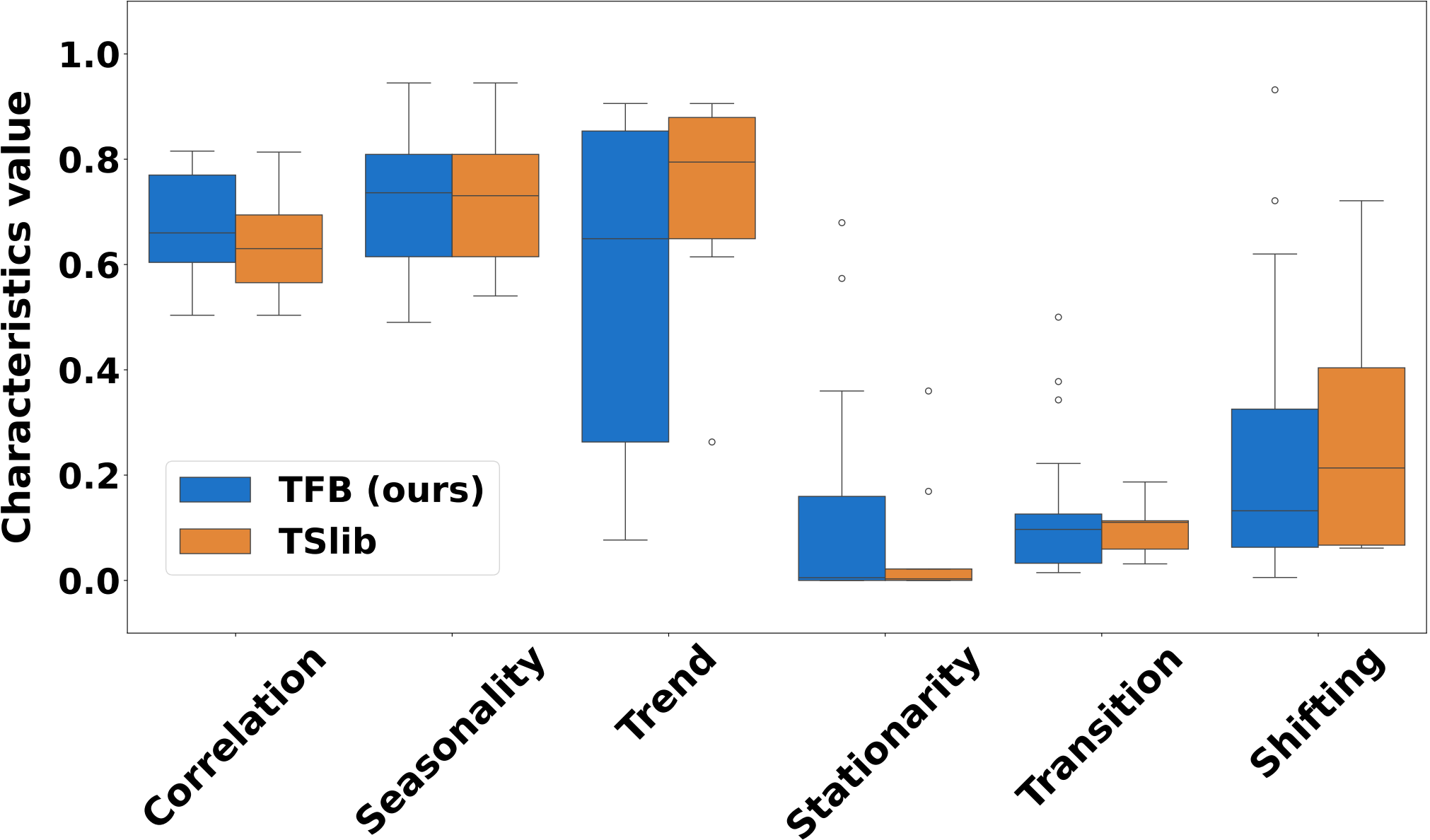}
    \caption{Box plot of the variations in normalized values of characteristics across the multivariate datasets in the TFB and TSlib.}
    \label{fig:multivariate datasets features1}
\end{figure}

\noindent
\textbf{Issue 1. Insufficient Coverage of Data Domains.} Time series from different domains may exhibit diverse characteristics. Figure~\ref{fig:pattern_eg}a depicts a time series from the environment domain called AQShunyi~\cite{zhang2017cautionary} that records temperature information at hourly intervals, exhibiting a distinct seasonal pattern. This pattern is reasonable in this scenario because temperatures in nature often cycle around the year. 
Figure~\ref{fig:pattern_eg}b shows a time series from FRED-MD~\cite{mccracken2016fred} belongs to economic domain that describes the monthly macroeconomic from 114 regional, national, and international sources with a clear increasing tendency. This may be attributed to overall economic stability with minimal fluctuations, reflecting sustained growth in the macroeconomic indicators.
Figure~\ref{fig:pattern_eg}c depicts a series among Electricity~\cite{misc_electricityloaddiagrams20112014_321} which comes from electricity domain and has a significant change in the data at a certain point in time, which might indicate an abrupt event, etc. 
However, these simple patterns are only the tip of the iceberg, and time series from different domains may exhibit much more complex patterns that either combine the above characteristics or are entirely different.
Therefore, using only limited domains results in limited coverage of time series characteristics, which cannot offer a full picture.

However, few empirical studies and benchmarks cover a wide variety of data domains. 
Figure~\ref{fig:Stacked_bar_chart} summarizes the multivariate data domains used in existing forecasting benchmarks which include MTSF. 
We observe that TSlib~\cite{wu2022timesnet}, LTSF-Linear~\cite{zeng2023transformers}, BasicTS~\cite{liang2022basicts}, and BasicTS+~\cite{shao2023exploring} only include around 10 datasets, covering less than or equal to 5 domains. We observe that these datasets are concentrated in mainly two domains, namely traffic and electricity. 
Since the multivariate time series datasets in TSlib are the most used, we investigate the variations in the values of the characteristics of datasets in TSlib and TFB---see Figure \ref{fig:multivariate datasets features1}. We observe that the TFB datasets exhibit more diverse distributions than those of TSlib across the six  characteristics. 
\textit{We argue that it is beneficial to broaden the coverage of domains, thereby enabling a more extensive assessment of method performance.}

\begin{table}[t]
\footnotesize
\caption{VAR, LR versus other methods, using MAE as the evaluation metric and a forecasting horizon of 24 steps.}
\begin{tabular}{l|m{0.7cm}m{0.5cm}m{0.9cm}m{0.7cm}m{0.9cm}p{1.1cm}}
\hline
\centering Datasets & \centering VAR & \centering LR & \centering PatchTST & \centering NLinear & \centering FEDformer & \centering\arraybackslash Crossformer \\ \hline
\centering NASDAQ & \centering\textbf{0.462} & \centering 0.616 & \centering 0.567 & \centering\underline{0.522} & \centering 0.547 & \centering\arraybackslash 0.745 \\ 
\centering Wind & \centering0.620 & \centering\textbf{0.583} & \centering 0.652  & \centering0.640 & \centering 0.697 & \centering\arraybackslash\underline{0.590} \\ 
\centering ILI & \centering 1.012 & \centering 4.856 & \centering\textbf{0.835}  & \centering\underline{0.919} & \centering 1.020 & \centering\arraybackslash 1.096 \\ \hline
\end{tabular}
\label{table:VAR}
\end{table}

\noindent
\textbf{Issue 2. Stereotype bias against traditional methods.} 
It is difficult for a single method to exhibit the best performance across all datasets. Methods exhibit varying performance across different datasets. To illustrate the issue, we conduct experiments on three datasets~(NASDAQ~\cite{feng2019temporal}, Wind~\cite{li2022generative}, and ILI~\cite{wu2021autoformer}) from different domains~(stock markets, energy, health) on methods VAR~\cite{toda1994vector}, PatchTST~\cite{nie2022time}, LinearRegression (LR)~\cite{kedem2005regression,herzen2022darts}, NLinear~\cite{zeng2023transformers}, FEDformer~\cite{zhou2022fedformer}, and Crossformer~\cite{zhang2022crossformer}. Results are shown in Table~\ref{table:VAR}. Surprisingly, VAR outperforms all recently proposed SOTA methods on NASQAD and is better than FEDfomer and Crossformer on ILI. Furthermore, LR performs better than recently proposed SOTA methods on Wind. However, the experimental studies in their original papers~\cite{zhou2022fedformer,zhang2022crossformer, nie2022time} do not include VAR and LR in their baselines and rather assume that traditional methods are incapable of obtaining competitive performance. 
From Table~\ref{Benchmarks Comparisons}, it follows that no existing MTSF benchmark has evaluated statistical methods. Moreover, as the training mechanisms for statistical methods differ from those of deep learning-based methods, it is difficult for existing benchmarks to accommodate statistical methods. \textit{We argue that it is beneficial to eliminate stereotype bias against traditional methods by comparing a wide range of methods.}

\begin{figure}[t]
\centering
\begin{minipage}{0.55\columnwidth}
    \centering
    \includegraphics[width=\linewidth]{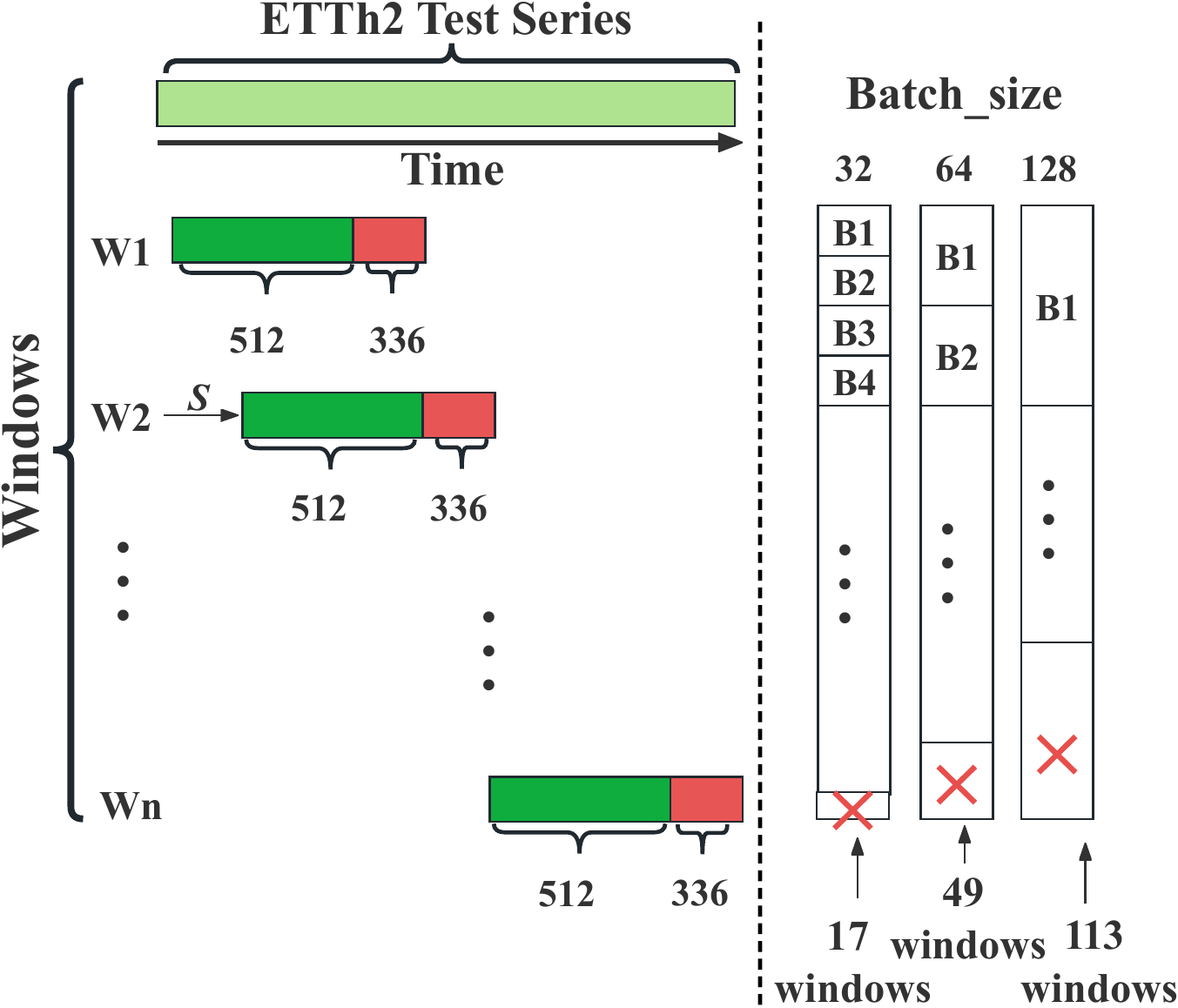}
    \caption{``Drop last'' illustration.}
    \label{fig:dataloader}
\end{minipage}\hspace{0.01\columnwidth}
\begin{minipage}{0.42\columnwidth}
\captionof{table}{Impact of batch sizes with ``drop last.''}
\setlength{\tabcolsep}{1.1pt}
    \centering
    \footnotesize
    \begin{tabularx}{\linewidth}{c|ccc}
        \hline
        Size      & PatchTST & DLinear & FEDformer \\ \hline
        1   & 0.4203   & 0.4874  & \textbf{0.4120}   \\ 
        32  & 0.4138   & 0.4831  & \textbf{0.4084}    \\ 
        64  & \textbf{0.3999}   & 0.4726  & 0.4022    \\ 
        128 & \textbf{0.3750}   & 0.4539  & 0.3921    \\ 
        256 & \textbf{0.3561}   & 0.4360  & 0.3825    \\ 
        512 & \textbf{0.3483}   & 0.4251  & 0.3736    \\ \hline
    \end{tabularx}
    
    \label{tab:batchsize}
\end{minipage}
\end{figure}

\noindent
\textbf{Issue 3. Lack of consistent and flexible pipelines.} The performance of different methods varies with changing experimental settings, e.g., splits among training/validation/testing data, the choice of normalization method, and the selection of hyper-parameter settings. For example, implementations of existing methods often employ a \textit{``Drop Last''} trick in the testing phase~\cite{zhou2022fedformer,nie2022time,wang2022micn}. To accelerate the testing, it is common to split the data into batches. However, if we discard the last incomplete batch with fewer instances than the batch size, this causes unfair comparisons. 
For example, in Figure~\ref{fig:dataloader}, the ETTh2~\cite{zhou2021informer} has a testing series of length 2,880, and we need to predict 336 future time steps using a look-back window of size 512. If we select the batch sizes to be 32, 64, and 128, the number of samples in the last batch are 17, 49, and 113, respectively. 
Discarding those last-batch testing samples is inappropriate unless all methods use the same strategy. Table~\ref{tab:batchsize} shows the testing results of PatchTST~\cite{nie2022time}, DLinear~\cite{zeng2023transformers} and FEDformer~\cite{zhou2022fedformer} on the ETTh2 with different batch sizes and the \textit{``Drop Last''} trick turned on. We observe that the performance of the methods changes when varying the batch size. 
In addition, the pipelines in most benchmarks are inflexible and do not support simultaneous evaluation of statistical learning, machine learning, and deep learning methods. 
\textit{We argue that it is crucial to ensure a consistent and flexible pipeline so that methods are evaluated in the same setting, thereby improving the fairness of the findings.}

Robust and extensive benchmarks enable researchers to evaluate proposals for new methods more rigorously across diverse range of datasets, which is crucial for advancing the state-of-the-art~\cite{tanmonash,qiao2024class}. 
For example, ImageNet~\cite{deng2009imagenet}, that encompasses a substantial dataset has been instrumental in ensuring progress in computer vision. ImageNet has established itself as a standard for assessing methods in image processing due to its support for rigorous evaluation. 
Table~\ref{Benchmarks Comparisons} compares existing benchmarks for TSF according to seven properties. No existing benchmark possesses all properties.

We propose the Time series Forecasting Benchmark~(TFB) to facilitate the empirical evaluation and comparison of TSF methods more comprehensively across application domains and methods, and with improved fairness. TFB contributes a collection of challenging and realistic datasets and provides a user-friendly, flexible, and scalable evaluation pipeline that offers robust evaluation support. TFB possesses the following key characteristics: 
\begin{itemize}[left=0.1cm]
\item A comprehensive collection of datasets organized according to a taxonomy~(to address \textbf{Issue 1}): Its collection of datasets offer diverse characteristics, encompassing time series from multiple domains and complex settings. This contributes to ensuring more robust and extensive evaluations.
\item Broad coverage of existing methods and extended support for evaluation strategies and metrics~(to address \textbf{Issue 2}): TFB covers a diverse range of methods, including statistical learning, machine learning, and deep learning methods, accompanied by a variety of evaluation strategies and metrics. This richness enables more comprehensive evaluations across methods and evaluation settings.
\item Flexible and scalable pipeline~(to address \textbf{Issue 3}): By its design, TFB improves the fairness of comparisons of methods. Methods are evaluated using a unified pipeline, consistent and standardized evaluation strategy and datasets are employed, eliminating biases and enabling more accurate performance comparisons. This enables more fair and meaningful conclusions regarding the effectiveness and efficiency of methods.
\end{itemize}

\begin{table*}[t!]
\small
  \caption{Time series forecasting benchmark comparison.}
  \label{Benchmarks Comparisons}
  \begin{tabular}{lccccccc}
    \toprule
     \diagbox{Benchmark}{Property} & \makecell{Univariate  \\ forecasting}  & \makecell{Multivariate  \\ forecasting}  & \makecell{Statistical  \\ method}  & \makecell{Machine learning  \\ method}  & \makecell{Deep learning  \\ method}  & \makecell{Taxonomy  \\ of data} & \makecell{Flexible \& scalable\\pipeline} \\
    \midrule
    M3~\cite{makridakis2000m3} & $\surd$ & \scalebox{1.25}[1]{$\times$} & $\surd$ & $\surd$  & \scalebox{1.25}[1]{$\times$} & \scalebox{1.25}[1]{$\times$} & \scalebox{1.25}[1]{$\times$}\\
    M4~\cite{makridakis2018m4} & $\surd$ & \scalebox{1.25}[1]{$\times$} & $\surd$ & $\surd$  & $\surd$ & \scalebox{1.25}[1]{$\times$} & \scalebox{1.25}[1]{$\times$}\\
    LTSF-Linear~\cite{zeng2023transformers} & \scalebox{1.25}[1]{$\times$} & $\surd$ & \scalebox{1.25}[1]{$\times$} & \scalebox{1.25}[1]{$\times$} & $\surd$ & \scalebox{1.25}[1]{$\times$} & $\bigcirc$\\
    TSlib~\cite{wu2022timesnet} & $\surd$ & $\surd$ & \scalebox{1.25}[1]{$\times$} & \scalebox{1.25}[1]{$\times$} & $\surd$ & \scalebox{1.25}[1]{$\times$} & $\bigcirc$\\
    BasicTS~\cite{liang2022basicts} & \scalebox{1.25}[1]{$\times$} & $\surd$ & \scalebox{1.25}[1]{$\times$}  & $\surd$ & $\surd$ & \scalebox{1.25}[1]{$\times$} & $\bigcirc$\\
    BasicTS+~\cite{shao2023exploring} & \scalebox{1.25}[1]{$\times$} & $\surd$ & \scalebox{1.25}[1]{$\times$}  & \scalebox{1.25}[1]{$\times$} & $\surd$ & $\bigcirc$ & $\bigcirc$\\
    Monash~\cite{godahewa2021monash} & $\surd$ & \scalebox{1.25}[1]{$\times$} & $\surd$ & $\surd$ & \scalebox{1.25}[1]{$\times$} & \scalebox{1.25}[1]{$\times$} & $\bigcirc$\\
    Libra~\cite{bauer2021libra} & $\surd$ & \scalebox{1.25}[1]{$\times$} & $\surd$  & $\surd$ & \scalebox{1.25}[1]{$\times$} & \scalebox{1.25}[1]{$\times$} & $\bigcirc$\\
    TFB (Ours) & $\surd$ & $\surd$ & $\surd$ & $\surd$ & $\surd$ & $\surd$ & $\surd$\\
    \bottomrule
\multicolumn{8}{l}{\scalebox{1.25}[1]{$\times$} indicates absent, $\surd$ indicates present, $\bigcirc$ indicates incomplete.}
  \end{tabular}
\end{table*}

Based on the experiments conducted using TFB, we make the following key observations:~(1)~ The statistical methods VAR~\cite{toda1994vector} and LinearRegression (LR)~\cite{kedem2005regression,herzen2022darts} perform better than recently proposed SOTA methods on some datasets---see Table~\ref{New Multivariate forecasting results}. (2)~Linear-based methods perform well when datasets exhibit an increasing trend or significant shifts. (3)~Transformer-based methods outperform linear-based methods on datasets with marked seasonality, and nonlinear patterns, as well as more pronounced patterns or strong internal similarities. (4)~Methods that consider dependencies between channels can enhance the performance of MTSF substantially, particularly on datasets with strong correlations, compared to methods that assume channel independence.

We make the following main contributions.
\begin{itemize}[left=0.1cm]
\item We present TFB which is specifically designed to further improve fair comparisons in TSF, include UTSF and MTSF. TFB facilitates comparisons with 20+ UTSF methods on 8,068 univariate time series and 14 MTSF methods on 25 multivariate datasets. 
\item We identify, collect, and process previously proposed TSF datasets to determine a comprehensive dataset covering diverse domains and characteristics and organize them in a standardized format. 
Then, we design experiments to study the performance of different methods over different characteristics of datasets. 
\item TFB offers an automated end-to-end pipeline for evaluating forecasting methods, streamlining and standardizing the steps of loading time series datasets, configuring experiments, and evaluating methods. This simplifies the evaluation process for researchers. Additionally, all datasets and code are available at \href{https://github.com/decisionintelligence/TFB}{https://github.com/decisionintelligence/TFB}.
\item We use TFB to evaluate and compare a diverse set of methods, covering statistical learning, machine learning, and deep learning methods and a rich array of evaluation tasks and strategies. We summarize the evaluation results into some key findings. 
\item We have also launched an online time series leaderboard: \url{https://decisionintelligence.github.io/OpenTS/OpenTS-Bench/}.
\end{itemize}

The rest of the paper is structured as follows. We review related work in Section~\ref{sec:RELATED WORK} and introduce terms and definitions in Section~\ref{sec:Preliminaries}. In Section~\ref{sec:TFB}, we cover the design of TFB, and in Section~\ref{sec:EXPERIMENTS}, we use TFB to benchmark existing forecasting methods. We offer conclusions in Section ~\ref{Conclusions}.

\section{RELATED WORK}
\label{sec:RELATED WORK}

We proceed to first cover approaches to TSF and then review existing benchmarking proposals.

\noindent
\textbf{Time series forecasting:}~Existing methods for TSF can be categorized broadly into three main categories: statistical learning, machine learning, and deep learning methods.

Early proposals primarily employed statistical learning methods. ARIMA~\cite{box1970distribution}, ETS~\cite{hyndman2008forecasting}, Theta~\cite{garza2022statsforecast}, VAR~\cite{toda1994vector}, and Kalman Filter~(KF)~\cite{harvey1990forecasting} are classical and widely utilized methods. These forecasting methods are based on the idea that future values of time series can be predicted from observed, past values. With the rapid advances in machine learning technologies, machine learning methods for TSF have emerged~\cite{fischer2020machine}. Notably, XGBoost~\cite{chen2016xgboost,zhang2021time}, Gradient Boosting Regression Trees (GBRT)~\cite{friedman2001greedy}, Random Forests\cite{breiman2001random, mei2014random} and LightGBM~\cite{ke2017lightgbm} have been applied extensively to better accommodate nonlinear relationships and complex patterns. 
Machine learning methods are flexibile at handling different types and lengths of time series and generally offer better forecasting accuracy than traditional methods.

However, these methods still require manual feature engineering and model design. Taking advantage of the representation learning capabilities offered by deep neural networks (DNNs) on rich data, numerous deep learning methods have been proposed. In many cases, these methods outperform traditional techniques in terms of forecasting accuracy. TCN~\cite{bai2018empirical} and DeepAR~\cite{salinas2020deepar} treat time series as sequences of vectors and utilize CNNs or RNNs to capture temporal dependencies. Additionally, Transformer architectures, including Informer~\cite{zhou2021informer}, FEDformer~\cite{zhou2022fedformer}, Autoformer~\cite{wu2021autoformer}, Triformer~\cite{cirstea2022triformer}, and PatchTST~\cite{nie2022time} can capture more complex temporal dynamics, significantly improving forecasting performance. MLP-based models such as N-HiTS~\cite{challu2023nhits}, N-BEATS~\cite{oreshkin2019n}, NLinear~\cite{zeng2023transformers}, and DLinear~\cite{zeng2023transformers} employ a simple architecture with relatively few parameters and have also demonstrated good forecasting accuracy.

\noindent
\textbf{Benchmarks:}~Several benchmarks for TSF have been proposed: Libra~\cite{bauer2021libra}, BasicTS~\cite{liang2022basicts}, BasicTS+~\cite{shao2023exploring}, Monash~\cite{godahewa2021monash}, M3~\cite{makridakis2000m3}, M4~\cite{makridakis2018m4}, LTSF-Linear~\cite{zeng2023transformers} and TSlib~\cite{wu2022timesnet}. However, these benchmarks fall short in different aspects---see Table~\ref{Benchmarks Comparisons}.

First, most benchmarks consider either univariate or multivariate time series forecasting. Early works, i.e., M3~\cite{makridakis2000m3}, M4~\cite{makridakis2018m4}, Libra~\cite{bauer2021libra} and Monash~\cite{godahewa2021monash}, only pay attention to univariate time series. And recent works, e.g., LTSF-Linear~\cite{zeng2023transformers}, BasicTS~\cite{liang2022basicts} and BasicTS+~\cite{shao2023exploring}, just consider multivariate. Only TSlib~\cite{wu2022timesnet} considers both. 

Second, a notable issue arises regarding the assessment of method diversity. In the early times, the deep learning methods are not dominant in TSF, so most early benchmarks do not have deep learning methods in their comparison, e.g., M3~\cite{makridakis2000m3}, M4~\cite{makridakis2018m4} and Libra~\cite{bauer2021libra}. On the contrary, with more deep learning methods demonstrating their performance in TSF, recent benchmarks focus solely on deep learning methods, e.g., LTSF-Linear~\cite{zeng2023transformers}, TSlib~\cite{wu2022timesnet}, BasicTS~\cite{liang2022basicts} and BasicTS+~\cite{shao2023exploring}. As mentioned in Section~\ref{sec:intro}, this limited coverage of different method paradigms is a concern.

Moreover, there is a challenge regarding scalable and unified pipelines. In the earliest works, M3~\cite{makridakis2000m3} and M4~\cite{makridakis2018m4} do not provide a pipeline in their benchmark at all, making it hard to integrate new methods. Monash~\cite{godahewa2021monash} and Libra~\cite{bauer2021libra} have pipelines but it is designed for statistical methods, which are unable to integrate deep learning methods. On the contrary, LTSF-Linear~\cite{zeng2023transformers} and TSlib~\cite{wu2022timesnet} have pipelines specific to deep learning methods, which are unable to evaluate statistical methods. We consider these pipelines to be incomplete, reducing scalability and hindering their broader application of these benchmarks.

Furthermore, the majority of benchmarks do not categorize the dataset; only BasicTS+~\cite{shao2023exploring} provides a coarse-grained classification. We consider this taxonomy to be incomplete. In contrast, TFB involves a fine-grained classification of the dataset.

Generally, TFB aims to enable a more reliable, thorough, and user-friendly evaluation. Thus, the proposed benchmark encompasses a broader range of methods and utilizes more scalable implementation pipelines than existing benchmarks, ensuring a fair comparison of TSF methods and promoting progress in the field of TSF.

\section{Preliminaries}
\label{sec:Preliminaries}
We provide definitions of time series and time series forecasting, and we cover key dataset characteristics, including trend, seasonality, stationarity, shifting, transition, and correlation.

\begin{definition}[Time series]
A time series  $X \in \mathbb{R}^{T\times N}$ is a time-oriented sequence of N-dimensional time points,  where $T$ is the number of time points, and $N$ is the number of variables. When $N=1$, a time series is called univariate. When $N>1$, it is called multivariate.     
\end{definition}

\begin{definition}[Time series forecasting]
Given a historical time series $X\in\mathbb{R}^{H\times N}$ of $H$ time points, time series forecasting aims to predict the next $F$ future time points, i.e., $Y\in\mathbb{R}^{F\times N}$, where $F$ is called the forecasting horizon.   
\end{definition}

\begin{definition}[Trend]
The trend of a time series refers to the long-term changes or patterns that occur over time. Intuitively, it represents the general direction in which the data is moving. Referring to the explained variance~\cite{o1982measures}, Trend Strength can be defined as follows.
\begin{equation}
\footnotesize
\mathit{Trend\_Strength} = \mathit{max} \left(0, 1 - \frac{\mathit{var}\left(R\right)}{\mathit{var}\left(X - S\right)}\right),
\end{equation}
where $X$ is a time series that can be decomposed into a trend~$(T)$, seasonality~$(S)$, and the remainder~$(R)$: $X=S+T+R$ by employing Seasonal and Trend decomposition using Loess, which is a highly versatile and robust method for time series decomposition~\cite{cleveland1990stl}.
\end{definition}

\begin{algorithm}[t]
\caption{Calculating Shifting Values of Time Series}
\footnotesize
\begin{flushleft}
{\bf Input:} 
Time series $X\in \mathbb{R}^{T\times 1}$

{\bf Output:}
Shifting value $\delta$ ~$\in(0,1)$ of $X$
\end{flushleft}
\begin{algorithmic}[1]
\State Normalize $X$ by calculating the z-score to obtain $Z\in \mathbb{R}^{T\times 1}$
\State $Z_{\min} \leftarrow \min(Z),~Z_{\max} \leftarrow \max(Z)$
\State $S = \{s_{i}~|~s_{i} \leftarrow Z_{\min}+(i-1)\frac{Z_{\max}-Z_{\min}}{m},1\leq i\leq m\}$ where m is the number of thresholds
\State {\bf for} $s_i$ in $S$ {\bf do}
\State \hspace{0.1in} $K \leftarrow \{j~|~ Z_j>s_{i},1\leq j\leq T\}$,~$M_i$ $\leftarrow$ $\mathit{median}(K)$,~ $1 \leq i \leq m$
\State {\bf end for}
\State  $M ^\prime \leftarrow$ $\mathit{Min}$--$\mathit{Max}$ $\mathit{Normalization}(M)$
\State {\bf return} $\text{$\delta$} \leftarrow \mathit{median}(\{M_1 ^\prime, M_2^\prime, ..., M_m^\prime\})$
\end{algorithmic}
\label{alg1}
\end{algorithm}

\begin{algorithm}[t]
\caption{Calculating Transition Values of Time Series}
\footnotesize
\begin{flushleft}
{\bf Input:} 
Time series $X\in \mathbb{R}^{T\times 1}$

{\bf Output:}
Transition value $\Delta$ ~$\in(0, \frac{1}{3})$ of $X$
\end{flushleft}
\begin{algorithmic}[1]
\State Calculate the first zero crossing of the autocorrelation function: 
\Statex $\tau \leftarrow \mathit{firstzero\_ac}(X)$  
\State Generate $Y\in \mathbb{R}^{T^\prime\times 1}$ by downsampling X with stride $\tau$
\State Define index  $I = \mathit{argsort}(Y) \in \mathbb{R}^{T^\prime\times 1}$, then characterize  $Y$ to obtain $Z\in \mathbb{R}^{T^\prime\times 1}$:
\State {\bf for} $j \in [0: T^\prime]$ {\bf do}
\State \hspace{0.1in} $Z[j] \leftarrow \mathit{floor}(~I[j] / \frac{1}{3}T^\prime)$
\State {\bf end for}
\State Generate a transition matrix $M\in \mathbb{R}^{3\times 3}$:
\State {\bf for} $j \in [0: T^\prime] $ {\bf do}
\State \hspace{0.1in} $M[Z[j]-1][Z[j+1]-1]$++
\State {\bf end for}
\State $M^\prime \leftarrow \frac{1}{T^\prime}M$
\State Compute the covariance matrix $C$ between the columns of $M^\prime$
\State {\bf return} $\Delta \leftarrow \mathit{tr}(C)$
\end{algorithmic}
\label{alg2}
\end{algorithm}

\begin{definition}[Seasonality]
Seasonality refers to the phenomenon where changes in a time series repeat at specific intervals. Similar to the measurement of trend, the strength of Seasonality of a time series $X$ can be estimated as follows.
\begin{equation}
\footnotesize
\mathit{Seasonality\_Strength}=\max \left(0, 1-\frac{\mathit{var}\left(R\right)}{\mathit{var}\left(X-T\right)}\right)
\end{equation}
\end{definition}

\begin{definition}[Stationarity]
Stationarity refers to the mean of any observation in a time series $X=\langle x_1,x_2,...,x_n\rangle$ is constant, and the variance is finite for all observations. Also, the covariance $\mathit{cov}(x_i,x_j)$ between any two observations $x_i$ and $x_j$ depends only on their distance $|j-i|$, i.e., $\forall i+r\leq n,j+r\leq n$ $(\mathit{cov}(x_i,x_j)=\mathit{cov}(x_{i+r},x_{j+r})).$ Strictly stationary time series are rare in practice. Therefore, weak stationarity conditions are commonly applied~\cite{lee2017anomaly} ~\cite{nason2006stationary}. In our paper, we also exclusively focus on weak stationarity.

We adopt the Augmented Dick-Fuller (ADF) test statistic~\cite{elliott1992efficient} to quntify stationarity. The stationarity can be calculated as follows.
\begin{equation}
\footnotesize
\mathit{Stationarity}=\begin{cases}\mathit{True} , &\mathit{ADF}(X)\leq0.05\\\mathit{False} ,&\mathit{ADF}(X)>0.05\end{cases}
\end{equation}
\end{definition}

\begin{definition}[Shifting]
Shifting refers to the phenomenon where the probability distribution of time series changes over time. This behavior can stem from structural changes within the system, external influences, or the occurrence of random events. As the value approaches 1, the degree of shifting becomes more severe. Algorithm~\ref{alg1} details the calculation process.

\end{definition}

\begin{definition}[Transition]
Transition refers to the trace of the covariance of transition
matrix between symbols in a 3-letter alphabet~\cite{lubba2019catch22}. It captures the regular and identifiable fixed features present in a time series, such as the clear manifestation of trends, periodicity, or the simultaneous presence of both seasonality and trend. Algorithm~\ref{alg2} details the calculation process.
\end{definition}

\begin{definition}[Correlation]
Correlation refers to the possibility that different variables in a multivariate time series may share common trends or patterns, indicating that they are influenced by similar factors or have some underlying relationship.
Correlation is calculated as follows.

First, use Catch22~\cite{lubba2019catch22} to extract 22 features for each variable in $X$, serving as a representation for each variable. $F=\langle F^1,F^2,...,F^N\rangle\in \mathbb{R}^{22\times N}$ is obtained using the formulation specified in Equation ~\ref{F}.

\begin{equation}
\label{F}
\footnotesize
F=\mathit{Catch22}(X)
\end{equation}
Second, according to Equation ~\ref{R} we calculate the Pearson correlation coefficient between all variable pairs. 
\begin{equation}
\label{R}
\footnotesize
P=\left\{r(F^{i},F^{j})\mid 1\leq i\leq N,i+1\leq j\leq N,i,j\in N^{\ast}\right\}
\end{equation}
Third, compute the mean and variance of all Pearson correlation coefficients~(PCCs)~\cite{cohen2009pearson} to obtain the correlation, as shown in Equation~\ref{Similarity}.
\begin{equation}
\label{Similarity}
\footnotesize
\mathit{Correlation}=\mathit{mean}\left(P\right)+\frac{1}{1+\mathit{var}\left(P\right)}
\end{equation}
\end{definition}

Figure ~\ref{fig:pattern_eg} shows time series exhibit diverse characteristics in real-world scenarios. For each characteristic, the calculated characteristic values in the upper right corner of the image corresponding to series with distinct and indistinct characteristic are significantly different.
In summary, the time series characteristics define here enable a deep understanding of time series. Using these characteristics for classifying time series, we can more selectively choose forecasting methods that are appropriate for specific scenarios.

\section{TFB: BENCHMARK DETAILS}
\label{sec:TFB}
We proceed to cover the design of the TFB benchmark. First, we provide a detailed overview of the dataset in TFB, including the process of collection, accompanied by a comprehensive demonstration.~(Section~\ref{sec:Datasets}). Second, we categorize the existing methods supported~(Section~\ref{sec:Baselines}). Third, we cover evaluation strategies and metrics~(Section~\ref{sec:Evaluation Settings}). Finally, we describe the full benchmark pipeline (Section~\ref{sec:Unified Pipeline}).

\begin{table}[t]
\caption{Statistics of univariate datasets.}
\footnotesize
\label{Univariate datasets}
\setlength{\tabcolsep}{1.5pt}
\begin{threeparttable}
\begin{tabular}{l|cccccccc}
\hline
Frequency &
  \textit{\#Series} &
  \textit{\begin{tabular}[c]{@{}c@{}}Seasonality\end{tabular}} &
  \textit{\begin{tabular}[c]{@{}c@{}}Trend\end{tabular}} &
  \textit{\begin{tabular}[c]{@{}c@{}}Shifting\end{tabular}} &
  \textit{\begin{tabular}[c]{@{}c@{}}Transition\end{tabular}} &
  \textit{\begin{tabular}[c]{@{}c@{}}Stationarity\end{tabular}} &
  \begin{tabular}[c]{@{}c@{}}$|TS|\textless300^1$\end{tabular} &
  \textit{\begin{tabular}[c]{@{}c@{}}$F^2$\end{tabular}}  
   \\ \hline
Yearly     & 1,500 & 611 & 1,086 & 978 & 633 & 354 & 1,499& 6 \\ 
Quarterly  & 1,514 & 486 & 933 & 889 & 894 & 471  & 1,508& 8\\
Monthly    & 1,674 & 883 & 884 & 778 & 1,212 & 667 & 1,026& 18\\
Weekly     & 805 & 253 & 330 & 445 & 407 & 372  & 473& 13\\
Daily      & 1,484 & 374 & 502 & 487 & 1,176 & 714  & 442 & 14\\
Hourly     & 706 & 435 & 276 & 284 & 680 & 472  & 0& 48\\
Other      & 385 & 75 & 248 & 236 & 195 & 124 & 215& 8\\\hline
Total      & 8,068 & 3,117 & 4,259 & 4,097 & 5,197 & 3,174 & 5,163\\\hline
\end{tabular}
    \begin{tablenotes}
    \tiny
      \item[1] $|TS|\textless300$: the length of univariate time series \textless300
      \item[2] $F$: the forecasting horizon
    \end{tablenotes}
\end{threeparttable}
\end{table}

\subsection{Datasets}
\label{sec:Datasets} 
We equip TFB with a set of 25 multivariate and 8,068 univariate datasets with the following desirable properties. 
All datasets are formatted consistently. 
The collection is comprehensive, covering a wide range of domains and characteristics.
The values of key characteristics of the multivariate and univariate datasets, as well as their classification based on the characteristics, can be found in our code repository.
This marks an improvement, addressing challenges such as different formats, varied documentation, and the time-consuming nature of dataset collection. 
We will provide a brief introduction to these datasets (Section~\ref{sec:Datasets introduction}), demonstrating their comprehensiveness (Section~\ref{sec:Comprehensive Datasets}).

\renewcommand{\arraystretch}{0.85} %
\begin{table}[t]
\caption{Statistics of multivariate datasets.}
\label{Multivariate datasets}
\footnotesize
\begin{tabular}{@{}lllrrc@{}}

\toprule
Dataset      & Domain      & Frequency & Lengths & Dim & Split  \\ \midrule
METR-LA     & Traffic     & 5 mins    & 34,272      & 207      & 7:1:2  \\
PEMS-BAY    & Traffic     & 5 mins    & 52,116      & 325      & 7:1:2 \\
PEMS04       & Traffic     & 5 mins    & 16,992      & 307      & 6:2:2 \\
PEMS08       & Traffic     & 5 mins    & 17,856      & 170      & 6:2:2 \\
Traffic      & Traffic     & 1 hour    & 17,544      & 862      & 7:1:2 \\
ETTh1       & Electricity & 1 hour     & 14,400      & 7        & 6:2:2 \\
ETTh2       & Electricity & 1 hour    & 14,400      & 7        & 6:2:2 \\
ETTm1       & Electricity & 15 mins   & 57,600      & 7        & 6:2:2 \\
ETTm2       & Electricity & 15 mins   & 57,600      & 7        & 6:2:2 \\
Electricity  & Electricity & 1 hour    & 26,304      & 321      & 7:1:2 \\
Solar       & Energy      & 10 mins   & 52,560      & 137      & 6:2:2  \\
Wind        & Energy & 15 mins   & 48,673      & 7        & 7:1:2  \\
Weather     & Environment & 10 mins   & 52,696      & 21       & 7:1:2 \\
AQShunyi     & Environment & 1 hour    & 35,064      & 11       & 6:2:2 \\
AQWan        & Environment & 1 hour    & 35,064      & 11       & 6:2:2 \\
ZafNoo       & Nature      & 30 mins   & 19,225      & 11       & 7:1:2 \\
CzeLan      & Nature      & 30 mins   & 19,934      & 11       & 7:1:2 \\
FRED-MD       & Economic    & 1 month   & 728        & 107      & 7:1:2 \\
Exchange & Economic    & 1 day      & 7,588       & 8        & 7:1:2 \\
NASDAQ       & Stock       & 1 day     & 1,244       & 5        & 7:1:2 \\
NYSE         & Stock       & 1 day     & 1,243       & 5        & 7:1:2 \\
NN5         & Banking     & 1 day     & 791        & 111      & 7:1:2 \\
ILI          & Health      & 1 week     & 966        & 7        & 7:1:2 \\
Covid-19     & Health      & 1 day     & 1,392       & 948      & 7:1:2 \\ 
Wike2000     & Web         & 1 day     & 792        & 2,000     & 7:1:2  \\ \bottomrule
\end{tabular}
\end{table}
\renewcommand{\arraystretch}{1}

\subsubsection{Dataset overview} 
\label{sec:Datasets introduction}

\noindent
\\\textbf{Univariate time series.} The univariate datasets are carefully curated from 16 open-source datasets, thus covering dozens of domains.
To fully reflect the complexity of real-world time series, we employed Principal Feature Analysis~(PFA)~\cite{lu2007feature}, a variation of Principal Component Analysis~(PCA)~\cite{bro2014principal}. 
PFA preserves the original values of individual time series data points. We employ the concept of explained variance, representing the ratio between the variance of a single time series and the sum of variances across all individual time series. A threshold $t$ for the explained variance is set to 0.9. This implies that for each dataset, we choose to retain the minimum number of time series required to encompass 90\% of the variance contributed by the remaining time series. As a result, the selected data exhibits pronounced heterogeneity. Compared to datasets with strong homogeneity, it can better reflect methods performance. 
In the end, we select 8,068 time series to enable the combined dataset to capture the diversity of real-world time series. Statistical information is reported in Table~\ref{Univariate datasets}. 

\noindent
\textbf{Multivariate time series. } 
Table~\ref{Multivariate datasets} lists statistics of the 25 multivariate time series datasets, which cover 10 domains. The frequencies vary from 5 minutes to 1 month, the range of feature dimensions varies from 5 to 2,000, and the sequence length varies from 728 to 57,600. This substantial diversity of the datasets enables comprehensive studies of forecasting methods. To ensure fair comparisons, we choose a fixed data split ratio for each dataset chronologically, i.e., 7:1:2 or 6:2:2, for training, validation and testing. 

    \begin{figure}[t]
    \centering
    \includegraphics[width=1\linewidth, trim=30 15 20 20, clip]{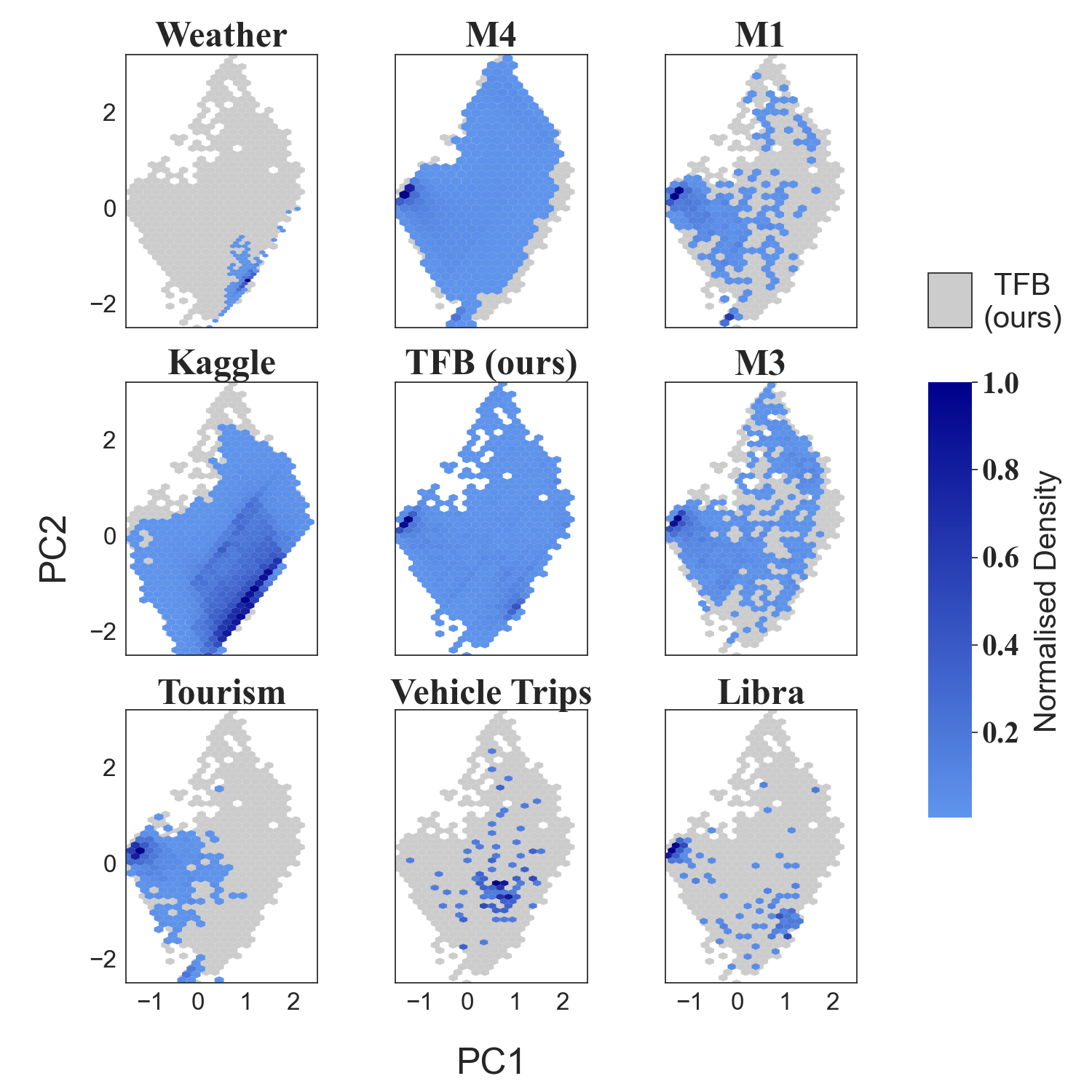}
    \caption{Hexbin plots showing normalised density values of the low-dimensional feature spaces generated by PCA across trend, seasonality, stationarity, shifting, and transition for 9 univariate datasets.}
    \label{fig:Hexbin}
\end{figure}

\subsubsection{Dataset comprehensiveness}
\label{sec:Comprehensive Datasets}
We proceed to investigate the coverage of the selected univariate and multivariate datasets.

\noindent
\textbf{Univariate time series. }
Since time series have different lengths, we first represent time series as vectors that consist of five feature indicators: trend, seasonality, stationarity, shifting, and transition. For ease of visualization, we adopt PCA~\cite{bro2014principal} to reduce the dimensionality from five to two and visualize the eight most widely distributed univariate time series datasets in hexbin---see Figure~\ref{fig:Hexbin}. We observe that TFB and M4 cover the most cells, while all other benchmarks are smaller than TFB. This emphasizes the coverage of our dataset in terms of diversity in characteristic distribution. In addition, compared to M4, our dataset covers a wider range of domains. Further, we note that M4 has a much larger sample size, totaling 100,000, compared to our dataset that contains only 8,068 time series. 
We believe that testing on diverse datasets is essential to better reflect the practical performance of methods. Yet, we can run much fewer experiments with TFB than the M4 dataset, i.e., around 8\%.

\noindent
\textbf{Multivariate time series. }Figure~\ref{fig:Stacked_bar_chart} shows a comparison of TFB and existing multivariate time series benchmarks according to dataset domain distribution. Our benchmark includes more diverse datasets in terms of both quantity and domain categories.
Next, we select TSlib, whose multivariate time series datasets are the most popular, to compare the characteristic distributions with TFB, which are shown in Figure \ref{fig:multivariate datasets features1}. We can observe that the datasets in TFB also represent a more diverse characteristic distribution.

\subsection{Comparison Methods}
\label{sec:Baselines}
To investigate the strengths and limitations of different forecasting methods, we include 22 methods that can be categorized as statistical learning, machine learning, and deep learning methods. In terms of statistical learning methods, we include ARIMA~\cite{box1970distribution}, ETS~\cite{hyndman2008forecasting}, Kalman Filter~(KF)~\cite{harvey1990forecasting}, and VAR~\cite{toda1994vector}. Among the machine learning methods, we include XGBModel~(XGB)~\cite{herzen2022darts}, LinearRegression~(LR)~\cite{kedem2005regression, herzen2022darts}, and Random Forest~(RF)~\cite{breiman2001random}. 
Finally, we further split the deep learning methods into RNN-based models~(RNN~\cite{kim2021reversible}), CNN-based models~(MICN~\cite{wang2022micn}, TimesNet~\cite{wu2022timesnet}, and TCN~\cite{bai2018empirical}), MLP-based models (NLinear~\cite{zeng2023transformers}, DLinear~\cite{zeng2023transformers}, TiDE~\cite{das2023long}, N-HiTS~\cite{challu2023nhits}, and N-BEATS~\cite{oreshkin2019n}), Transformer-based models~(PatchTST~\cite{nie2022time}, Crossformer~\cite{zhang2022crossformer}, and FEDformer~\cite{zhou2022fedformer}, Non-stationary Transformer (Stationary)~\cite{liu2022non}, Informer~\cite{zhou2021informer}, and Triformer~\cite{cirstea2022triformer}), and Model-Agnostic models (FiLM~\cite{zhou2022FiLM}).

\subsection{Evaluation Settings}
\label{sec:Evaluation Settings}

\subsubsection{Evaluation strategies}
To assess the forecasting accuracy of a method, TFB implements two distinct evaluation strategies: 1)~\textit{Fixed forecasting;} and 2)~\textit{Rolling forecasting}.

\begin{figure}[t]
  \centering
  \subfloat[Fixed Forecasting.]
  {\includegraphics[width=0.23\textwidth]{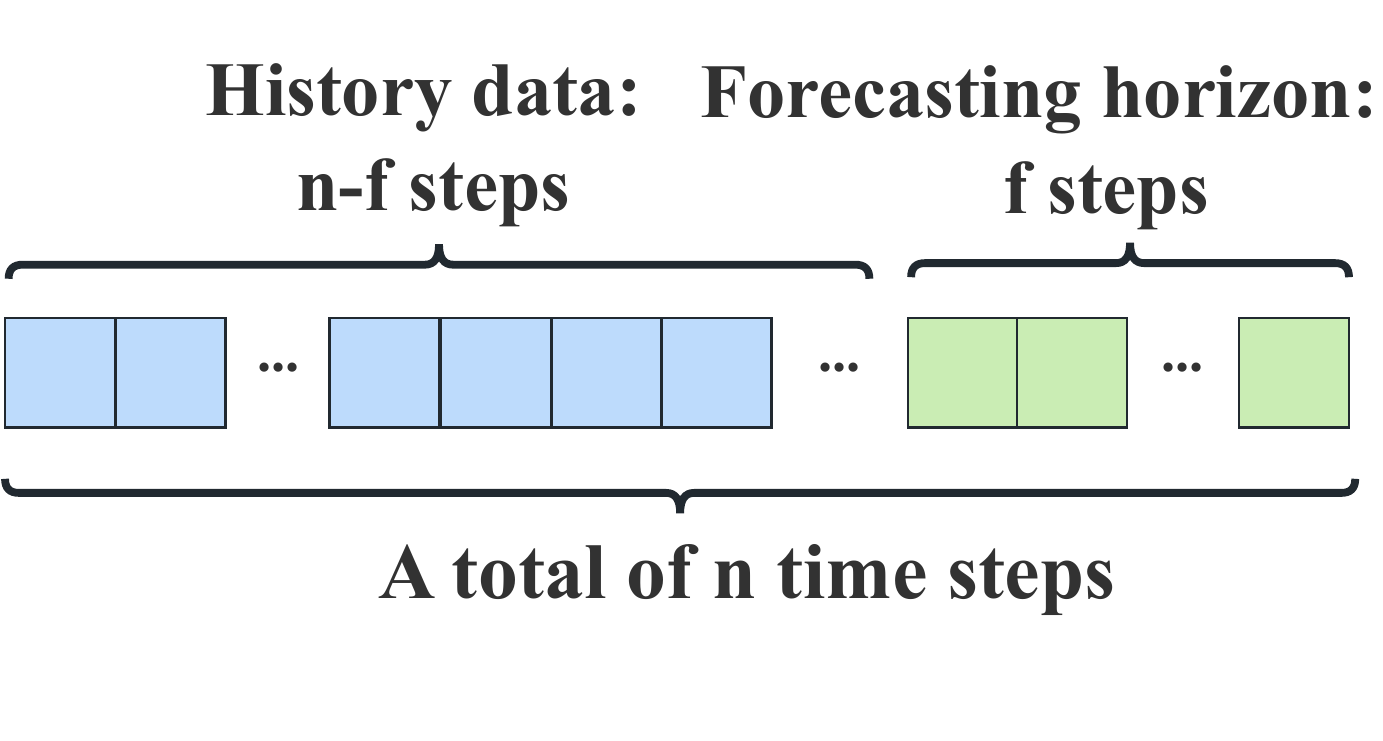}\label{fig:fixed}}
  \subfloat[Rolling Forecasting.]
  {\includegraphics[width=0.23\textwidth]{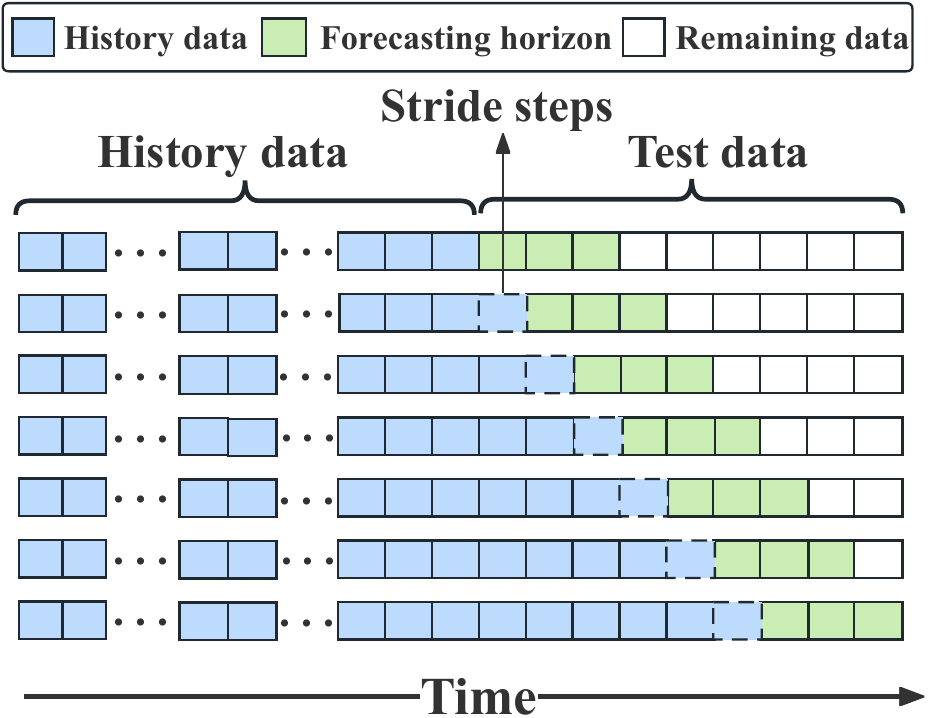}\label{fig:rolling}}
  \caption{Time series forecasting evaluation strategies.}
\end{figure}

\noindent
\textbf{Fixed forecasting. }Given a time series on length $n$, $f$ future time points are predicted from $n - f$ historical time points, which is illustrated in Figure~\ref{fig:fixed}.

\noindent
\textbf{Rolling forecasting. }
In Rolling forecasting, illustrated in Figure \ref{fig:rolling}, the blue squares represent historical data, the green squares represent the forecasting horizon, and the white squares represent the remaining data in the time series. During rolling forecast, excluding the last iteration, the historical data expands by a fixed number of steps (called the stride) in each iteration. In the final iteration, the historical data is extended to cover the entire time series along with the forecasting horizon. In each iteration of the inference process, the forecasting model is applied based on the historical data to forecast the designated forecasting horizon. Then, we calculate the average of the error metrics for each iteration. In statistical learning methods~(e.g., ARIMA~\cite{box1970distribution}, ETS~\cite{garza2022statsforecast}), it is common to use the entire or a subset of the historical data for training and then make forecasting during each iteration. In contrast, in deep learning or machine learning methods, each iteration often involves using only the last portion of the historical data, with a length equal to the look-back windows size, for inference to make forecasting. The current evaluation strategy for TSF, such as TimesNet~\cite{wu2022timesnet}, aligns with our defined criteria.

From a practical perspective, despite some limitations in the generalization capabilities of statistical learning methods, their relatively short runtimes prompts us to adopt the approach of retraining and then making inference predictions to support rolling forecasting. However, in the case of deep learning and machine learning methods, each iteration of retraining typically takes much longer time. Therefore, to balance timeliness and prediction accuracy, machine learning and deep learning methods opt for the strategy of reinferring during each iteration in rolling forecasting.
\subsubsection{Evaluation metrics}

To achieve an extensive evaluation of forecasting performance, we employ eight error metrics, namely the Mean Absolute Error (MAE), Mean Absolute Percentage Error (MAPE), Mean Squared Error (MSE), Symmetric Mean Absolute Percentage Error (SMAPE), Root Mean Squared Error (RMSE), Weighted Absolute Percent Error (WAPE), Modified Symmetric Mean Absolute Percentage Error (MSMAPE)~\cite{msmape}, and Mean Absolute Scaled Error (MASE)~\cite{hyndman2006another}, which are defined in Equations~~\ref{MAE}--~\ref{MASE}.

\begin{minipage}{.42\linewidth}
\footnotesize
\begin{align}
    &\mathit{MAE} = \frac{1}{h}\sum_{k=1}^{h}|F_k-Y_k| \label{MAE} \tag{7}\\
    &\mathit{MSE} = \frac{1}{h}\sum_{k=1}^{h}(F_k - Y_k)^2\tag{9}\\
    &\mathit{RMSE} = \sqrt{\frac{1}{h}\sum_{k=1}^{h}(F_k-Y_k)^2}\tag{11}
\end{align}
\end{minipage}%
\begin{minipage}{.54\linewidth}
\footnotesize
\begin{align}
&\mathit{MAPE} = \frac{1}{h}\sum_{k=1}^{h}\frac{|Y_k - F_k|}{Y_k}\times 100\% \tag{8}\\
    &\mathit{SMAPE} = \frac{100\%}{h}\sum_{k=1}^{h}\frac{|F_k-Y_k|}{\frac{|Y_k|+|F_k|}{2}}\tag{10}\\
    &\mathit{WAPE} = \frac{\sum_{k=1}^{h}|Y_k - F_k|}{\sum_{k=1}^{h}|Y_k|}\tag{12}
\end{align}
\end{minipage}

\begin{equation}
\footnotesize
\mathit{MSMAPE} = \frac{100\%}{h}\sum_{k=1}^{h}\frac{|F_k-Y_k|}{\max(|Y_k|+|F_k|+\epsilon,0.5+\epsilon)/2} \label{MSMAPE}\tag{13}
\end{equation}
\begin{equation}
\label{MASE}
\footnotesize
\mathit{MASE} = \frac{\sum_{k=M+1}^{M+h}|F_k-Y_k|}{\frac{h}{M-S}\sum_{k=S+1}^{M}|Y_k-Y_{k-S}|},\tag{14}
\end{equation}

\noindent
where $M$ is the length of the training series, $S$ is the seasonality of the dataset, $h$ is the forecasting horizon, $F_k$ are the
generated forecasts , and $Y_k$ are the actual values. We set the parameter $\epsilon$ in Equation~\ref{MSMAPE} to its proposed default of 0.1. For rolling forecasting, we further calculate the average of error metrics for all samples~(windows) on each time series to assess the method performance.

\subsection{Unified Pipeline}
\label{sec:Unified Pipeline}

\begin{figure}[t]
    \centering
    \includegraphics[width=1.0\linewidth]{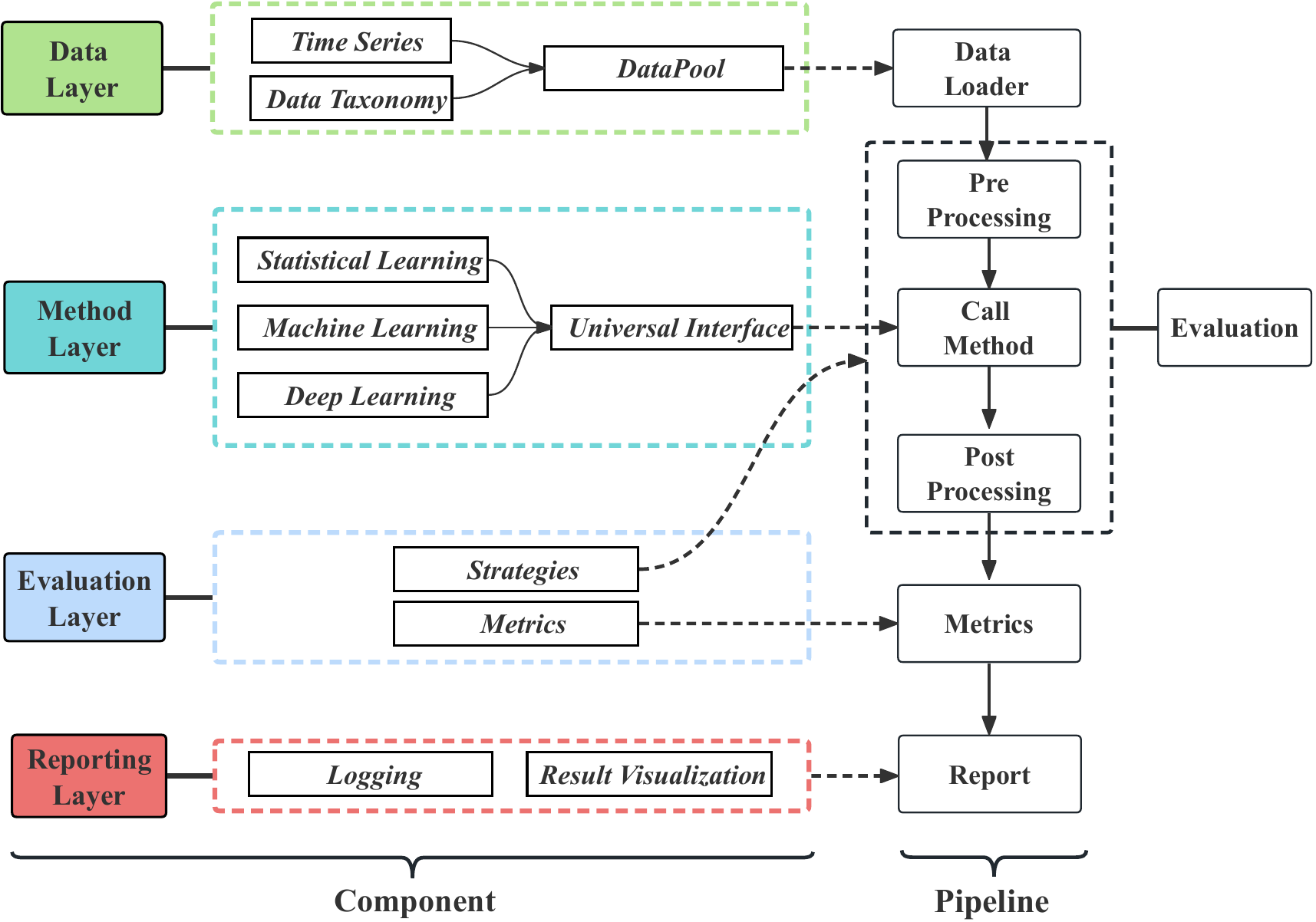}
    \caption{The TFB pipeline.}
    \label{fig:pipeline}
\end{figure}

As mentioned in Section~\ref{sec:intro}, small implementation differences can impact evaluation results profoundly. To enable fair and comprehensive comparisons of methods, we introduce a unified pipeline that is structured into a data layer, a method layer, an evaluation layer, and a reporting layer---see Figure~\ref{fig:pipeline}. The details of each component are listed as follows. 

\begin{itemize}[left=0.1cm]

\item The data layer is a repository of univariate and multivariate time series from diverse domains, structured according to their distinct characteristics, frequencies, and sequence lengths. The data is uniformly according to a standardized format. When a new dataset becomes available, this layer can assess whether the distribution of existing datasets across the six features can be expanded. If yes, it is accepted as a new dataset. 
\item The method layer supports embedding statistical learning, machine learning and deep learning methods. However, other benchmarks have not achieved this, most of them can only embed deep learning methods. Additionally, TFB is designed to be compatible with any third-party TSF library, such as Darts~\cite{herzen2022darts}, TSlib~\cite{wu2022timesnet}. Users can easily integrate forecasting methods implemented in third-party libraries into TFB by writing a simple Universal Interface, facilitating fair comparisons. TFB goes beyond benchmarks that only support Direct Multi-Step (DMS) forecasting to support also Iterative Multi-Step (IMS) forecasting. The method layer thus contributes to the applicability of TFB by offering support for a broad range of methods.

\item The evaluation layer offers support for a diverse range of evaluation strategies and metrics. It thus supports both fixed and rolling forecasting strategies, increasing again the applicability of the benchmark to a wider range of methods and applications. The layer also covers evaluation metrics found in other studies and enables the use of customized metrics for a more comprehensive assessment of method performance. Besides, for each evaluation strategy, TFB offers standardized dataset handling, splitting, and normalization. Additionally, it provides a standard configuration file that can be customized by users. This aims to facilitate deeper understanding of method performance in different settings.

\item The reporting layer encompasses a logging system for tracking information, enabling the capture of experimental settings to enable traceability. Further, it encompasses a visualization module to facilitate a clear understanding of method performance. This design is to offer exhaustive support and transparency throughout the evaluation process. 
\end{itemize}

Users only need to deploy their method architecture at the method layer and choose or configure the configuration file, then TFB can automatically run the pipeline in Figure~\ref{fig:pipeline}.

The TFB pipeline supports various extensible features. Compatibility with CPU and GPU hardware enables evaluation in different computing environments. TFB also supports both sequential and parallel program execution, providing users with multiple options.

In summary, TFB is a unified, flexible, scalable, and user-friendly benchmarking tool for TSF methods. It enables to better understand, compare, and select TSF methods for specific application scenarios.

\section{EXPERIMENTS}
\label{sec:EXPERIMENTS}

We report on experiments with the 14 multivariate and 22 univariate forecasting methods covered in Section~\ref{sec:Baselines} on all the datasets covered in Section ~\ref{sec:Datasets}. We adopt the pipeline covered in Section~\ref{sec:Unified Pipeline} to do so. For each method, we conduct comprehensive hyper-parameters selection, such that its performance result approach or surpass the performance reported in its original paper. Our goal is to showcase TFB as an easy-to-use and powerful resource, providing a convenient and fair evaluation platform for TSF methods.

\subsection{Experimental Setup}

\subsubsection{Datasets and Comparison methods} We include all the datasets included in TFB and all methods mentioned in Section~\ref{sec:Baselines}.

\subsubsection{Implementation details} 
For multivariate forecasting, we use the rolling forecasting strategy. We consider four forecasting horizon $F$: {24, 36, 48, and 60,} for FredMd, NASDAQ, NYSE, NN5, ILI, Covid-19, and Wike2000, and we use another four forecasting horizon, {96, 192, 336, and 720,} for all other datasets which have longer lengths. The look-back window $H$ underwent testing with lengths {36 and 104} for FredMd, NASDAQ, NYSE, NN5, ILI, Covid-19, and Wike2000, and {96, 336, and 512} for all other datasets. For univariate forecasting, we adopt a fixed forecasting strategy to maintain consistency with the setting of the M4 competition~\cite{makridakis2018m4}, forecasting horizons go from 6 to 48, with the look-back window $H$ set to 1.25 times forecasting horizon $F$. 

For each method, we adhere to the hyper-parameter as specified in their original papers. Additionally, we perform hyper-parameter searches across multiple sets, with a limit of 8 sets. The optimal result is then selected from these evaluations, contributing to a comprehensive and unbiased assessment of each method's performance. Due to space constraints, we only report the results of a subset of metrics in the paper, additional metrics results can be found in the code repository.

All experiments are conducted using PyTorch~\cite{paszke2019pytorch} in Python 3.8 and execute on an NVIDIA Tesla-A800 GPU. The training process is guided by the L2 loss, employing the ADAM optimizer. Initially, the batch size is set to 32, with the option to reduce it by half (to a minimum of 8) in case of an Out-Of-Memory (OOM) situation. We do not use the \textit {"Drop Last"} operation during testing. To ensure reproducibility and facilitate experimentation, datasets and code are available at: \href{https://github.com/decisionintelligence/TFB}{https://github.com/decisionintelligence/TFB}.

\renewcommand{\arraystretch}{1.1}
\setlength{\tabcolsep}{1.1pt}
\begin{table*}[ht]
\caption{Univariate forecasting results.}
\label{Univariate forecasting results}
\resizebox{1.8\columnwidth}{!}{
\Huge
\begin{tabular}{cc|c@{\hspace{3pt}}|@{\hspace{3pt}}cccccc@{\hspace{3pt}}|@{\hspace{3pt}}ccccc@{\hspace{3pt}}|@{\hspace{3pt}}cc@{\hspace{3pt}}|@{\hspace{3pt}}c@{\hspace{3pt}}|@{\hspace{3pt}}c@{\hspace{3pt}}|@{\hspace{3pt}}ccc@{\hspace{3pt}}|@{\hspace{3pt}}ccc}
\toprule
\multicolumn{2}{c}{\textbf{Dataset}} & \textbf{Metric} & \textbf{PatchTST} & \textbf{Crossformer} & \textbf{FEDformer} & \textbf{Stationary} & \textbf{Informer} & \textbf{Triformer} & \textbf{DLinear} & \textbf{NLinear} & \textbf{TiDE} & \textbf{N-BEATS} & \textbf{N-HiTS} & \textbf{TimesNet} & \textbf{TCN} & \textbf{RNN} & \textbf{FiLM}  & \textbf{LR} & \textbf{RF} & \textbf{XGB} & \textbf{ARIMA} & \textbf{ETS} & \textbf{KF}\\\addlinespace
\midrule
\addlinespace
\multirow[c]{6}{*}{\rotatebox{90}{\begin{tabular}[c]{@{}c@{}}Seasonality\end{tabular}}} & \multirow[c]{3}{*}{{\begin{tabular}[c]{@{}c@{}}$\surd$\end{tabular}}}& mase & \underline{1.660} & 29.704 & 2.100 & 2.384 & 2.390 & 19.378 & 2.409 & 2.850 & 2.074 & 2.081 & 2.189 & \textbf{1.446} & 24.159 & 30.231 & 1.882  & 7.8e+9 & \uuline{1.649} & 1.715 & 2.830 & 4.091 & 9.002 \\

 & & msmape & \uuline{12.263} & 161.074 & 19.041 & 15.190 & 14.301 & 107.957 & 19.628 & 20.938  & 14.566 & 15.794 & 13.557 & \textbf{10.927} & 121.335 & 149.830 & 13.537 & 19.183 & \underline{12.790} & 13.404 & 19.868 & 26.386 & 57.409\\

 & & Ranks & 167 & 14 & 59 & 40 & 85 & 15 & 22 & 45 & 162 & 168 & 213 & \underline{314} & 14 & 8 & 141  & \textbf{603} & \uuline{425} & 250 & 225 & 92 & 43\\ 
\addlinespace\cline{2-24} \addlinespace
& \multirow[c]{3}{*}{\rotatebox{90}{\begin{tabular}[c]{@{}c@{}}$\times$\end{tabular}}} & mase & 1.639 & 23.704 & 1.879 & 1.638 & 1.601 & 16.496 & 1.949 & 1.733 & 2.526 & 1.677 & 1.678 & \textbf{1.478} & 15.441 & 23.889 & 1.769  & 2.5e+10 & 1.731 & 1.814 & \uuline{1.496} & \underline{1.544} & 3.318\\

& & msmape & \underline{21.671} & 166.859 & 26.766 & 22.050 & \uuline{21.413} & 138.945 & 26.966 & 27.154 & 30.968 & 25.705 & 24.127 & \textbf{20.497} & 140.174 & 162.648 & 22.331 & 33.413 & 25.316 & 26.838 & 24.491 & 25.190 & 44.551 \\

& & Ranks & 214 & 35 & 208 & 200 & 303 & 72 & 61 & 157 & 189 & \underline{383} & \textbf{556} & 371 & 64 & 40 & 138 & 326 & \uuline{423} & 273 & 355 & 268 & 292 \\
\addlinespace \Xhline{3pt} \addlinespace

\multirow[c]{6}{*}{\rotatebox{90}{\begin{tabular}[c]{@{}c@{}}Trend\end{tabular}}} & \multirow[c]{3}{*}{{\begin{tabular}[c]{@{}c@{}}$\surd$ \end{tabular}}}& mase & \uuline{2.220} & 41.287 & 2.758 & 2.651 & 2.658 & 28.091 & 3.016 & 2.914 & 3.316 & 2.512 & 2.553 & \textbf{1.911} & 28.716 & 44.365 & 2.492 & 7.9e+9 & \underline{2.271} & 2.355 & 2.822 & 3.615 & 8.486 \\

 && msmape & \uuline{10.679} & 184.125 & 13.917 & 11.709 & 11.216 & 133.424 & 14.435 & 13.463  & 13.747 & 11.583 & 11.090 & \textbf{9.247} & 136.243 & 180.218 & 11.442 & 16.920 & \underline{10.832} & 11.221 & 11.686 & 12.986 & 50.062 \\

 && Ranks & 201 & 2 & 114 & 113 & 188 & 20 & 51 & 113 & 201 & 270 & 375 & \underline{402} & 6 & 2 & 162 & \textbf{737} & 298 & 246 & \uuline{403} & 222 & 123 \\
\addlinespace\cline{2-24} \addlinespace

& \multirow[c]{3}{*}{{\begin{tabular}[c]{@{}c@{}}$\times$\end{tabular}}} & mase & \uuline{1.007} & 8.941 & 1.077 & 1.127 & 1.064 & 5.878 & 1.131 & 1.326 & 1.272 & 1.073 & 1.116 & \textbf{0.968} & 7.718 & 8.282 & \underline{1.052} & 3.1e+10 & 1.059 & 1.127 & 1.104 & 1.311 & 2.185 \\

 && msmape & \uuline{26.261} & 142.824 & 34.808 & 27.894 & \underline{26.993} & 119.760 & 34.972 & 37.374 & 36.802 & 33.385 & 30.053 & \textbf{25.243} & 129.153 & 136.076 & 27.307 & 40.210 & 31.262 & 33.307 & 35.019 & 39.814 & 48.911 \\

& & Ranks & 180 & 47 & 153 & 127 & 200 & 67 & 32 & 89 & 150 & 281 & \uuline{394} & \underline{283} & 72 & 46 & 117 & 192 & \textbf{550} & 277 & 177 & 138 & 212 \\
\addlinespace \Xhline{3pt} \addlinespace

\multirow[c]{6}{*}{\rotatebox{90}{\begin{tabular}[c]{@{}c@{}}Stationarity\end{tabular}}} & \multirow[c]{3}{*}{{\begin{tabular}[c]{@{}c@{}}$\surd$\end{tabular}}}& mase & \uuline{1.004} & 9.380 & 1.057 & 1.132 & 1.133 & 6.309 & 1.139 & 1.290 & 1.343 & 1.162 & 1.212 & \textbf{0.961} & 7.870 & 8.305 & 1.066 & 15.848 & \underline{1.043} & 1.100 & 1.257 & 1.618 & 3.172 \\

 && msmape & \uuline{27.024} & 135.888 & 35.122 & 28.539 & \underline{27.546} & 114.323 & 35.434 & 37.306 & 37.594 & 33.519 & 31.080 & \textbf{26.120} & 122.194 & 129.738 & 28.172 & 38.320 & 32.232 & 34.281 & 36.212 & 41.513 & 52.234 \\

 && Ranks & 154 & 45 & 128 & 105 & 177 & 67 & 24 & 69 & 124 & 214 & \uuline{285} & \underline{242} & 66 & 42 & 99 & 197 & \textbf{444} & 219 & 150 & 117 & 180 \\
\addlinespace\cline{2-24} \addlinespace

& \multirow[c]{3}{*}{{\begin{tabular}[c]{@{}c@{}}$\times$\end{tabular}}} &  mase & \uuline{2.065} & 36.826 & 2.553 & 2.451 & 2.408 & 24.945 & 2.768 & 2.732 & 3.007 & 2.269 & 2.305 & \textbf{1.793} & 25.907 & 39.090 & 2.297 & 3.1e+10 & \underline{2.125} & 2.214 & 2.500 & 3.118 & 7.032 \\

 && msmape & \uuline{12.206} & 183.264 & 16.425 & 13.397 & 12.904 & 135.177 & 16.800 & 16.610 & 16.225 & 14.325 & 12.885 & \textbf{10.754} & 139.836 & 178.312 & 12.940 & 21.168 & \underline{12.853} & 13.455 & 13.942 & 15.365 & 47.757 \\
 
& & Ranks & 227 & 4 & 139 & 135 & 211 & 20 & 59 & 133 & 227 & 337 & \uuline{484} & \underline{443} & 12 & 6 & 180 & \textbf{732} & 404 & 304 & 430 & 243 & 155 \\
\addlinespace \Xhline{3pt} \addlinespace

\multirow[c]{6}{*}{\rotatebox{90}{\begin{tabular}[c]{@{}c@{}}Transition\end{tabular}}} & \multirow[c]{3}{*}{{\begin{tabular}[c]{@{}c@{}}$\surd$\end{tabular}}}& mase & \underline{1.397} & 19.759 & 1.571 & 1.744 & 1.779 & 11.972 & 1.820 & 1.985 & 1.827 & 1.543 & 1.601 & \textbf{1.282} & 13.744 & 19.901 & 1.505 & 5.7e+4 & \uuline{1.380} & 1.474 & 1.930 & 2.723 & 3.998 \\

 & & msmape & \uuline{21.932} & 155.700 & 24.803 & 23.707 & 22.978 & 117.240 & 25.741 & 29.013  & 27.664 & 23.031 & \underline{22.672} & \textbf{20.869} & 125.136 & 150.624 & 22.973 & 29.179 & 23.002 & 24.545 & 25.020 & 28.725 & 45.521 \\
 
 & & Ranks & 242 & 44 & 187 & 153 & 263 & 79 & 44 & 121 & 230 & 373 & \underline{527} & 464 & 73 & 48 & 187 & \textbf{560} & \uuline{533} & 382 & 304 & 186 & 166 \\
\addlinespace\cline{2-24} \addlinespace

& \multirow[c]{3}{*}{{\begin{tabular}[c]{@{}c@{}}$\times$\end{tabular}}} & mase & \uuline{2.102} & 37.372 & 2.676 & 2.277 & \underline{2.136} & 27.831 & 2.682 & 2.489 &  3.303 & 2.358 & 2.371 & \textbf{1.799} & 27.987 & 39.651 & 2.369 & 5.3e+10 & 2.278 & 2.323 & 2.157 & 2.172 & 8.259 \\

  && msmape & \underline{10.984} & 180.775 & 21.932 & 11.399 & \uuline{10.860} & 144.591 & 21.216 & 17.039 &  19.143 & 19.785 & 15.284 & \textbf{9.435} & 146.942 & 173.676 & 11.622 & 25.629 & 15.905 & 16.404 & 18.520 & 20.089 & 56.754 \\

 & & Ranks & 139 & 5 & 80 & 87 & 125 & 8 & 39 & 81 & 121 & 178 & 242 & 221 & 5 & 0 & 92 & \textbf{369} & \uuline{315} & 141 & \underline{276} & 174 & 169 \\
\addlinespace \Xhline{3pt} \addlinespace

\multirow[c]{6}{*}{\rotatebox{90}{\begin{tabular}[c]{@{}c@{}}Shifting\end{tabular}}} & \multirow[c]{3}{*}{{\begin{tabular}[c]{@{}c@{}}$\surd$\end{tabular}}}& mase & \uuline{2.138} & 36.092 & 2.646 & 2.507 & 2.314 & 25.570 & 2.823 & 2.747 & 2.975 & 2.289 & 2.345 & \textbf{1.857} & 24.925 & 37.921 & 2.352 & 3.7e+10 & \underline{2.224} & 2.306 & 2.331 & 2.799 & 6.862 \\

 & & msmape & \uuline{13.453} & 173.924 & 19.874 & 14.454 & \underline{13.509} & 133.554 & 19.930 & 19.013 &  17.860 & 16.573 & 14.248 & \textbf{11.873} & 137.113 & 167.496 & 14.074 & 21.775 & 14.877 & 16.159 & 17.550 & 20.517 & 50.844 \\

 & & Ranks & 181 & 13 & 107 & 86 & 194 & 27 & 47 & 123 & 173 & 268 & 380 & \underline{384} & 14 & 18 & 143 & \textbf{556} & \uuline{401} & 220 & 370 & 227 & 157 \\
\addlinespace\cline{2-24} \addlinespace

& \multirow[c]{3}{*}{{\begin{tabular}[c]{@{}c@{}}$\times$\end{tabular}}} & mase & \uuline{1.142} & 15.639 & 1.262 & 1.355 & 1.485 & 9.401 & 1.410 & 1.563 & 1.709 & 1.363 & 1.390 & \textbf{1.062} & 12.499 & 15.066 & 1.257 & 7.4e+4 & \underline{1.159} & 1.229 & 1.681 & 2.248 & 4.122 \\

  && msmape & \uuline{22.763} & 155.031 & 27.811 & 24.258 & 23.983 & 120.183 & 28.465 & 30.674  & 31.618 & 27.347 & 26.022 & \textbf{21.881} & 128.544 & 148.943 & \underline{23.945} & 34.251 & 26.254 & 27.310 & 28.022 & 30.950 & 48.150 \\

 & & Ranks & 200 & 36 & 160 & 154 & 194 & 60 & 36 & 79 & 178 & 283 & \uuline{389} & 301 & 64 & 30 & 136 & \underline{373} & \textbf{447} & 303 & 210 & 133 & 178\\
\addlinespace
\bottomrule
\end{tabular}
}
\end{table*}

\subsection{Experimental Results}

\subsubsection{Univariate time series forecasting}
\label{Progress in univariate time series forecasting}

Table~\ref{Univariate forecasting results} reports the average results of UTSF in terms of the metrics MASE, MSMAPE, and the Ranks in MSMAPE that indicates how many times the best performance is achieved on the datasets. 
We observe that the recently proposed deep learning methods, including TimesNet~\cite{wu2022timesnet}, PatchTST~\cite{nie2022time}, and N-HiTS~\cite{challu2023nhits}, exhibit substantially better average performance on univariate datasets in terms of MASE and MSMAPE. However, when considering Ranks, the (non-deep) machine learning methods LinearRegression~(LR)~\cite{kedem2005regression,herzen2022darts} and RandomForest~(RF)~\cite{breiman2001random} outperform all competitors. This suggests that the machine learning methods may be more suitable in specific scenarios. In this scenario, each individual univariate time series is adopted to train a separate model, and deep learning methods require large amount of training data to be effective. Therefore, 
the performance of the deep learning methods falls short. Next, we can observe that LR performs better on the time series with seasonality, trend, and shifting characteristics, while RF is better when those patterns are absent. Further, we notice that LR is more suitable for data without stationarity than RF. Finally, both LR and RF are sensitive to the transition characteristic: the stronger the characteristic, the better. These results offer guidance on choosing the right method for a specific setting.

\subsubsection{Multivariate time series forecasting}
Due to the large number of results, we report them in two tables, Tables ~\ref{Common Multivariate forecasting results} and ~\ref{New Multivariate forecasting results}. The datasets in these tables are ordered based on their trend characteristic, where datasets with weaker trend are occur first. In both tables, we report the MAE and MSE on normalized data considering four different forecasting horizons for each dataset. "nan" represents the inability of a method to generate effective predictions, while "inf" represents infinite results. 
We see that no single method achieves the best performance on all datasets. We also see that the Transformer-based methods generally outperform other methods on datasets with weak trend. Next, Linear-based methods tend to perform moderately better on datasets with strong trend. Surprisingly, we observe that recent methods not consistently outperform earlier studies, such as Informer, LR, and VAR. This discovery highlights the need for evaluating method performance across diverse datasets.
Evaluating methods on relatively few datasets render it difficult to accurately assess their universality and overall performance. Therefore, it is crucial to expand the range of datasets used in evaluations.

\begin{figure}[t]
    \centering
    \includegraphics[width=0.9\linewidth]{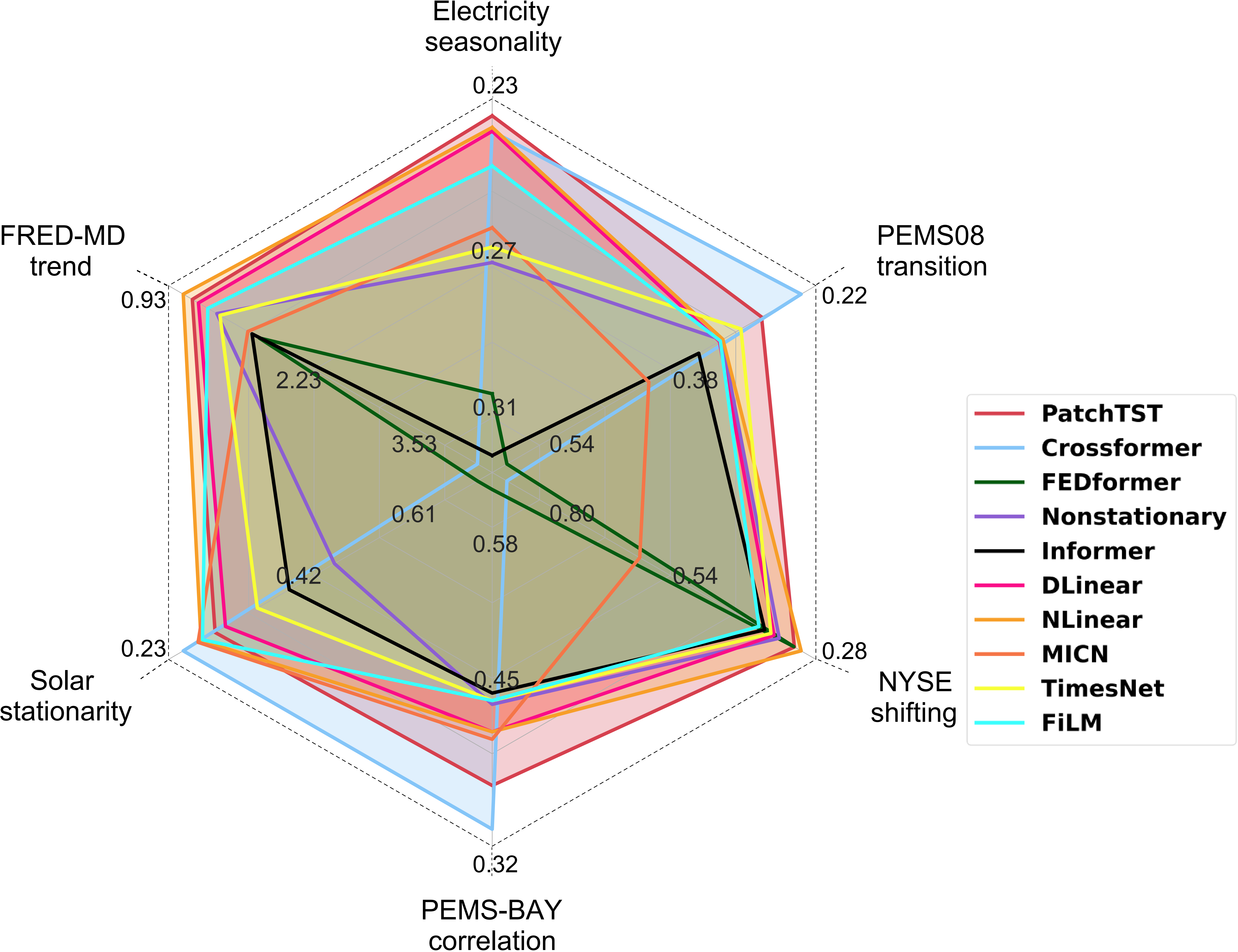}
    \caption{MAE results of methods across six characteristics.}
    \label{fig:leida}
\end{figure}
\subsubsection{Performance on different characteristics}
We proceed to study the performance of different deep learning methods with respect to different characteristics. First, we score the multivariate time series with respect to the six characteristics we consider. Then, we select the dataset with the highest score for each characteristic, which are FRED-MD on trend, Electricity on seasonality, PEMS08 on transition, NYSE on shifting, PEMS-BAY on correlation, and Solar on stationarity. Next, we show the best MAE results for the methods in a radar figure. We use a forecasting horizon of 24 for FRED-MD and NYSE and of 96 for other datasets---see Figure~\ref{fig:leida}. 
We see that no deep learning method excels on all datasets. In particular, Crossformer demonstrates exceptional performance on datasets where transition is highly pronounced~(PEMS08), the data is most stationary~(Solar), and the data is most correlated~(PEMS-BAY). However, Crossformer's performance is noticeably inferior to other methods on time series with 
other characteristics. Next, PatchTST achieves optimal performance on datasets with strong seasonality~(Electricity). Similarly, NLinear delivers outstanding results on time series with the most significant trend~(FRED-MD) and severe drift~(NYSE). Both PatchTST and NLinear perform well consistently, without any notably poor outcomes.

\renewcommand{\arraystretch}{1}
\begin{table*}[t]
\caption{Multivariate forecasting results I.}
\label{Common Multivariate forecasting results}

\resizebox{1.9\columnwidth}{!}{
\Huge
\begin{tabular}{@{}cc@{\hspace{3pt}}|@{\hspace{3pt}}cccccccccc@{\hspace{3pt}}|@{\hspace{3pt}}cccc@{\hspace{3pt}}|@{\hspace{3pt}}cccccc@{\hspace{3pt}}|@{\hspace{3pt}}cc@{\hspace{3pt}}|@{\hspace{3pt}}cc@{\hspace{3pt}}|@{\hspace{3pt}}cc@{\hspace{3pt}}|@{\hspace{3pt}}cc@{}}
\toprule
  \multicolumn{2}{c@{\hspace{3pt}}|@{\hspace{3pt}}}{\textbf{Model}} & \multicolumn{2}{c}{\textbf{PatchTST}} & \multicolumn{2}{c}{\textbf{Crossformer}} & \multicolumn{2}{c}{\textbf{FEDformer}} & \multicolumn{2}{c}{\textbf{Informer}} & \multicolumn{2}{c@{\hspace{3pt}}|@{\hspace{3pt}}}{\textbf{Triformer}} & \multicolumn{2}{c}{\textbf{DLinear}} & \multicolumn{2}{c@{\hspace{3pt}}|@{\hspace{3pt}}}{\textbf{NLinear}} & \multicolumn{2}{c}{\textbf{MICN}} & \multicolumn{2}{c}{\textbf{TimesNet}} & \multicolumn{2}{c@{\hspace{3pt}}|@{\hspace{3pt}}}{\textbf{TCN}} & \multicolumn{2}{c@{\hspace{3pt}}|@{\hspace{3pt}}}{\textbf{FiLM}} & \multicolumn{2}{c@{\hspace{3pt}}|@{\hspace{3pt}}}{\textbf{RNN}} & \multicolumn{2}{c@{\hspace{3pt}}|@{\hspace{3pt}}}{\textbf{LR}} & \multicolumn{2}{c}{\textbf{VAR}} \\
\multicolumn{2}{c@{\hspace{3pt}}|@{\hspace{3pt}}}{Metrics}  & mae & mse & mae & mse & mae & mse & mae & mse & mae & mse & mae & mse & mae & mse & mae & mse & mae & mse & mae & mse & mae & mse & mae & mse & mae & mse & mae & mse \\

\midrule
\multirow[c]{4}{*}{\rotatebox{90}{PEMS04}} & 96 & \underline{0.280} & \underline{0.161} & \textbf{0.224} & \textbf{0.112} & 0.565 & 0.573 & 0.304 & 0.189 & 0.325 & 0.218 & 0.296 & 0.196 & 0.294 & 0.202 & 0.318 & 0.199 & \uuline{0.266} & \uuline{0.159} & 0.291 & 0.168 & 0.297 & 0.204 & 0.916 & 1.318 & 0.474 & 0.423 & 0.313 & 0.192 \\

 & 192 & \underline{0.290} & \uuline{0.178} & \textbf{0.236} & \textbf{0.134} & 0.624 & 0.655 & 0.335 & 0.229 & 0.353 & 0.257 & 0.310 & 0.213 & 0.305 & 0.223 &  0.380 & 0.290 & \uuline{0.282} & \underline{0.179} & 0.303 & 0.188 & 0.307 & 0.219 & 0.933 & 1.359 & 0.497 & 0.465 & 0.336 & 0.216 \\

 & 336 & \underline{0.302} & \underline{0.193} & \uuline{0.286} & \uuline{0.190} & 0.920 & 1.365 & 0.323 & 0.217 & 0.347 & 0.250 & 0.327 & 0.235 & 0.324 & 0.245 &  0.453 & 0.384 & \textbf{0.269} & \textbf{0.169} & 0.336 & 0.237 & 0.327 & 0.243 & 0.939 & 1.373 & 0.519 & 0.513 & 0.362 & 0.245 \\

 & 720 & \underline{0.338} & \uuline{0.233} & \uuline{0.331} & \underline{0.235} & 0.728 & 0.873 & 0.391 & 0.310 & 0.375 & 0.284 & 0.395 & 0.327 & 0.396 & 0.350 &  0.629 & 0.677 & \textbf{0.286} & \textbf{0.187} & 0.354 & 0.246 & 0.386 & 0.322 & 0.938 & 1.372 & 0.658 & 0.831 & 0.409 & 0.298 \\
\addlinespace\cline{1-30} \addlinespace
\multirow[c]{4}{*}{\rotatebox{90}{PEMS-BAY}} & 96 & \uuline{0.361} & \uuline{0.567} & \textbf{0.326} & \textbf{0.492} & 0.540 & 0.815 & 0.435 & 0.844 & 0.472 & 0.785 & 0.404 & \underline{0.628} & 0.404 & 0.642  & \underline{0.398} & 0.664 & 0.429 & 0.908 & 0.680 & 1.300 & 0.429 & 0.660 & 0.659 & 1.198 & 1.192 & 2.934 & 0.516 & 0.761 \\

 & 192 & \uuline{0.382} & \uuline{0.619} & \textbf{0.336} & \textbf{0.490} & 0.583 & 0.872 & 0.461 & 0.912 & 0.500 & 0.854 & 0.419 & \underline{0.666} & 0.420 & 0.687  & 0.429 & 0.727 & 0.448 & 0.997 & 0.710 & 1.327 & \underline{0.404} & 0.677 & 0.659 & 1.201 & 1.176 & 2.776 & 0.545 & 0.863 \\

 & 336 & \uuline{0.403} & \uuline{0.670} & \textbf{0.342} & \textbf{0.529} & 0.609 & 1.026 & 0.447 & 0.834 & 0.477 & 0.805 & 0.439 & \underline{0.710} & 0.438 & 0.735 &  0.481 & 0.919 & 0.429 & 0.893 & 0.573 & 1.040 & \underline{0.421} & 0.726 & 0.660 & 1.204 & 1.071 & 2.310 & 0.581 & 0.966 \\

 & 720 & \uuline{0.448} & \textbf{0.795} & 0.494 & \uuline{0.850} & 0.659 & 1.124 & 0.498 & 1.005 & 0.519 & 0.900 & 0.508 & \underline{0.872} & 0.516 & 0.929 & 0.596 & 1.105 & \textbf{0.447} & 0.962 & 0.544 & 1.089 & \underline{0.482} & 0.895 & 0.660 & 1.210 & 0.959 & 1.844 & 0.617 & 1.082 \\
\addlinespace\cline{1-30} \addlinespace
\multirow[c]{4}{*}{\rotatebox{90}{METR-LA}} & 96 & \textbf{0.629} & 1.058 & \uuline{0.633} & 1.232 & 0.770 & 1.335 & 0.664 & 1.375 & 0.640 & \underline{1.048} & 0.658 & \textbf{1.009} & 0.650 & \uuline{1.044}  & 0.685 & 1.256 & \underline{0.640} & 1.323 & 0.716 & 1.409 & 0.682 & 1.282 & 0.774 & 1.668 & 1.968 & 7.219 & 0.706 & 1.100 \\

 & 192 & \uuline{0.676} & \uuline{1.177} & \textbf{0.663} & 1.312 & 0.887 & 1.645 & 0.707 & 1.554 & 0.714 & 1.222 & 0.716 & \textbf{1.148} & 0.720 & \underline{1.218} & 0.701 & 1.359 & \underline{0.681} & 1.490 & 0.860 & 1.715 & 0.733 & 1.223 & 0.784 & 1.727 & 1.955 & 6.797 & 0.754 & 1.276 \\

 & 336 & 0.733 & \underline{1.300} & \uuline{0.711} & 1.367 & 0.840 & 1.687 & \underline{0.721} & 1.615 & 0.735 & \uuline{1.286} & 0.742 & \textbf{1.241} & 0.758 & 1.339  & 0.741 & 1.390 & \textbf{0.688} & 1.502 & 0.899 & 1.811 & 0.745 & 1.323 & 0.791 & 1.765 & 1.762 & 5.511 & 0.786 & 1.375 \\

 & 720 & \uuline{0.779} & \underline{1.466} & \textbf{0.773} & 1.516 & 0.962 & 2.018 & 0.805 & 1.878 & 0.782 & \uuline{1.437} & 0.793 & \textbf{1.415} & 0.874 & 1.656 &  \underline{0.780} & 1.486 & 0.784 & 1.801 & 0.854 & 1.898 & 0.825 & 1.574 & 0.809 & 1.849 & 1.382 & 3.472 & 0.817 & 1.478 \\
\addlinespace\cline{1-30} \addlinespace
\multirow[c]{4}{*}{\rotatebox{90}{PEMS08}} & 96 & \uuline{0.272} & \uuline{0.163} & \textbf{0.226} & \textbf{0.119} & 0.568 & 0.623 & 0.345 & 0.247 & 0.329 & 0.226 & 0.321 & 0.241 & 0.317 & 0.259 & 0.403 & 0.310 & \underline{0.296} & \underline{0.213} & 0.393 & 0.299 & 0.320 & 0.252 & 0.879 & 1.071 & 4.198 & 78.523 & 0.363 & 0.267 \\

 & 192 & \uuline{0.295} & \uuline{0.201} & \textbf{0.259} & \textbf{0.151} & 0.690 & 0.782 & 0.409 & 0.351 & 0.365 & 0.279 & 0.342 & 0.273 & 0.338 & 0.300 & 0.408 & 0.346 & \underline{0.321} & \underline{0.264} & 0.393 & 0.289 & 0.341 & 0.290 & 0.880 & 1.072 & 4.919 & 52.019 & 0.435 & 0.393 \\

 & 336 & \uuline{0.311} & \uuline{0.225} & \textbf{0.276} & \textbf{0.180} & 0.859 & 1.256 & 0.386 & 0.342 & 0.365 & 0.287 & 0.362 & 0.299 & 0.358 & 0.325 & 0.438 & 0.358 & \underline{0.315} & \underline{0.264} & 0.446 & 0.394 & 0.360 & 0.309 & 0.880 & 1.074 & 3.205 & 20.396 & 0.474 & 0.467 \\

 & 720 & \underline{0.347} & \uuline{0.264} & \textbf{0.332} & \textbf{0.254} & 0.706 & 0.843 & 0.467 & 0.461 & 0.398 & 0.330 & 0.429 & 0.378 & 0.426 & 0.403 & 0.641 & 0.732 & \uuline{0.343} & \underline{0.311} & 0.470 & 0.466 & 0.413 & 0.365 & 0.883 & 1.080 & 2.331 & 10.680 & 0.556 & 0.591 \\
\addlinespace\cline{1-30} \addlinespace
\multirow[c]{4}{*}{\rotatebox{90}{Traffic}} & 96 & \textbf{0.271} & \textbf{0.379} & \underline{0.282} & 0.514 & 0.365 & 0.593 & 0.371 & 0.664 & 0.323 & 0.589 & 0.282 & \uuline{0.410} & \uuline{0.279} & \underline{0.410}  & 0.295 & 0.494 & 0.313 & 0.600 & 0.606 & 1.139 & 0.284 & 0.411 & 1.604 & 5.834 & 0.548 & 0.799 & 1.056 & 2.355 \\

 & 192 & \uuline{0.277} & \textbf{0.394} & \textbf{0.273} & 0.501 & 0.375 & 0.614 & 0.396 & 0.724 & 0.325 & 0.597 & 0.288 & \underline{0.423} & 0.284 & 0.423 & 0.301 & 0.521 & 0.328 & 0.619 & 0.559 & 1.040 & \underline{0.283} & \uuline{0.406} & 2.120 & 8.563 & 0.551 & 0.794 & 0.869 & 1.751 \\

 & 336 & \uuline{0.281} & \textbf{0.404} & \textbf{0.279} & 0.507 & 0.373 & 0.609 & 0.435 & 0.796 & 0.332 & 0.617 & 0.296 & \underline{0.436} & \underline{0.291} & 0.436 &  0.309 & 0.552 & 0.330 & 0.627 & 0.540 & 1.006 & 0.298 & \uuline{0.425} & 2.405 & 10.153 & 0.552 & 0.802 & 0.752 & 1.397 \\

 & 720 & \uuline{0.302} & \textbf{0.442} & \textbf{0.301} & 0.571 & 0.394 & 0.646 & 0.453 & 0.823 & 0.350 & 0.650 & 0.315 & \underline{0.466} & \underline{0.308} & \uuline{0.464} &  0.328 & 0.569 & 0.342 & 0.659 & 0.626 & 1.063 & 0.370 & 0.523 & 2.644 & 11.596 & 0.568 & 0.837 & 0.644 & 1.113 \\
\addlinespace\cline{1-30} \addlinespace
\multirow[c]{4}{*}{\rotatebox{90}{Solar}} & 96 & 0.273 & \uuline{0.190} & \textbf{0.230} & \textbf{0.166} & 0.530 & 0.509 & 0.373 & 0.338 & 0.279 & 0.225 & 0.287 & 0.216 & \underline{0.254} & 0.222 & \uuline{0.250} & \underline{0.193} & 0.330 & 0.285 & 0.265 & 0.206 & 0.256 & 0.225 & 2.271 & 6.661 & 0.365 & 0.235 & 0.593 & 0.556 \\

 & 192 & 0.302 & \textbf{0.204} & \textbf{0.251} & \uuline{0.214} & 0.500 & 0.474 & 0.391 & 0.375 & 0.295 & 0.250 & 0.305 & 0.244 & \uuline{0.269} & 0.252 & \underline{0.270} & \underline{0.225} & 0.342 & 0.309 & 0.347 & 0.304 & 0.276 & 0.266 & 2.258 & 6.457 & 0.369 & 0.253 & 0.686 & 0.668 \\

 & 336 & \underline{0.293} & \uuline{0.212} & \textbf{0.260} & \textbf{0.203} & 0.439 & 0.338 & 0.416 & 0.417 & 0.298 & 0.261 & 0.319 & 0.263 & \uuline{0.282} & 0.273 & 0.301 & \underline{0.250} & 0.365 & 0.335 & 0.306 & 0.279 & 0.299 & 0.308 & 2.291 & 6.604 & 0.402 & 0.291 & 0.724 & 0.710 \\

 & 720 & 0.310 & \textbf{0.221} & 0.721 & 0.735 & 0.459 & 0.365 & 0.407 & 0.390 & \underline{0.292} & \uuline{0.258} & 0.324 & \underline{0.264} & \uuline{0.282} & 0.273 & 0.362 & 0.323 & 0.355 & 0.346 & \textbf{0.269} & 0.296 & 0.300 & 0.310 & 2.321 & 6.743 & 0.452 & 0.333 & 0.751 & 0.738 \\
\addlinespace\cline{1-30} \addlinespace
\multirow[c]{4}{*}{\rotatebox{90}{ETTm1}} & 96 & 0.343 & \textbf{0.290} & 0.361 & 0.310 & 0.465 & 0.467 & 0.424 & 0.430 & 0.384 & 0.342 & \textbf{0.343} & \uuline{0.299} & \underline{0.343} & \underline{0.301} & 0.349 & 0.303 & 0.398 & 0.377 & 0.643 & 0.700 & \uuline{0.343} & 0.301 & 0.908 & 2.087 & 0.393 & 0.359 & 0.678 & 0.889 \\

 & 192 & 0.368 & \textbf{0.329} & 0.402 & 0.363 & 0.524 & 0.610 & 0.479 & 0.550 & 0.415 & 0.389 & \textbf{0.364} & \uuline{0.334} & \underline{0.365} & 0.337  & 0.368 & \underline{0.335} & 0.411 & 0.405 & 0.616 & 0.698 & \uuline{0.365} & 0.339 & 1.121 & 2.961 & 0.442 & 0.434 & 0.762 & 1.031 \\

 & 336 & 0.390 & \textbf{0.360} & 0.430 & 0.408 & 0.544 & 0.618 & 0.529 & 0.654 & 0.446 & 0.427 & \textbf{0.384} & \uuline{0.365} & \underline{0.384} & 0.371 & 0.388 & \underline{0.366} & 0.437 & 0.443 & 0.764 & 0.900 & \uuline{0.384} & 0.373 & 1.394 & 4.439 & 0.488 & 0.503 & 0.803 & 1.093 \\

 & 720 & 0.422 & \uuline{0.416} & 0.637 & 0.777 & 0.551 & 0.615 & 0.578 & 0.714 & 0.484 & 0.488 & \uuline{0.415} & \underline{0.418} & \underline{0.415} & 0.426 & 0.423 & \textbf{0.409} & 0.464 & 0.495 & 0.779 & 0.979 & \textbf{0.414} & 0.423 & 1.594 & 5.515 & 0.568 & 0.632 & 0.826 & 1.111 \\
\addlinespace\cline{1-30} \addlinespace
\multirow[c]{4}{*}{\rotatebox{90}{Weather}} & 96 & \textbf{0.196} & \uuline{0.149} & \uuline{0.212} & \textbf{0.146} & 0.292 & 0.223 & 0.255 & 0.218 & 0.231 & \underline{0.168} & 0.230 & 0.170 & 0.222 & 0.179 & 0.232 & 0.172 & \underline{0.219} & 0.170 & 0.332 & 0.262 & 0.229 & 0.178 & 0.545 & 0.543 & 0.293 & 0.201 & 0.401 & 0.360 \\

 & 192 & \textbf{0.240} & \textbf{0.193} & \uuline{0.261} & \uuline{0.195} & 0.322 & 0.252 & 0.306 & 0.269 & 0.279 & 0.216 & 0.267 & \underline{0.212} & \underline{0.261} & 0.218 & 0.270 & 0.214 & 0.264 & 0.222 & 0.447 & 0.420 & 0.263 & 0.218 & 0.545 & 0.544 & 0.330 & 0.240 & 0.453 & 0.413 \\

 & 336 & \textbf{0.281} & \textbf{0.244} & 0.325 & 0.268 & 0.371 & 0.327 & 0.340 & 0.320 & 0.322 & 0.271 & 0.305 & \uuline{0.257} & \underline{0.296} & 0.266 & 0.309 & \underline{0.259} & 0.310 & 0.293 & 0.496 & 0.509 & \uuline{0.295} & 0.266 & 0.548 & 0.549 & 0.379 & 0.303 & 0.491 & 0.459 \\

 & 720 & \textbf{0.332} & \uuline{0.314} & 0.380 & 0.330 & 0.419 & 0.424 & 0.390 & 0.392 & 0.379 & 0.358 & 0.356 & \underline{0.318} & 0.344 & 0.334  & \underline{0.342} & \textbf{0.308} & 0.355 & 0.360 & 0.474 & 0.499 & \uuline{0.340} & 0.332 & 0.548 & 0.549 & 0.447 & 0.396 & 0.539 & 0.529 \\
\addlinespace\cline{1-30} \addlinespace
\multirow[c]{4}{*}{\rotatebox{90}{ILI}} & 24 & \textbf{0.835} & \textbf{1.840} & 1.096 & 2.981 & 1.020 & 2.400 & 1.151 & 2.738 & 1.728 & 6.044 & 1.031 & 2.208 & \uuline{0.919} & \uuline{1.998}  & 1.020 & 2.279 & \underline{0.926} & \underline{2.009} & 1.419 & 4.194 & 1.011 & 2.454 & 3.125 & 15.131 & 4.856 & 47.856 & 1.012 & 2.429 \\

 & 36 & \textbf{0.845} & \textbf{1.724} & 1.162 & 3.295 & 1.005 & 2.410 & 1.145 & 2.890 & 1.762 & 6.226 & \underline{0.981} & \underline{2.032} & \uuline{0.916} & \uuline{1.920} & 1.085 & 2.451 & 0.997 & 2.552 & 1.514 & 4.999 & 1.016 & 2.412 & 3.133 & 15.196 & 3.803 & 31.573 & 1.081 & 2.851 \\

 & 48 & \textbf{0.863} & \textbf{1.762} & 1.230 & 3.586 & 1.033 & 2.592 & 1.136 & 2.742 & 1.764 & 6.230 & 1.063 & 2.209 & \underline{0.924} & \uuline{1.895} & 1.077 & 2.440 & \uuline{0.919} & \underline{1.956} & 1.458 & 4.636 & 1.040 & 2.398 & 3.152 & 15.340 & 3.173 & 18.711 & 1.130 & 3.060 \\

 & 60 & \textbf{0.894} & \textbf{1.752} & 1.256 & 3.693 & 1.070 & 2.539 & 1.139 & 2.825 & 1.828 & 6.596 & 1.086 & 2.292 & \uuline{0.947} & \uuline{1.964} & 1.009 & 2.305 & \underline{0.962} & \underline{2.178} & 1.639 & 5.676 & 0.969 & 2.227 & 3.203 & 15.704 & 3.057 & 17.222 & 1.152 & 3.151 \\
\addlinespace\cline{1-30} \addlinespace
\multirow[c]{4}{*}{\rotatebox{90}{Electricity}} & 96 & \textbf{0.233} & \textbf{0.133} & \underline{0.237} & \uuline{0.135} & 0.302 & 0.186 & 0.321 & 0.214 & 0.285 & 0.185 & 0.237 & \underline{0.140} & \uuline{0.236} & 0.141 & 0.262 & 0.156 & 0.267 & 0.164 & 0.433 & 0.371 & 0.246 & 0.154 & 1.651 & 4.368 & 0.631 & 0.731 & 0.359 & 0.264 \\

 & 192 & \uuline{0.248} & \textbf{0.150} & 0.262 & 0.160 & 0.315 & 0.201 & 0.350 & 0.245 & 0.297 & 0.196 & \underline{0.250} & \uuline{0.154} & \textbf{0.248} & \underline{0.155} & 0.285 & 0.173 & 0.280 & 0.180 & 0.435 & 0.371 & 0.261 & 0.168 & 1.657 & 4.392 & 0.651 & 0.771 & 0.370 & 0.282 \\

 & 336 & \uuline{0.267} & \textbf{0.168} & 0.282 & 0.182 & 0.330 & 0.218 & 0.393 & 0.294 & 0.315 & 0.215 & \underline{0.268} & \uuline{0.169} & \textbf{0.264} & \underline{0.171} & 0.296 & 0.184 & 0.292 & 0.190 & 0.442 & 0.371 & 0.284 & 0.189 & 1.663 & 4.415 & 1.004 & 1.796 & 0.380 & 0.296 \\

 & 720 & \textbf{0.295} & \textbf{0.202} & 0.337 & 0.246 & 0.350 & 0.241 & 0.393 & 0.306 & 0.348 & 0.265 & \underline{0.301} & \underline{0.204} & \uuline{0.297} & 0.210 & 0.311 & \uuline{0.203} & 0.307 & 0.209 & 0.468 & 0.420 & 0.340 & 0.249 & 1.670 & 4.445 & 1.039 & 1.931 & 0.399 & 0.320 \\
\addlinespace\cline{1-30} \addlinespace
\multirow[c]{4}{*}{\rotatebox{90}{ETTh1}} & 96 & 0.396 & 0.376 & 0.426 & 0.405 & 0.419 & 0.379 & 0.563 & 0.709 & 0.424 & 0.399 & \textbf{0.392} & \textbf{0.371} & \uuline{0.393} & \underline{0.372} & 0.413 & 0.377 & 0.412 & 0.389 & 0.676 & 0.780 & \underline{0.395} & \uuline{0.372} & 1.093 & 2.778 & 0.440 & 0.420 & 0.708 & 0.923 \\

 & 192 & \underline{0.416} & \textbf{0.399} & 0.442 & 0.413 & 0.443 & 0.419 & 0.570 & 0.724 & 0.448 & 0.444 & \textbf{0.413} & \uuline{0.404} & \uuline{0.413} & 0.405 & 0.435 & \underline{0.405} & 0.443 & 0.440 & 0.723 & 0.870 & 0.423 & 0.415 & 1.561 & 5.760 & 0.506 & 0.512 & 0.770 & 1.025 \\

 & 336 & \uuline{0.432} & \textbf{0.418} & 0.460 & 0.442 & 0.464 & 0.455 & 0.581 & 0.732 & 0.479 & 0.492 & \underline{0.435} & 0.434 & \textbf{0.427} & \underline{0.429} & 0.446 & \uuline{0.426} & 0.465 & 0.482 & 0.775 & 0.978 & 0.446 & 0.449 & 1.912 & 8.029 & 0.584 & 0.633 & 0.799 & 1.058 \\

 & 720 & \uuline{0.469} & \uuline{0.450} & 0.539 & 0.550 & 0.488 & 0.474 & 0.616 & 0.760 & 0.528 & 0.549 & 0.489 & 0.469 & \textbf{0.452} & \textbf{0.436} & 0.500 & 0.483 & 0.501 & 0.525 & 0.804 & 1.006 & \underline{0.472} & \underline{0.461} & 2.359 & 11.006 & 0.699 & 0.808 & 0.810 & 1.057 \\
\addlinespace\cline{1-30} \addlinespace
\multirow[c]{4}{*}{\rotatebox{90}{Exchange}} & 96 & \textbf{0.200} & \underline{0.083} & 0.364 & 0.248 & 0.267 & 0.136 & 0.254 & 0.122 & 0.336 & 0.219 & \uuline{0.204} & \textbf{0.082} & \underline{0.204} & 0.085 & 0.208 & \uuline{0.082} & 0.238 & 0.109 & 0.671 & 0.727 & 0.212 & 0.092 & 0.700 & 0.831 & 0.331 & 0.226 & 0.263 & 0.135 \\

 & 192 & \uuline{0.298} & \underline{0.176} & 0.511 & 0.474 & 0.353 & 0.239 & 0.324 & 0.197 & 0.509 & 0.483 & 0.325 & 0.186 & \textbf{0.297} & \uuline{0.175} & \underline{0.302} & \textbf{0.163} & 0.331 & 0.213 & 0.876 & 1.175 & 0.308 & 0.182 & 1.019 & 1.633 & 0.550 & 0.625 & 0.399 & 0.298 \\

 & 336 & \textbf{0.397} & \uuline{0.301} & 0.723 & 0.829 & 0.486 & 0.438 & 0.421 & 0.330 & 0.681 & 0.812 & 0.435 & 0.328 & \underline{0.409} & 0.320 & 0.420 & \textbf{0.298} & 0.435 & 0.358 & 1.065 & 1.763 & \uuline{0.409} & \underline{0.318} & 1.289 & 2.489 & 0.740 & 1.071 & 0.548 & 0.540 \\

 & 720 & 0.693 & 0.847 & 0.998 & 1.491 & 0.811 & 1.117 & \textbf{0.609} & \textbf{0.615} & 0.887 & 1.346 & \underline{0.679} & \underline{0.801} & 0.689 & 0.834 & 0.686 & \uuline{0.797} & 0.764 & 1.004 & 1.308 & 2.550 & \uuline{0.677} & 0.808 & 1.545 & 3.346 & 0.841 & 1.189 & 0.776 & 1.079 \\
\addlinespace\cline{1-30} \addlinespace
\multirow[c]{4}{*}{\rotatebox{90}{ETTm2}} & 96 & \underline{0.254} & 0.165 & 0.359 & 0.263 & 0.306 & 0.219 & 0.302 & 0.216 & 0.357 & 0.298 & 0.255 & \uuline{0.164} & \textbf{0.252} & \textbf{0.163} & 0.273 & 0.175 & 0.266 & 0.190 & 0.805 & 0.989 & \uuline{0.254} & \underline{0.165} & 1.233 & 3.444 & 0.284 & 0.188 & 0.370 & 0.288 \\

 & 192 & \underline{0.292} & \underline{0.221} & 0.425 & 0.361 & 0.357 & 0.294 & 0.367 & 0.324 & 0.460 & 0.491 & 0.304 & 0.224 & \textbf{0.290} & \uuline{0.218} & 0.323 & 0.241 & 0.308 & 0.251 & 0.939 & 1.368 & \uuline{0.292} & \textbf{0.218} & 1.484 & 4.724 & 0.342 & 0.259 & 0.485 & 0.481 \\

 & 336 & \textbf{0.325} & \uuline{0.275} & 0.496 & 0.469 & 0.401 & 0.362 & 0.429 & 0.424 & 0.557 & 0.726 & 0.337 & \underline{0.277} & \uuline{0.326} & \textbf{0.273} & 0.373 & 0.312 & 0.350 & 0.322 & 1.044 & 1.635 & \underline{0.329} & 0.277 & 1.601 & 5.309 & 0.402 & 0.336 & 0.621 & 0.775 \\

 & 720 & \uuline{0.380} & \uuline{0.360} & 0.857 & 1.263 & 0.450 & 0.459 & 0.500 & 0.581 & 0.863 & 2.009 & 0.401 & 0.371 & \underline{0.382} & \underline{0.361} & 0.492 & 0.491 & 0.403 & 0.414 & 1.968 & 5.564 & \textbf{0.374} & \textbf{0.327} & 1.670 & 5.670 & 0.496 & 0.470 & 0.833 & 1.333 \\
\addlinespace\cline{1-30} \addlinespace
\multirow[c]{4}{*}{\rotatebox{90}{ETTh2}} & 96 & \uuline{0.339} & \uuline{0.277} & 0.557 & 0.611 & 0.380 & 0.338 & 0.402 & 0.378 & 0.828 & 1.894 & 0.368 & 0.302 & \textbf{0.338} & \textbf{0.275} & 0.392 & 0.339 & 0.363 & 0.319 & 1.092 & 1.706 & \underline{0.347} & \underline{0.284} & 1.444 & 4.642 & 0.411 & 0.336 & 0.437 & 0.377 \\

 & 192 & \uuline{0.381} & \uuline{0.345} & 0.651 & 0.810 & 0.428 & 0.415 & 0.449 & 0.462 & 0.908 & 2.115 & 0.433 & 0.405 & \textbf{0.379} & \textbf{0.336} & 0.434 & 0.412 & 0.416 & 0.411 & 1.701 & 4.123 & \underline{0.401} & \underline{0.358} & 1.846 & 7.099 & 0.468 & 0.414 & 0.575 & 0.611 \\

 & 336 & \uuline{0.404} & \uuline{0.368} & 0.698 & 0.928 & 0.451 & 0.378 & 0.449 & 0.426 & 0.875 & 1.816 & 0.490 & 0.496 & \textbf{0.403} & \textbf{0.362} & 0.490 & 0.497 & 0.443 & 0.415 & 1.619 & 3.371 & \underline{0.425} & \underline{0.372} & 2.007 & 8.134 & 0.521 & 0.478 & 0.738 & 0.973 \\

 & 720 & \textbf{0.432} & \uuline{0.397} & 0.775 & 1.094 & 0.485 & 0.479 & 0.449 & \underline{0.401} & 0.931 & 1.698 & 0.622 & 0.766 & \uuline{0.437} & \textbf{0.396} & 0.605 & 0.732 & \underline{0.445} & 0.429 & 1.786 & 4.166 & 0.454 & 0.427 & 2.295 & 9.981 & 0.645 & 0.666 & 0.953 & 1.585 \\
\addlinespace
\bottomrule
\end{tabular}
}
\end{table*}

\renewcommand{\arraystretch}{1}

\begin{table*}[h]
\caption{Multivariate forecasting results II.}
\label{New Multivariate forecasting results}

\resizebox{1.95\columnwidth}{!}{
\Huge
\begin{tabular}{@{}cc@{\hspace{3pt}}|@{\hspace{3pt}}cccccccccc@{\hspace{3pt}}|@{\hspace{3pt}}cccc@{\hspace{3pt}}|@{\hspace{3pt}}cccccc@{\hspace{3pt}}|@{\hspace{3pt}}cc@{\hspace{3pt}}|@{\hspace{3pt}}cc@{\hspace{3pt}}|@{\hspace{3pt}}cc@{\hspace{3pt}}|@{\hspace{3pt}}cc@{}}
\toprule
  \multicolumn{2}{c@{\hspace{3pt}}|@{\hspace{3pt}}}{\textbf{Model}} & \multicolumn{2}{c}{\textbf{PatchTST}} & \multicolumn{2}{c}{\textbf{Crossformer}} & \multicolumn{2}{c}{\textbf{FEDformer}} & \multicolumn{2}{c}{\textbf{Informer}} & \multicolumn{2}{c@{\hspace{3pt}}|@{\hspace{3pt}}}{\textbf{Triformer}} & \multicolumn{2}{c}{\textbf{DLinear}} & \multicolumn{2}{c@{\hspace{3pt}}|@{\hspace{3pt}}}{\textbf{NLinear}} & \multicolumn{2}{c}{\textbf{MICN}} & \multicolumn{2}{c}{\textbf{TimesNet}} & \multicolumn{2}{c@{\hspace{3pt}}|@{\hspace{3pt}}}{\textbf{TCN}} & \multicolumn{2}{c@{\hspace{3pt}}|@{\hspace{3pt}}}{\textbf{FiLM}} & \multicolumn{2}{c@{\hspace{3pt}}|@{\hspace{3pt}}}{\textbf{RNN}} & \multicolumn{2}{c@{\hspace{3pt}}|@{\hspace{3pt}}}{\textbf{LR}} & \multicolumn{2}{c}{\textbf{VAR}} \\
\multicolumn{2}{c@{\hspace{3pt}}|@{\hspace{3pt}}}{Metrics}  & mae & mse & mae & mse & mae & mse & mae & mse & mae & mse & mae & mse & mae  & mse & mae & mse & mae & mse & mae & mse & mae & mse & mae & mse & mae & mse & mae & mse \\

\midrule
\multirow[c]{4}{*}{\rotatebox{90}{AQShunyi}} & 96 & \textbf{0.481} & \uuline{0.648} & \uuline{0.484} & 0.652 & 0.525 & 0.706 & 0.542 & 0.754 & 0.492 & 0.665 & 0.492 & \underline{0.651} & 0.486 & 0.653 & 0.513 & 0.698 & 0.488 & 0.658 & 0.550 & 0.807 & \underline{0.486} & 0.664 & 0.741 & 1.112 & 0.489 & \textbf{0.633} & 0.517 & 0.684 \\

 & 192 & \uuline{0.501} & 0.690 & \textbf{0.499} & \uuline{0.674} & 0.531 & 0.729 & 0.536 & 0.759 & \underline{0.501} & \underline{0.681} & 0.512 & 0.691 & 0.506 & 0.701 & 0.530 & 0.722 & 0.511 & 0.707 & 0.546 & 0.760 & 0.504 & 0.705 & 0.750 & 1.147 & 0.508 & \textbf{0.673} & 0.552 & 0.746 \\

 & 336 & \uuline{0.515} & \underline{0.711} & \textbf{0.515} & \uuline{0.704} & 0.569 & 0.824 & 0.560 & 0.837 & 0.525 & 0.731 & 0.529 & 0.716 & 0.519 & 0.722 & 0.544 & 0.729 & 0.537 & 0.785 & 0.537 & 0.757 & \underline{0.517} & 0.725 & 0.754 & 1.161 & 0.522 & \textbf{0.702} & 0.578 & 0.794 \\

 & 720 & 0.538 & 0.770 & \textbf{0.518} & \uuline{0.747} & 0.561 & 0.794 & 0.543 & 0.777 & \underline{0.534} & \textbf{0.742} & 0.556 & 0.765 & 0.545 & 0.777 & 0.549 & 0.789 & \uuline{0.527} & 0.755 & 0.544 & 0.787 & 0.544 & 0.782 & 0.755 & 1.172 & 0.544 & \underline{0.749} & 0.624 & 0.877 \\
\addlinespace\cline{1-30} \addlinespace
\multirow[c]{4}{*}{\rotatebox{90}{AQWan}} & 96 & \uuline{0.470} & \uuline{0.745} & \textbf{0.465} & \underline{0.750} & 0.508 & 0.796 & 0.522 & 0.901 & \underline{0.474} & 0.762 & 0.481 & 0.756 & 0.475 & 0.758 & 0.496 & 0.789 & 0.488 & 0.791 & 0.546 & 0.874 & 0.475 & 0.766 & 0.734 & 1.197 & 0.476 & \textbf{0.730} & 0.506 & 0.775 \\

 & 192 & 0.491 & 0.792 & \textbf{0.479} & \textbf{0.762} & 0.517 & 0.825 & 0.521 & 0.833 & \uuline{0.484} & 0.786 & 0.502 & 0.800 & 0.496 & 0.809 & 0.519 & 0.821 & \underline{0.490} & \underline{0.779} & 0.521 & 0.842 & 0.494 & 0.809 & 0.743 & 1.236 & 0.496 & \uuline{0.770} & 0.543 & 0.841 \\

 & 336 & \uuline{0.503} & 0.819 & \underline{0.504} & \uuline{0.802} & 0.537 & 0.863 & 0.525 & 0.847 & \textbf{0.495} & \underline{0.802} & 0.516 & 0.823 & 0.508 & 0.830 & 0.525 & 0.826 & 0.505 & 0.814 & 0.544 & 0.900 & 0.505 & 0.831 & 0.747 & 1.251 & 0.509 & \textbf{0.801} & 0.570 & 0.893 \\

 & 720 & 0.533 & 0.890 & \textbf{0.511} & \textbf{0.830} & 0.552 & 0.907 & 0.532 & 0.883 & \underline{0.519} & \uuline{0.852} & 0.548 & 0.891 & 0.538 & 0.906 & 0.544 & 0.875 & \uuline{0.519} & 0.869 & 0.532 & 0.870 & 0.536 & 0.906 & 0.748 & 1.269 & 0.533 & \underline{0.859} & 0.618 & 0.988 \\
\addlinespace\cline{1-30} \addlinespace
\multirow[c]{4}{*}{\rotatebox{90}{NN5}} & 24 & 0.612 & 0.755 & \uuline{0.591} & \uuline{0.741} & 0.618 & 0.785 & 0.949 & 1.362 & 0.929 & 1.382 & \underline{0.593} & \underline{0.750} & 0.646 & 0.836 & \textbf{0.587} & \textbf{0.727} & 0.625 & 0.815 & 0.749 & 1.062 & 0.651 & 0.846 & 0.939 & 1.381 & 0.635 & 0.806 & 0.902 & 1.442 \\

 & 36 & 0.595 & \underline{0.694} & \underline{0.589} & 0.703 & 0.606 & 0.727 & 0.945 & 1.342 & 0.920 & 1.351 & \uuline{0.577} & \uuline{0.690} & 0.639 & 0.790 & \textbf{0.562} & \textbf{0.659} & 0.602 & 0.755 & 0.807 & 1.140 & 0.702 & 0.883 & 0.931 & 1.351 & 0.618 & 0.747 & 0.900 & 1.410 \\

 & 48 & 0.585 & \underline{0.667} & \underline{0.575} & 0.669 & 0.607 & 0.725 & 0.942 & 1.337 & 0.918 & 1.348 & \uuline{0.573} & \uuline{0.665} & 0.632 & 0.761 & \textbf{0.562} & \textbf{0.646} & 0.637 & 0.794 & 0.746 & 0.977 & 0.741 & 0.969 & 0.929 & 1.345 & 0.611 & 0.722 & 0.911 & 1.436 \\

 & 60 & 0.582 & \underline{0.653} & 0.587 & 0.683 & 0.615 & 0.726 & 0.946 & 1.334 & 0.918 & 1.346 & \underline{0.574} & 0.658 & 0.629 & 0.748 & \uuline{0.564} & \uuline{0.645} & 0.609 & 0.735 & 0.740 & 0.983 & \textbf{0.556} & \textbf{0.633} & 0.929 & 1.342 & 0.610 & 0.713 & 0.919 & 1.455 \\
\addlinespace\cline{1-30} \addlinespace
\multirow[c]{4}{*}{\rotatebox{90}{Wike2000}} & 24 & \textbf{1.023} & \textbf{457.187} & 1.707 & \underline{637.822} & 3.745 & 681.415 & 1.569 & 1219.935 & 2.028 & 641.407 & 1.446 & 697.008 & \underline{1.281} & 961.455  & 1.354 & \uuline{515.430} & \uuline{1.240} & 643.931 & 11.513 & 1485.661 & 1.318 & 959.455 & 3.137 & 653.172 & 2.056 & 856.400 & 5.624 & 2.3e+5 \\

 & 36 & \textbf{1.115} & \textbf{511.946} & 1.818 & 691.940 & 3.233 & 716.180 & 1.647 & 1291.983 & 2.080 & 694.272 & 1.613 & 773.376 & \underline{1.381} & 1040.933 & 1.523 & \underline{570.032} & \uuline{1.208} & \uuline{515.191} & 17.005 & 2150.399 & 1.419 & 983.238 & 3.186 & 706.354 & 2.105 & 887.756 & 5.710 & 1.7e+5 \\

 & 48 & \textbf{1.179} & \uuline{531.902} & 1.899 & 718.072 & 3.026 & 748.497 & 1.724 & 1342.577 & 2.130 & 719.606 & 1.623 & 786.374 & 1.577 & 1448.341 & 1.672 & \underline{614.153} & \uuline{1.239} & \textbf{517.910} & 14.732 & 1161.917 & \underline{1.513} & 1067.060 & 3.212 & 731.465 & 2.134 & 898.313 & 5.525 & 1.5e+5 \\

 & 60 & \textbf{1.244} & \textbf{554.829} & 1.920 & 730.350 & 2.953 & 786.667 & 1.800 & 1459.240 & 2.152 & 731.595 & 1.868 & 839.076 & 1.570 & 1281.245 & 1.769 & \underline{644.523} & \uuline{1.338} & \uuline{568.802} & 16.414 & 1799.562 & \underline{1.477} & 765.995 & 3.210 & 743.347 & 2.142 & 901.402 & 5.551 & 1.3e+5 \\
\addlinespace\cline{1-30} \addlinespace
\multirow[c]{4}{*}{\rotatebox{90}{Wind}} & 96 & 0.652 & 0.889 & \uuline{0.590} & \textbf{0.784} & 0.697 & 1.030 & 0.674 & 0.953 & \underline{0.612} & \underline{0.845} & 0.632 & 0.881 & 0.640 & 0.923 & 0.627 & 0.875 & 0.658 & 0.998 & 0.748 & 1.124 & 0.649 & 0.933 & 0.860 & 1.319 & \textbf{0.583} & \uuline{0.786} & 0.620 & 0.855 \\

 & 192 & 0.747 & 1.076 & \underline{0.698} & \uuline{0.965} & 0.807 & 1.275 & 0.764 & 1.147 & \uuline{0.692} & \underline{1.028} & 0.715 & 1.034 & 0.734 & 1.081 & 0.719 & 1.066 & 0.752 & 1.214 & 0.828 & 1.359 & 0.743 & 1.097 & 0.861 & 1.323 & \textbf{0.670} & \textbf{0.947} & 0.711 & 1.032 \\

 & 336 & 0.809 & 1.209 & \underline{0.755} & \uuline{1.073} & 0.864 & 1.307 & 0.825 & 1.310 & \uuline{0.749} & \underline{1.139} & 0.779 & 1.159 & 0.805 & 1.228 & 0.782 & 1.164 & 0.828 & 1.318 & 0.882 & 1.414 & 0.809 & 1.241 & 0.862 & 1.327 & \textbf{0.729} & \textbf{1.069} & 0.767 & 1.150 \\

 & 720 & 0.851 & 1.304 & \underline{0.803} & \underline{1.191} & 0.890 & 1.402 & 0.864 & 1.323 & \uuline{0.783} & \uuline{1.169} & 0.815 & 1.233 & 0.852 & 1.328 & 0.821 & 1.245 & 0.887 & 1.474 & 0.896 & 1.404 & 0.837 & 1.281 & 0.859 & 1.325 & \textbf{0.775} & \textbf{1.169} & 0.807 & 1.232 \\
\addlinespace\cline{1-30} \addlinespace
\multirow[c]{4}{*}{\rotatebox{90}{ZafNoo}} & 96 & 0.426 & 0.444 & 0.419 & \uuline{0.432} & 0.441 & 0.475 & 0.473 & 0.541 & 0.420 & 0.442 & \underline{0.411} & \underline{0.434} & \uuline{0.410} & 0.446 & 0.421 & 0.442 & 0.424 & 0.479 & 0.474 & 0.533 & 0.411 & 0.451 & 1.274 & 3.928 & \textbf{0.401} & \textbf{0.412} & 1.4e+229 & inf \\

 & 192 & 0.456 & 0.498 & 0.449 & \uuline{0.479} & 0.479 & 0.544 & 0.575 & 0.708 & 0.448 & 0.493 & \uuline{0.444} & \underline{0.484} & 0.447 & 0.503 & 0.455 & 0.493 & \underline{0.446} & 0.491 & 0.476 & 0.546 & 0.448 & 0.508 & 1.271 & 3.858 & \textbf{0.438} & \textbf{0.472} & nan & nan \\

 & 336 & 0.480 & 0.530 & 0.469 & 0.521 & 0.517 & 0.595 & 0.661 & 0.851 & 0.467 & 0.523 & \underline{0.464} & \underline{0.518} & 0.470 & 0.544 & \uuline{0.455} & \textbf{0.504} & 0.479 & 0.551 & 0.507 & 0.624 & 0.471 & 0.549 & 1.238 & 3.586 & \textbf{0.454} & \uuline{0.506} & nan & nan \\

 & 720 & 0.499 & 0.574 & \underline{0.483} & \uuline{0.543} & 0.560 & 0.697 & 0.699 & 0.876 & 0.494 & 0.563 & 0.486 & \underline{0.548} & 0.504 & 0.595 & \textbf{0.476} & \textbf{0.540} & 0.511 & 0.627 & 0.508 & 0.599 & 0.502 & 0.586 & 1.169 & 3.053 & \uuline{0.476} & 0.554 & nan & nan \\
\addlinespace\cline{1-30} \addlinespace
\multirow[c]{4}{*}{\rotatebox{90}{CzeLan}} & 96 & 0.251 & 0.183 & 0.443 & 0.581 & 0.310 & 0.231 & 0.305 & 0.250 & 0.519 & 0.818 & 0.289 & 0.211 & \textbf{0.229} & \uuline{0.178} & 0.275 & 0.194 & \underline{0.237} & \textbf{0.176} & 0.553 & 0.519 & \uuline{0.232} & \underline{0.180} & 1.312 & 3.143 & 0.262 & 0.184 & 0.351 & 0.302 \\

 & 192 & \underline{0.271} & \textbf{0.208} & 0.503 & 0.705 & 0.339 & 0.268 & 0.337 & 0.295 & 0.589 & 0.962 & 0.323 & 0.252 & \textbf{0.252} & \uuline{0.210} & 0.322 & 0.254 & 0.279 & 0.215 & 0.559 & 0.539 & \uuline{0.255} & \underline{0.212} & 1.233 & 2.963 & 0.300 & 0.222 & 0.401 & 0.368 \\

 & 336 & 0.302 & 0.243 & 0.596 & 0.971 & 0.361 & 0.298 & 0.361 & 0.335 & 0.659 & 1.161 & 0.366 & 0.317 & \textbf{0.280} & \underline{0.243} & 0.362 & 0.314 & \underline{0.288} & \textbf{0.224} & 0.692 & 0.966 & \uuline{0.281} & \uuline{0.243} & 1.243 & 2.985 & 0.357 & 0.287 & 0.454 & 0.443 \\

 & 720 & \underline{0.335} & \uuline{0.273} & 0.762 & 1.566 & 0.446 & 0.416 & 0.416 & 0.384 & 0.741 & 1.496 & 0.392 & 0.358 & \uuline{0.317} & 0.284 & 0.396 & 0.354 & 0.337 & \underline{0.282} & 0.848 & 1.143 & \textbf{0.305} & \textbf{0.268} & 1.257 & 3.020 & 0.445 & 0.405 & 0.524 & 0.544 \\
\addlinespace\cline{1-30} \addlinespace
\multirow[c]{4}{*}{\rotatebox{90}{Covid-19}} & 24 & \textbf{0.042} & \textbf{1.052} & 2.297 & 1768.432 & 0.177 & 2.125 & 0.079 & 2.642 & 2.395 & 1768.993 & 0.240 & 9.587 & 0.070 & \uuline{1.139} & 0.438 & 33.908 & \underline{0.064} & 2.231 & 55.555 & 6274.889 & \uuline{0.047} & \underline{1.183} & 3.601 & 1711.553 & 0.727 & 80.511 & 7.056 & 5.8e+4 \\

 & 36 & \textbf{0.051} & \textbf{1.398} & 2.394 & 1770.695 & 0.177 & 2.463 & 0.087 & 3.064 & 2.432 & 1769.541 & 0.173 & 5.333 & 0.091 & \underline{1.582} & 0.494 & 46.864 & \underline{0.081} & 2.530 & 72.837 & 1.1e+4 & \uuline{0.059} & \uuline{1.470} & 3.600 & 1711.515 & 0.644 & 51.959 & 10.333 & 9.3e+4 \\

 & 48 & \textbf{0.060} & \textbf{1.814} & 2.465 & 1773.221 & 0.192 & 2.847 & 0.097 & 3.698 & 2.481 & 1770.224 & 0.246 & 9.598 & 0.099 & \underline{1.932} & 0.566 & 46.563 & \underline{0.079} & 2.561 & 84.309 & 1.3e+4 & \uuline{0.069} & \uuline{1.859} & 3.600 & 1711.519 & 0.635 & 52.277 & 11.473 & 8.5e+4 \\

 & 60 & \textbf{0.068} & \textbf{2.265} & 2.427 & 1772.835 & 0.212 & 3.275 & 0.106 & 4.232 & 2.533 & 1773.691 & 0.297 & 7.778 & 0.127 & \underline{2.682} & 0.711 & 94.629 & \underline{0.098} & 3.675 & 93.938 & 1.6e+4 & \uuline{0.078} & \uuline{2.278} & 3.600 & 1711.542 & 0.737 & 102.415 & 15.914 & 9.6e+4 \\
\addlinespace\cline{1-30} \addlinespace
\multirow[c]{4}{*}{\rotatebox{90}{NASDAQ}} & 24 & 0.567 & 0.649 & 0.745 & 1.149 & 0.547 & 0.627 & 0.744 & 1.110 & 1.330 & 2.727 & 0.666 & 0.830 & \uuline{0.522} & \underline{0.566} & 0.630 & 0.776 & \underline{0.539} & \uuline{0.563} & 1.554 & 3.153 & 0.645 & 0.767 & 0.608 & 0.752 & 0.616 & 0.774 & \textbf{0.462} & \textbf{0.516} \\

 & 36 & 0.682 & \textbf{0.821} & 0.885 & 1.414 & \underline{0.659} & \underline{0.885} & 0.895 & 1.399 & 1.534 & 3.387 & 0.862 & 1.356 & \uuline{0.656} & \uuline{0.827} & 0.832 & 1.325 & 0.687 & 0.905 & 1.693 & 3.853 & 0.835 & 1.379 & 0.726 & 1.062 & 0.803 & 1.223 & \textbf{0.653} & 0.920 \\

 & 48 & 0.793 & \underline{1.169} & 1.136 & 2.108 & \underline{0.786} & \uuline{1.139} & 0.864 & 1.296 & 1.551 & 3.412 & 0.990 & 1.817 & \textbf{0.757} & \textbf{1.106} & 0.977 & 1.788 & \uuline{0.783} & 1.218 & 1.637 & 3.582 & 0.829 & 1.179 & 0.790 & 1.263 & 0.893 & 1.540 & 0.805 & 1.366 \\

 & 60 & 0.828 & 1.268 & 1.230 & 2.336 & \uuline{0.783} & \underline{1.251} & 0.834 & \textbf{1.171} & 1.528 & 3.287 & 1.037 & 2.011 & \underline{0.792} & \uuline{1.204} & 1.099 & 2.230 & \textbf{0.781} & 1.298 & 2.017 & 4.984 & 0.853 & 1.303 & 0.814 & 1.328 & 0.964 & 1.753 & 0.879 & 1.619 \\
\addlinespace\cline{1-30} \addlinespace
\multirow[c]{4}{*}{\rotatebox{90}{NYSE}} & 24 & \underline{0.296} & 0.226 & 0.848 & 0.825 & 0.297 & \uuline{0.194} & 0.353 & 0.280 & 1.258 & 2.353 & 0.335 & 0.247 & \uuline{0.283} & \textbf{0.193} & 0.593 & 0.578 & 0.341 & 0.257 & 0.548 & 0.521 & 0.364 & 0.313 & 0.560 & 0.603 & 0.422 & 0.428 & \textbf{0.279} & \underline{0.198} \\

 & 36 & \uuline{0.375} & \underline{0.365} & 0.904 & 0.942 & \underline{0.377} & \uuline{0.322} & 0.417 & 0.411 & 1.552 & 3.381 & 0.472 & 0.457 & \textbf{0.360} & \textbf{0.321} & 0.551 & 0.556 & 0.430 & 0.404 & 0.821 & 1.038 & 0.415 & 0.390 & 0.866 & 1.356 & 0.567 & 0.660 & 0.416 & 0.394 \\

 & 48 & \uuline{0.450} & \underline{0.523} & 0.962 & 1.064 & \underline{0.457} & \uuline{0.477} & 0.500 & 0.571 & 1.737 & 4.266 & 0.603 & 0.709 & \textbf{0.438} & \textbf{0.464} & 0.680 & 0.859 & 0.502 & 0.589 & 1.276 & 2.149 & 0.480 & 0.538 & 1.198 & 2.435 & 0.701 & 0.944 & 0.545 & 0.620 \\

 & 60 & 0.586 & 0.771 & 0.937 & 1.121 & \uuline{0.557} & \uuline{0.636} & 0.615 & 0.782 & 1.847 & 4.701 & 0.696 & 0.914 & \textbf{0.522} & \textbf{0.631} & 0.731 & 0.957 & 0.629 & 0.785 & 1.354 & 2.426 & \underline{0.563} & \underline{0.721} & 1.531 & 3.767 & 0.848 & 1.319 & 0.673 & 0.894 \\
\addlinespace\cline{1-30} \addlinespace
\multirow[c]{4}{*}{\rotatebox{90}{FRED-MD}} & 24 & \uuline{1.014} & \uuline{35.777} & 3.618 & 380.703 & 1.623 & 66.090 & 1.559 & 69.980 & 5.046 & 395.884 & \underline{1.070} & \underline{37.898} & \textbf{0.931} & \textbf{32.125} & 1.521 & 63.217 & 1.266 & 43.268 & 21.307 & 1440.311 & 1.155 & 40.680 & 8.090 & 482.839 & 7.689 & 439.730 & 44.514 & 8076.984 \\

 & 36 & \uuline{1.345} & \uuline{61.034} & 3.960 & 396.700 & 1.863 & 94.359 & 1.867 & 102.875 & 5.175 & 411.328 & \underline{1.477} & 71.047 & \textbf{1.258} & \textbf{58.332} & 1.945 & 102.800 & 1.533 & \underline{69.514} & 7.896 & 467.865 & 1.670 & 90.434 & 8.178 & 497.868 & 7.024 & 442.815 & 45.036 & 8561.171 \\

 & 48 & \uuline{1.667} & \underline{93.482} & 4.110 & 412.674 & 2.135 & 129.798 & 2.147 & 138.049 & 5.234 & 423.926 & 2.002 & 118.579 & \textbf{1.609} & \textbf{82.184} & 2.373 & 147.990 & \underline{1.742} & \uuline{89.913} & 10.404 & 613.432 & 1.943 & 109.945 & 8.247 & 511.313 & 6.471 & 446.561 & 42.856 & 7960.835 \\

 & 60 & \underline{2.011} & \underline{133.444} & 4.189 & 420.936 & 2.435 & 173.616 & 2.461 & 185.188 & 5.299 & 432.464 & 2.221 & 156.844 & \textbf{1.882} & \textbf{109.625} & 2.781 & 201.248 & \uuline{1.976} & \uuline{116.187} & 6.885 & 446.670 & 2.397 & 180.367 & 8.288 & 519.078 & 5.911 & 444.686 & 42.948 & 8564.820 \\
\addlinespace
\bottomrule
\end{tabular}
}
\end{table*}

\begin{figure}[t]
\centering
\begin{subfigure}{0.4\linewidth}
    \includegraphics[width=1.0\textwidth]{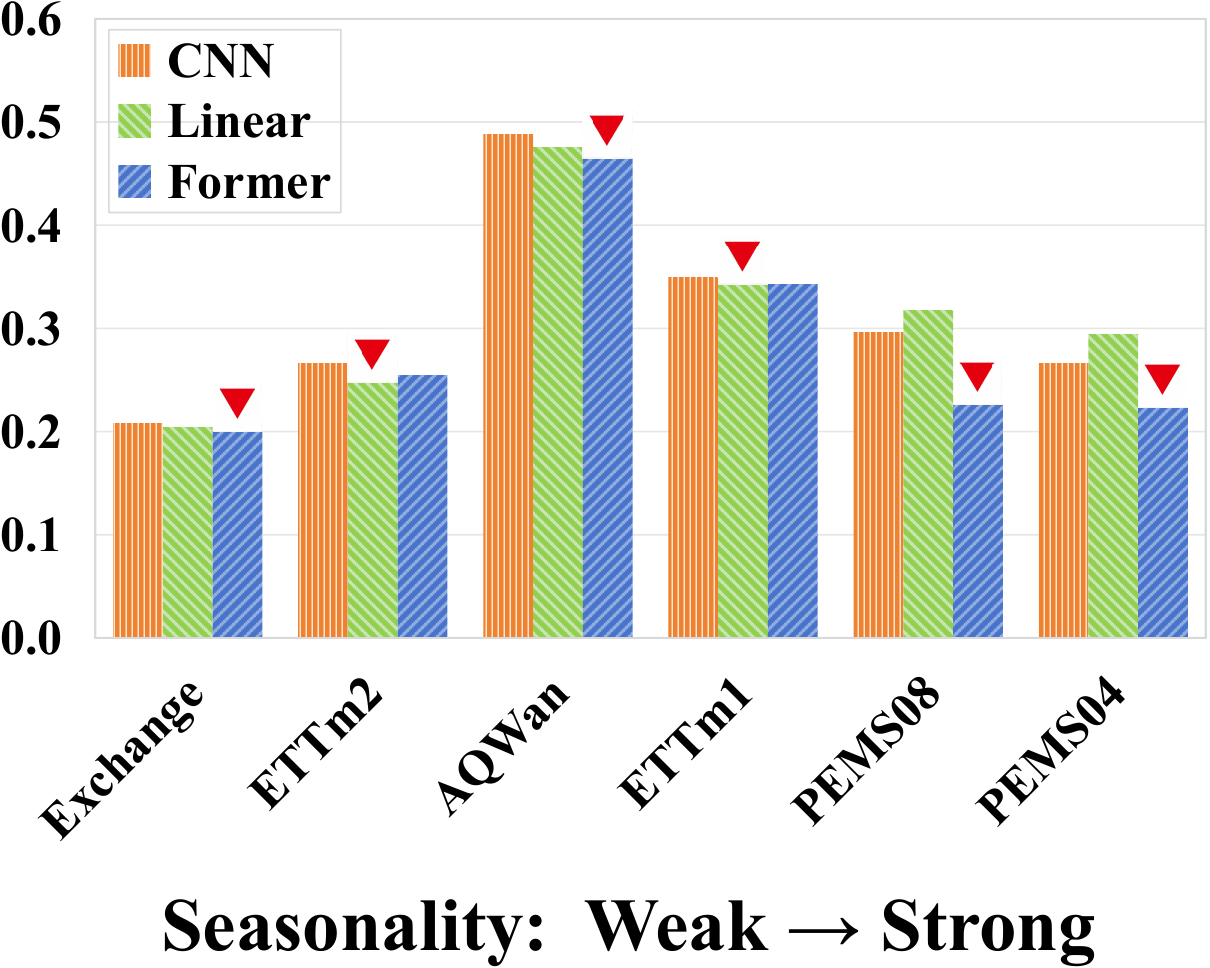}
    \caption{Seasonality}
    \label{fig:bar_seasonal}
\end{subfigure}
\begin{subfigure}{0.4\linewidth}
    \includegraphics[width=1.0\textwidth]{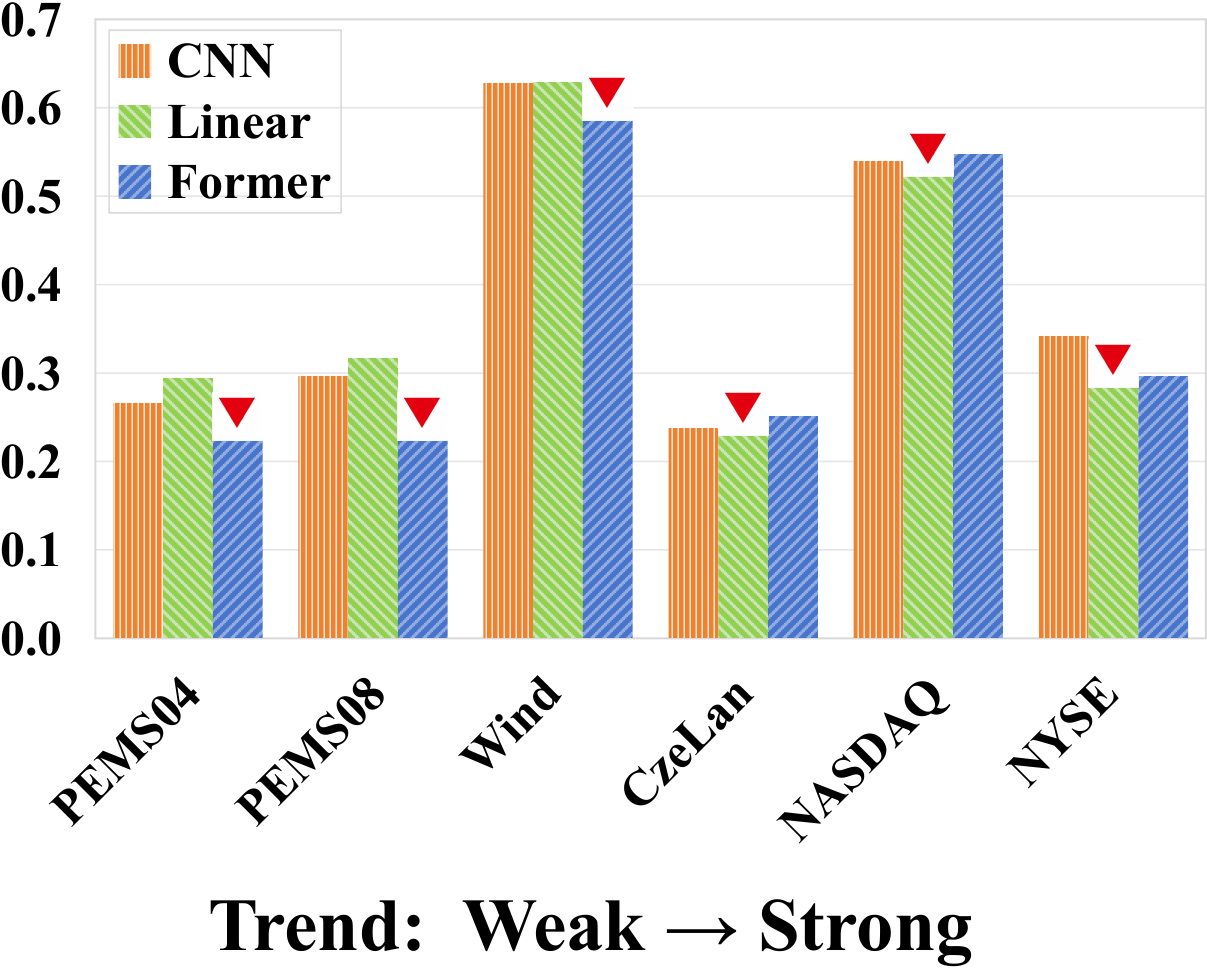}
    \caption{Trend}
    \label{fig:bar_trend}
\end{subfigure}
\begin{subfigure}{0.4\linewidth}
    \includegraphics[width=1.0\textwidth]{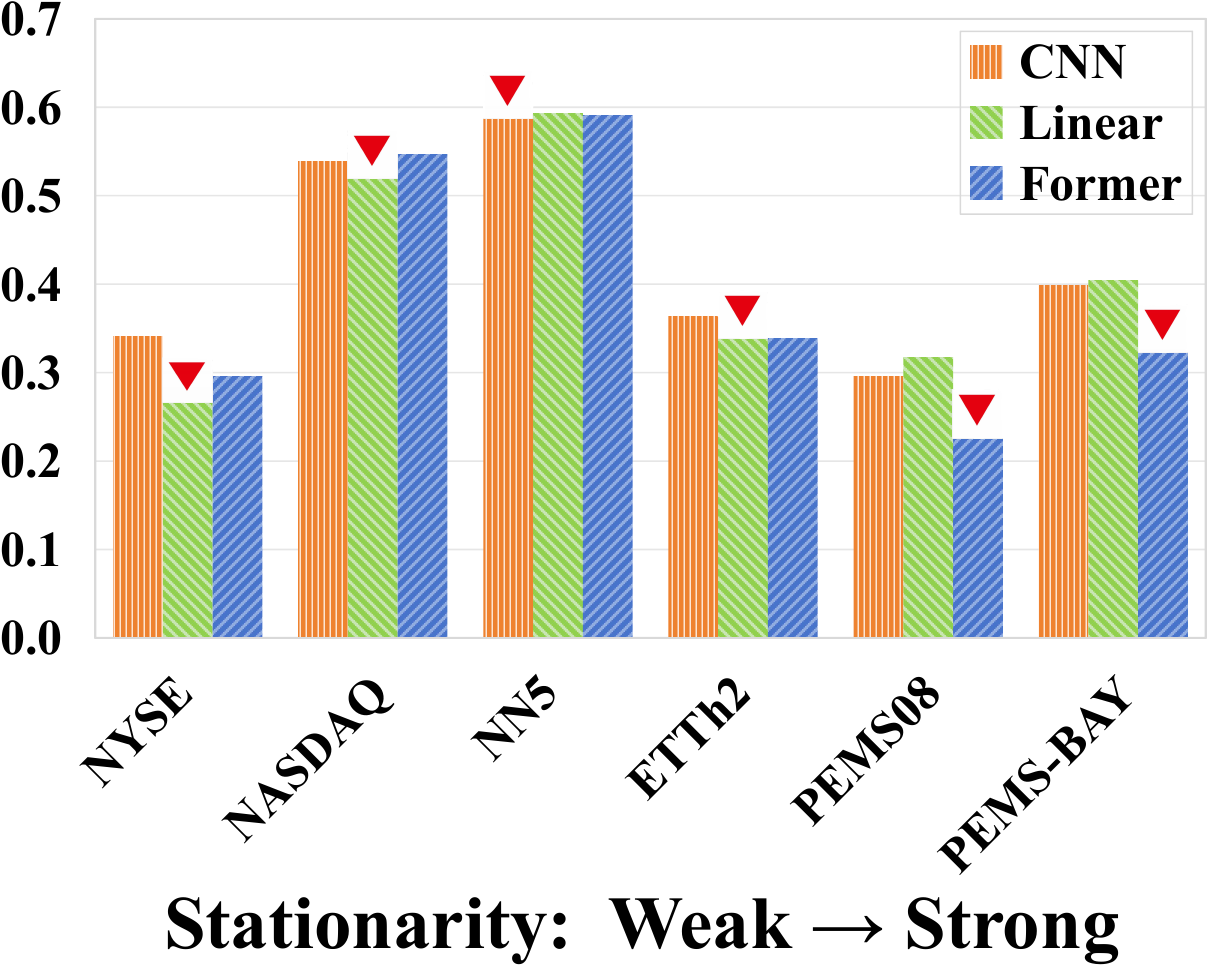}
    \caption{Stationarity}
    \label{fig:bar_stationary}
\end{subfigure}
\begin{subfigure}{0.4\linewidth}
    \includegraphics[width=1.0\textwidth]{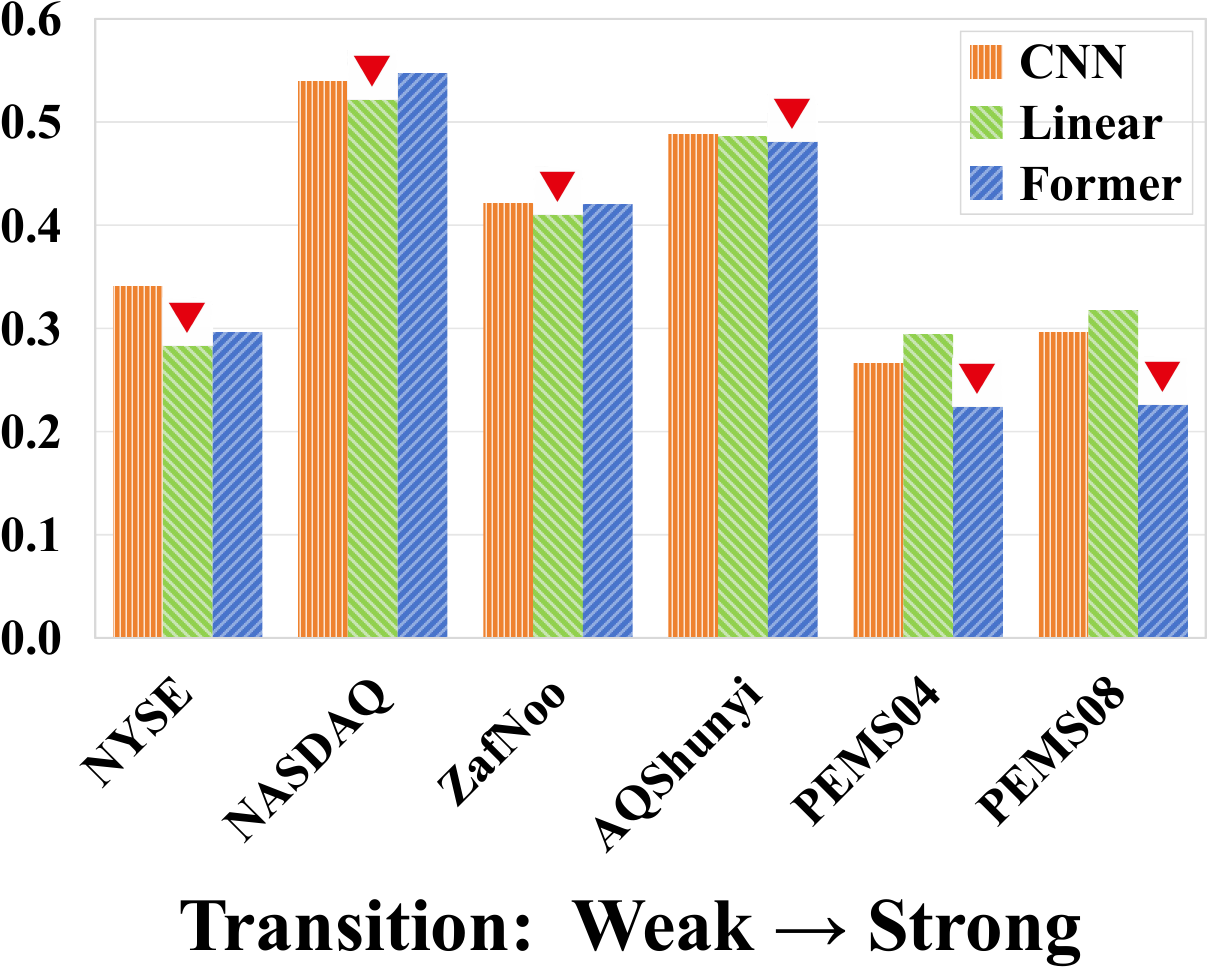}
    \caption{Transition}
    \label{fig:bar_pattern}
\end{subfigure}
\begin{subfigure}{0.4\linewidth}
    \includegraphics[width=1.0\textwidth]{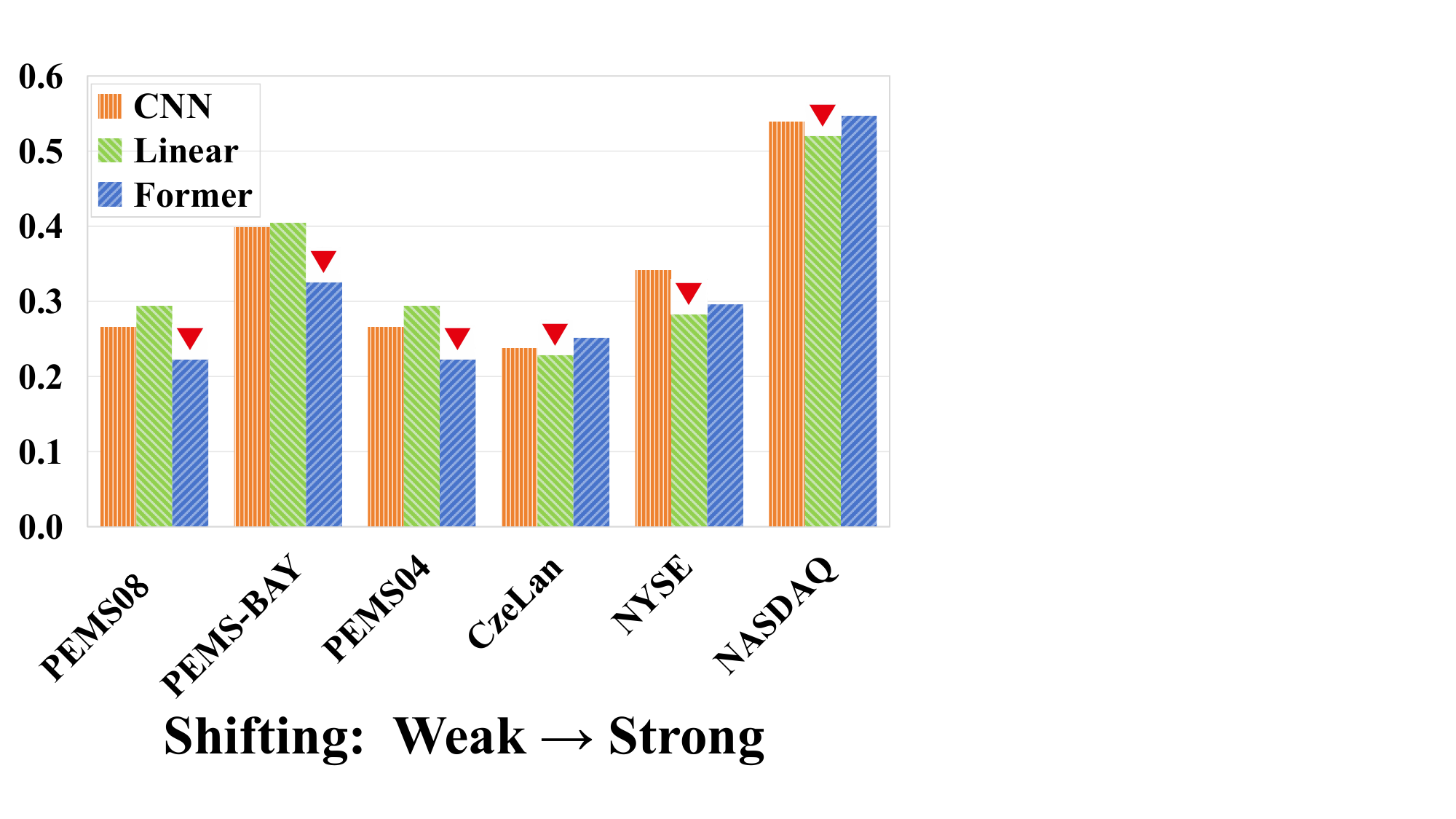}
    \caption{Shifting}
    \label{fig:bar_shifting}
\end{subfigure}
\begin{subfigure}{0.4\linewidth}
    \includegraphics[width=1.0\textwidth]{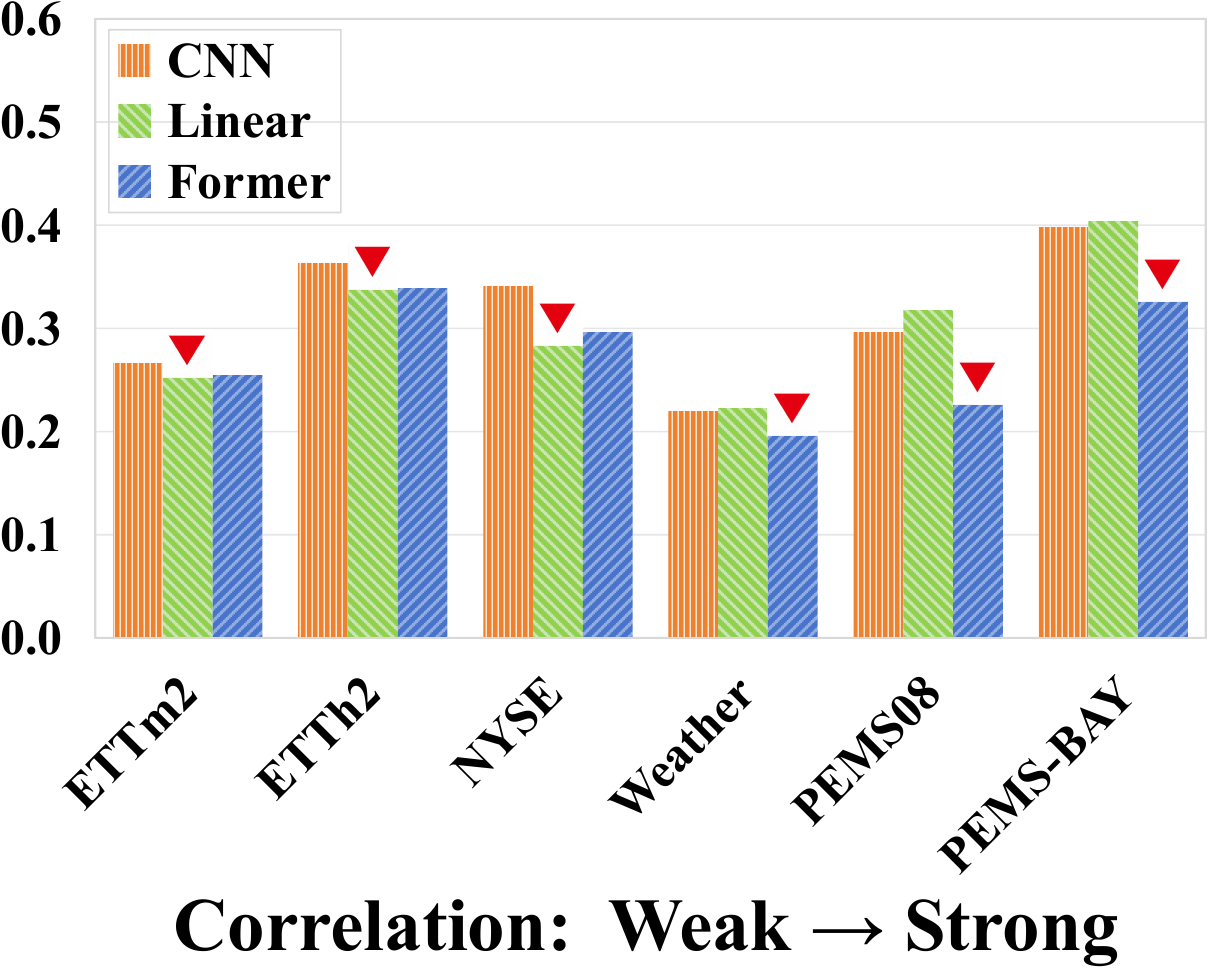}
    \caption{Correlation}
    \label{fig:bar_similarity}
\end{subfigure}
\caption{Comparison between Transformer-based, CNN-based, and Linear-based methods. Red triangles indicate methods with the best accuracy (minimum MAE).}
\label{fig:Comparison algorithms}
\end{figure}

\subsection{Hints to Method Design}
\subsubsection{Transformers vs. linear methods}

To study the impact of different data characteristics on these two types of methods, we consider the best MAE results obtained by CNN, Linear and Transformer. We use a forecasting horizon of 24 for NASDAQ, NYSE, and NN5 and of 96 for other datasets---see Figure~\ref{fig:Comparison algorithms}. 
We choose CNN as a reference to gain a more comprehensive understanding of the performance of Transformer and Linear methods. We have the following observations. 
First, each method exhibits distinct advantages on datasets with different characteristics. Second, Linear-based methods excel when the dataset shows an increasing trend or significant shifts. This can be attributed to the linear modeling capabilities of the Linear model, making it well-suited for capturing linear trends and shifts. Third, Transformer-based methods outperform Linear-based methods on datasets exhibiting marked seasonality, stationarity, and nonlinear patterns, as well as more pronounced patterns or strong internal similarities. This superiority may stem from the enhanced nonlinear modeling capabilities of Transformer-based methods, enabling them to flexibly adapt to complex time series patterns and intrinsic correlations. 

Therefore, the observed phenomenon underscores the inherent differences between Transformer-based and Linear-based methods at addressing diverse characteristics of time series. To achieve optimal performance, we recommend selecting the appropriate method based on the characteristics of the relevant time series, allowing the method to fully leverage its strengths.

\begin{figure}[t]
    \centering
    \includegraphics[width=0.75\linewidth]{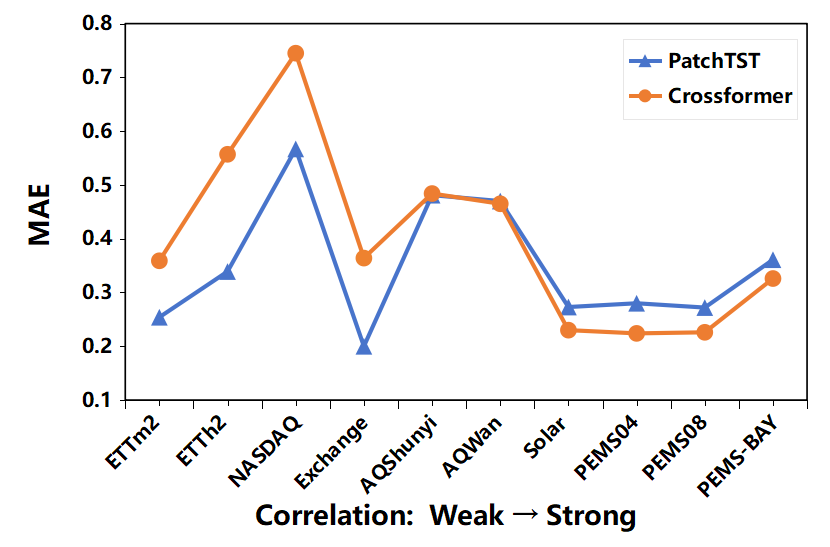}
    \caption{Method performance for varying correlation within datasets.}
    \label{fig:multi_linear_map}
\end{figure}
\subsubsection{Channel independence  vs. channel dependence}
In multivariate datasets, variables are occasionally referred to as channels. To study the impact of channel dependency in multivariate time series, we compare PatchTST and Crossformer on ten datasets with dependencies ranging from weak to strong. We report the MAE for forecasting horizon $F$ is 96 in Figure~\ref{fig:multi_linear_map}. We observe that as the correlations within a dataset increase, the performance of Crossformer gradually surpasses that of PatchTST, suggesting that it is better to consider channel dependencies when correlations are strong. However, when correlations among variables are not pronounced, PatchTST that does not consider channel dependencies is better. 

This observation indicates that introducing consideration for dependencies between channels can significantly improve the performance of multivariate time series prediction, especially on datasets with strong correlations, when compared to methods that assume channel independence. This suggests that when designing new forecasting methods, attention should be given to leveraging and fully utilizing the relationships between variables, thereby capturing more accurately the underlying structures and patterns in the dataset. However, when the correlation within a dataset is not pronounced, the performance of the Crossformer method, considering channel dependencies, may not outperform PatchTST. Therefore, it is of interest to find better ways to effectively leverage channel dependencies. This in-depth exploration and balancing of channel dependencies provides guidance for the design of new methods and the performance optimization of existing ones.

\begin{figure}[t]
    \centering
    \includegraphics[width=0.85\linewidth, trim=0 20 0 10, clip]{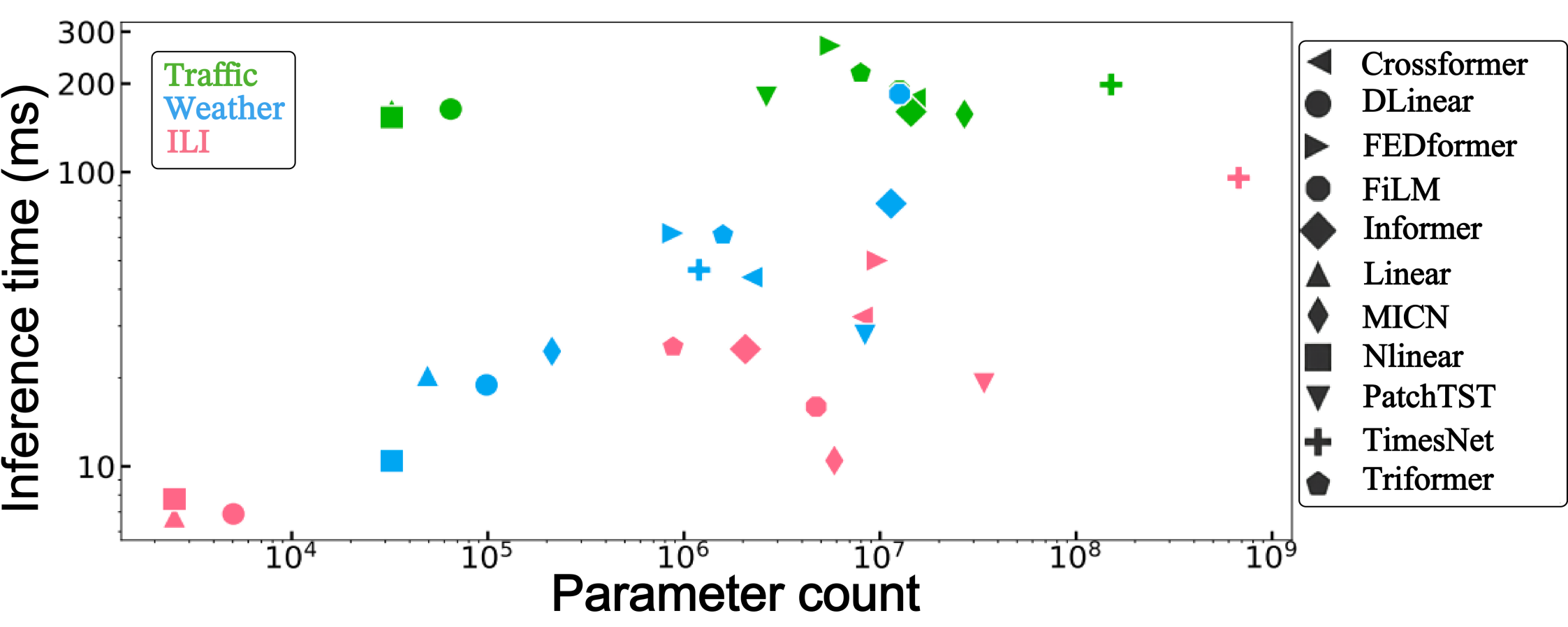}
    \caption{Comparison of parameter counts and inference time for deep learning methods.}
    \label{fig:parameter}
\end{figure}

\subsubsection{Running time and parameter}
\label{Runtime and Memory}
We study the performance of deep learning methods in MTSF about inference time and parameter counts. Due to space constraints, we choose three datasets with different scales, namely Traffic~(large), Weather~(medium), and ILI~(small). We use the forecasting horizon of 24 for ILI, while it is 96 for Traffic and Weather. The results are shown in Figure~\ref{fig:parameter}, where the horizontal axis represents the number of parameters on a logarithmic scale and the vertical axis represents the inference time per sample~(window in the test dataset) in milliseconds, which is also on a logarithmic scale. The reported parameter counts and inference times refer to the method's best performance on the dataset. We observe a general upward trend in the inference time as the number of parameters increases, which is intuitive. When considering the running time and number of parameters, Linear-based methods outperform CNN-based and Transformer-based methods. Further, CNN-based methods tend to have a larger number of parameters. Among the Transformer-based methods, we can also observe that PatchTST demonstrates a notable advantage over Triformer and Crossformer in terms of running time.

\section{Conclusions}
\label{Conclusions}
We present the \textbf{TFB} time series forecasting benchmark that addresses three issues to enable comprehensive and reliable comparison of TSF methods. \textit{To alleviate the insufficient data domains problem}, TFB includes datasets from a total of 10 different domains, covering traffic, electricity, energy, environment, nature, economics, stock, banking, health, and web. We also include a time series characteristics analysis to ensure that selected datasets are well distributed across different characteristics. \textit{To remove the bias on traditional methods,} TFB covers a diverse range of methods, including statistical learning, machine learning, and deep learning methods, accompanied by a variety of evaluation strategies and metrics. Thus, TFB can evaluate the performance of different methods comprehensively. \textit{To address the inconsistent and inflexible pipelines problem,} TFB offers a new flexible and scalable pipeline, eliminating biases and providing a better basis for performance comparison. Overall, TFB is a more comprehensive and fair benchmark that aims to promote the development of new TSF methods.

\begin{acks}
This work was partially supported by National Natural Science Foundation of China (62372179), Independent Research Fund Denmark under agreement 8048-00038B, and Huawei Cloud Algorithm Innovation Lab. Jilin Hu is the corresponding author of the work.
\end{acks}


\bibliographystyle{ACM-Reference-Format}
\bibliography{sample}


\begin{thebibliography}{99}


\ifx \showCODEN    \undefined \def \showCODEN     #1{\unskip}     \fi
\ifx \showDOI      \undefined \def \showDOI       #1{#1}\fi
\ifx \showISBNx    \undefined \def \showISBNx     #1{\unskip}     \fi
\ifx \showISBNxiii \undefined \def \showISBNxiii  #1{\unskip}     \fi
\ifx \showISSN     \undefined \def \showISSN      #1{\unskip}     \fi
\ifx \showLCCN     \undefined \def \showLCCN      #1{\unskip}     \fi
\ifx \shownote     \undefined \def \shownote      #1{#1}          \fi
\ifx \showarticletitle \undefined \def \showarticletitle #1{#1}   \fi
\ifx \showURL      \undefined \def \showURL       {\relax}        \fi
\providecommand\bibfield[2]{#2}
\providecommand\bibinfo[2]{#2}
\providecommand\natexlab[1]{#1}
\providecommand\showeprint[2][]{arXiv:#2}

\bibitem[\protect\citeauthoryear{Alvarez, Troncoso, Riquelme, and Ruiz}{Alvarez et~al\mbox{.}}{2010}]%
        {alvarez2010energy}
\bibfield{author}{\bibinfo{person}{Francisco~Martinez Alvarez}, \bibinfo{person}{Alicia Troncoso}, \bibinfo{person}{Jose~C Riquelme}, {and} \bibinfo{person}{Jesus S~Aguilar Ruiz}.} \bibinfo{year}{2010}\natexlab{}.
\newblock \showarticletitle{Energy time series forecasting based on pattern sequence similarity}.
\newblock \bibinfo{journal}{\emph{IEEE Transactions on Knowledge and Data Engineering}} \bibinfo{volume}{23}, \bibinfo{number}{8} (\bibinfo{year}{2010}), \bibinfo{pages}{1230--1243}.
\newblock


\bibitem[\protect\citeauthoryear{Bai, Kolter, and Koltun}{Bai et~al\mbox{.}}{2018}]%
        {bai2018empirical}
\bibfield{author}{\bibinfo{person}{Shaojie Bai}, \bibinfo{person}{J~Zico Kolter}, {and} \bibinfo{person}{Vladlen Koltun}.} \bibinfo{year}{2018}\natexlab{}.
\newblock \showarticletitle{An empirical evaluation of generic convolutional and recurrent networks for sequence modeling}.
\newblock \bibinfo{journal}{\emph{arXiv preprint arXiv:1803.01271}} (\bibinfo{year}{2018}).
\newblock


\bibitem[\protect\citeauthoryear{Bauer, Z{\"u}fle, Eismann, Grohmann, Herbst, and Kounev}{Bauer et~al\mbox{.}}{2021}]%
        {bauer2021libra}
\bibfield{author}{\bibinfo{person}{Andr{\'e} Bauer}, \bibinfo{person}{Marwin Z{\"u}fle}, \bibinfo{person}{Simon Eismann}, \bibinfo{person}{Johannes Grohmann}, \bibinfo{person}{Nikolas Herbst}, {and} \bibinfo{person}{Samuel Kounev}.} \bibinfo{year}{2021}\natexlab{}.
\newblock \showarticletitle{Libra: A benchmark for time series forecasting methods}. In \bibinfo{booktitle}{\emph{{ICPE}}}. \bibinfo{pages}{189--200}.
\newblock


\bibitem[\protect\citeauthoryear{Box and Pierce}{Box and Pierce}{1970}]%
        {box1970distribution}
\bibfield{author}{\bibinfo{person}{George~EP Box} {and} \bibinfo{person}{David~A Pierce}.} \bibinfo{year}{1970}\natexlab{}.
\newblock \showarticletitle{Distribution of residual autocorrelations in autoregressive-integrated moving average time series models}.
\newblock \bibinfo{journal}{\emph{Journal of the American statistical Association}} \bibinfo{volume}{65}, \bibinfo{number}{332} (\bibinfo{year}{1970}), \bibinfo{pages}{1509--1526}.
\newblock


\bibitem[\protect\citeauthoryear{Breiman}{Breiman}{2001}]%
        {breiman2001random}
\bibfield{author}{\bibinfo{person}{Leo Breiman}.} \bibinfo{year}{2001}\natexlab{}.
\newblock \showarticletitle{Random forests}.
\newblock \bibinfo{journal}{\emph{Machine learning}}  \bibinfo{volume}{45} (\bibinfo{year}{2001}), \bibinfo{pages}{5--32}.
\newblock


\bibitem[\protect\citeauthoryear{Bro and Smilde}{Bro and Smilde}{2014}]%
        {bro2014principal}
\bibfield{author}{\bibinfo{person}{Rasmus Bro} {and} \bibinfo{person}{Age~K Smilde}.} \bibinfo{year}{2014}\natexlab{}.
\newblock \showarticletitle{Principal component analysis}.
\newblock \bibinfo{journal}{\emph{Analytical methods}} \bibinfo{volume}{6}, \bibinfo{number}{9} (\bibinfo{year}{2014}), \bibinfo{pages}{2812--2831}.
\newblock


\bibitem[\protect\citeauthoryear{Campos, Kieu, Guo, Huang, Zheng, Yang, and Jensen}{Campos et~al\mbox{.}}{2022}]%
        {davidpvldb}
\bibfield{author}{\bibinfo{person}{David Campos}, \bibinfo{person}{Tung Kieu}, \bibinfo{person}{Chenjuan Guo}, \bibinfo{person}{Feiteng Huang}, \bibinfo{person}{Kai Zheng}, \bibinfo{person}{Bin Yang}, {and} \bibinfo{person}{Christian~S. Jensen}.} \bibinfo{year}{2022}\natexlab{}.
\newblock \showarticletitle{Unsupervised Time Series Outlier Detection with Diversity-Driven Convolutional Ensembles}.
\newblock \bibinfo{journal}{\emph{Proc. {VLDB} Endow.}} \bibinfo{volume}{15}, \bibinfo{number}{3} (\bibinfo{year}{2022}), \bibinfo{pages}{611--623}.
\newblock


\bibitem[\protect\citeauthoryear{Campos, Yang, Kieu, Zhang, Guo, and Jensen}{Campos et~al\mbox{.}}{2024}]%
        {campos2024qcore}
\bibfield{author}{\bibinfo{person}{David Campos}, \bibinfo{person}{Bin Yang}, \bibinfo{person}{Tung Kieu}, \bibinfo{person}{Miao Zhang}, \bibinfo{person}{Chenjuan Guo}, {and} \bibinfo{person}{Christian~S Jensen}.} \bibinfo{year}{2024}\natexlab{}.
\newblock \showarticletitle{QCore: Data-Efficient, On-Device Continual Calibration for Quantized Models--Extended Version}.
\newblock \bibinfo{journal}{\emph{arXiv preprint arXiv:2404.13990}} (\bibinfo{year}{2024}).
\newblock


\bibitem[\protect\citeauthoryear{Challu, Olivares, Oreshkin, Ramirez, Canseco, and Dubrawski}{Challu et~al\mbox{.}}{2023}]%
        {challu2023nhits}
\bibfield{author}{\bibinfo{person}{Cristian Challu}, \bibinfo{person}{Kin~G Olivares}, \bibinfo{person}{Boris~N Oreshkin}, \bibinfo{person}{Federico~Garza Ramirez}, \bibinfo{person}{Max~Mergenthaler Canseco}, {and} \bibinfo{person}{Artur Dubrawski}.} \bibinfo{year}{2023}\natexlab{}.
\newblock \showarticletitle{Nhits: Neural hierarchical interpolation for time series forecasting}. In \bibinfo{booktitle}{\emph{{AAAI}}}, Vol.~\bibinfo{volume}{37}. \bibinfo{pages}{6989--6997}.
\newblock


\bibitem[\protect\citeauthoryear{Chen, Zhang, Cheng, Shu, Wang, Wen, Yang, and Guo}{Chen et~al\mbox{.}}{2024}]%
        {chen2024pathformer}
\bibfield{author}{\bibinfo{person}{Peng Chen}, \bibinfo{person}{Yingying Zhang}, \bibinfo{person}{Yunyao Cheng}, \bibinfo{person}{Yang Shu}, \bibinfo{person}{Yihang Wang}, \bibinfo{person}{Qingsong Wen}, \bibinfo{person}{Bin Yang}, {and} \bibinfo{person}{Chenjuan Guo}.} \bibinfo{year}{2024}\natexlab{}.
\newblock \showarticletitle{Pathformer: Multi-scale transformers with Adaptive Pathways for Time Series Forecasting}.
\newblock \bibinfo{journal}{\emph{arXiv preprint arXiv:2402.05956}} (\bibinfo{year}{2024}).
\newblock


\bibitem[\protect\citeauthoryear{Chen and Guestrin}{Chen and Guestrin}{2016}]%
        {chen2016xgboost}
\bibfield{author}{\bibinfo{person}{Tianqi Chen} {and} \bibinfo{person}{Carlos Guestrin}.} \bibinfo{year}{2016}\natexlab{}.
\newblock \showarticletitle{Xgboost: A scalable tree boosting system}. In \bibinfo{booktitle}{\emph{{SIGKDD}}}. \bibinfo{pages}{785--794}.
\newblock


\bibitem[\protect\citeauthoryear{Cheng, Chen, Guo, Zhao, Wen, Yang, and Jensen}{Cheng et~al\mbox{.}}{2023}]%
        {cheng2023weakly}
\bibfield{author}{\bibinfo{person}{Yunyao Cheng}, \bibinfo{person}{Peng Chen}, \bibinfo{person}{Chenjuan Guo}, \bibinfo{person}{Kai Zhao}, \bibinfo{person}{Qingsong Wen}, \bibinfo{person}{Bin Yang}, {and} \bibinfo{person}{Christian~S Jensen}.} \bibinfo{year}{2023}\natexlab{}.
\newblock \showarticletitle{Weakly guided adaptation for robust time series forecasting}.
\newblock \bibinfo{journal}{\emph{Proc. {VLDB} Endow.}} \bibinfo{volume}{17}, \bibinfo{number}{4} (\bibinfo{year}{2023}), \bibinfo{pages}{766--779}.
\newblock


\bibitem[\protect\citeauthoryear{Cirstea, Guo, Yang, Kieu, Dong, and Pan}{Cirstea et~al\mbox{.}}{2022a}]%
        {cirstea2022triformer}
\bibfield{author}{\bibinfo{person}{Razvan{-}Gabriel Cirstea}, \bibinfo{person}{Chenjuan Guo}, \bibinfo{person}{Bin Yang}, \bibinfo{person}{Tung Kieu}, \bibinfo{person}{Xuanyi Dong}, {and} \bibinfo{person}{Shirui Pan}.} \bibinfo{year}{2022}\natexlab{a}.
\newblock \showarticletitle{Triformer: Triangular, Variable-Specific Attentions for Long Sequence Multivariate Time Series Forecasting}. In \bibinfo{booktitle}{\emph{{IJCAI}}}. \bibinfo{pages}{1994--2001}.
\newblock


\bibitem[\protect\citeauthoryear{Cirstea, Yang, Guo, Kieu, and Pan}{Cirstea et~al\mbox{.}}{2022b}]%
        {DBLP:conf/icde/CirsteaYGKP22}
\bibfield{author}{\bibinfo{person}{Razvan{-}Gabriel Cirstea}, \bibinfo{person}{Bin Yang}, \bibinfo{person}{Chenjuan Guo}, \bibinfo{person}{Tung Kieu}, {and} \bibinfo{person}{Shirui Pan}.} \bibinfo{year}{2022}\natexlab{b}.
\newblock \showarticletitle{Towards Spatio-Temporal Aware Traffic Time Series Forecasting}. In \bibinfo{booktitle}{\emph{{ICDE}}}. \bibinfo{pages}{2900--2913}.
\newblock


\bibitem[\protect\citeauthoryear{Cirstea, Kieu, Guo, Yang, and Pan}{Cirstea et~al\mbox{.}}{2021}]%
        {razvanicde2021}
\bibfield{author}{\bibinfo{person}{Razvan-Gabriel Cirstea}, \bibinfo{person}{Tung Kieu}, \bibinfo{person}{Chenjuan Guo}, \bibinfo{person}{Bin Yang}, {and} \bibinfo{person}{Sinno~Jialin Pan}.} \bibinfo{year}{2021}\natexlab{}.
\newblock \showarticletitle{Enhance{N}et: Plugin Neural Networks for Enhancing Correlated Time Series Forecasting}. In \bibinfo{booktitle}{\emph{{ICDE}}}. \bibinfo{pages}{1739--1750}.
\newblock


\bibitem[\protect\citeauthoryear{Cirstea, Yang, and Guo}{Cirstea et~al\mbox{.}}{2019}]%
        {MileTS}
\bibfield{author}{\bibinfo{person}{Razvan-Gabriel Cirstea}, \bibinfo{person}{Bin Yang}, {and} \bibinfo{person}{Chenjuan Guo}.} \bibinfo{year}{2019}\natexlab{}.
\newblock \showarticletitle{Graph Attention Recurrent Neural Networks for Correlated Time Series Forecasting}. In \bibinfo{booktitle}{\emph{{MileTS19@KDD}}}.
\newblock


\bibitem[\protect\citeauthoryear{Cleveland, Cleveland, McRae, and Terpenning}{Cleveland et~al\mbox{.}}{1990}]%
        {cleveland1990stl}
\bibfield{author}{\bibinfo{person}{Robert~B Cleveland}, \bibinfo{person}{William~S Cleveland}, \bibinfo{person}{Jean~E McRae}, {and} \bibinfo{person}{Irma Terpenning}.} \bibinfo{year}{1990}\natexlab{}.
\newblock \showarticletitle{STL: A seasonal-trend decomposition}.
\newblock \bibinfo{journal}{\emph{J. Off. Stat}} \bibinfo{volume}{6}, \bibinfo{number}{1} (\bibinfo{year}{1990}), \bibinfo{pages}{3--73}.
\newblock


\bibitem[\protect\citeauthoryear{Cohen, Huang, Chen, Benesty, Benesty, Chen, Huang, and Cohen}{Cohen et~al\mbox{.}}{2009}]%
        {cohen2009pearson}
\bibfield{author}{\bibinfo{person}{Israel Cohen}, \bibinfo{person}{Yiteng Huang}, \bibinfo{person}{Jingdong Chen}, \bibinfo{person}{Jacob Benesty}, \bibinfo{person}{Jacob Benesty}, \bibinfo{person}{Jingdong Chen}, \bibinfo{person}{Yiteng Huang}, {and} \bibinfo{person}{Israel Cohen}.} \bibinfo{year}{2009}\natexlab{}.
\newblock \showarticletitle{Pearson correlation coefficient}.
\newblock \bibinfo{journal}{\emph{Noise reduction in speech processing}} (\bibinfo{year}{2009}), \bibinfo{pages}{1--4}.
\newblock


\bibitem[\protect\citeauthoryear{Das, Kong, Leach, Sen, and Yu}{Das et~al\mbox{.}}{2023}]%
        {das2023long}
\bibfield{author}{\bibinfo{person}{Abhimanyu Das}, \bibinfo{person}{Weihao Kong}, \bibinfo{person}{Andrew Leach}, \bibinfo{person}{Rajat Sen}, {and} \bibinfo{person}{Rose Yu}.} \bibinfo{year}{2023}\natexlab{}.
\newblock \showarticletitle{Long-term Forecasting with TiDE: Time-series Dense Encoder}.
\newblock \bibinfo{journal}{\emph{arXiv preprint arXiv:2304.08424}} (\bibinfo{year}{2023}).
\newblock


\bibitem[\protect\citeauthoryear{Deng, Dong, Socher, Li, Li, and Fei-Fei}{Deng et~al\mbox{.}}{2009}]%
        {deng2009imagenet}
\bibfield{author}{\bibinfo{person}{Jia Deng}, \bibinfo{person}{Wei Dong}, \bibinfo{person}{Richard Socher}, \bibinfo{person}{Li-Jia Li}, \bibinfo{person}{Kai Li}, {and} \bibinfo{person}{Li Fei-Fei}.} \bibinfo{year}{2009}\natexlab{}.
\newblock \showarticletitle{Imagenet: A large-scale hierarchical image database}. In \bibinfo{booktitle}{\emph{CVPR}}. \bibinfo{pages}{248--255}.
\newblock


\bibitem[\protect\citeauthoryear{Elliott, Rothenberg, and Stock}{Elliott et~al\mbox{.}}{1992}]%
        {elliott1992efficient}
\bibfield{author}{\bibinfo{person}{Graham Elliott}, \bibinfo{person}{Thomas~J Rothenberg}, {and} \bibinfo{person}{James~H Stock}.} \bibinfo{year}{1992}\natexlab{}.
\newblock \bibinfo{title}{Efficient tests for an autoregressive unit root}.
\newblock
\newblock


\bibitem[\protect\citeauthoryear{Federico~Garza}{Federico~Garza}{2022}]%
        {garza2022statsforecast}
\bibfield{author}{\bibinfo{person}{Cristian Challú Kin G.~Olivares Federico~Garza, Max Mergenthaler~Canseco}.} \bibinfo{year}{2022}\natexlab{}.
\newblock \bibinfo{title}{{StatsForecast}: Lightning fast forecasting with statistical and econometric models}.
\newblock \bibinfo{howpublished}{{PyCon} Salt Lake City, Utah, US 2022}.
\newblock


\bibitem[\protect\citeauthoryear{Feng, He, Wang, Luo, Liu, and Chua}{Feng et~al\mbox{.}}{2019}]%
        {feng2019temporal}
\bibfield{author}{\bibinfo{person}{Fuli Feng}, \bibinfo{person}{Xiangnan He}, \bibinfo{person}{Xiang Wang}, \bibinfo{person}{Cheng Luo}, \bibinfo{person}{Yiqun Liu}, {and} \bibinfo{person}{Tat-Seng Chua}.} \bibinfo{year}{2019}\natexlab{}.
\newblock \showarticletitle{Temporal relational ranking for stock prediction}.
\newblock \bibinfo{journal}{\emph{ACM Transactions on Information Systems}} \bibinfo{volume}{37}, \bibinfo{number}{2} (\bibinfo{year}{2019}), \bibinfo{pages}{1--30}.
\newblock


\bibitem[\protect\citeauthoryear{Fischer, Pohl, and Ratz}{Fischer et~al\mbox{.}}{2020}]%
        {fischer2020machine}
\bibfield{author}{\bibinfo{person}{Jan~Alexander Fischer}, \bibinfo{person}{Philipp Pohl}, {and} \bibinfo{person}{Dietmar Ratz}.} \bibinfo{year}{2020}\natexlab{}.
\newblock \showarticletitle{A machine learning approach to univariate time series forecasting of quarterly earnings}.
\newblock \bibinfo{journal}{\emph{Review of Quantitative Finance and Accounting}}  \bibinfo{volume}{55} (\bibinfo{year}{2020}), \bibinfo{pages}{1163--1179}.
\newblock


\bibitem[\protect\citeauthoryear{Friedman}{Friedman}{2001}]%
        {friedman2001greedy}
\bibfield{author}{\bibinfo{person}{Jerome~H Friedman}.} \bibinfo{year}{2001}\natexlab{}.
\newblock \showarticletitle{Greedy function approximation: a gradient boosting machine}.
\newblock \bibinfo{journal}{\emph{Annals of statistics}} (\bibinfo{year}{2001}), \bibinfo{pages}{1189--1232}.
\newblock


\bibitem[\protect\citeauthoryear{Godahewa, Bergmeir, Webb, Hyndman, and Montero-Manso}{Godahewa et~al\mbox{.}}{2021}]%
        {godahewa2021monash}
\bibfield{author}{\bibinfo{person}{Rakshitha Godahewa}, \bibinfo{person}{Christoph Bergmeir}, \bibinfo{person}{Geoffrey~I Webb}, \bibinfo{person}{Rob~J Hyndman}, {and} \bibinfo{person}{Pablo Montero-Manso}.} \bibinfo{year}{2021}\natexlab{}.
\newblock \showarticletitle{Monash time series forecasting archive}.
\newblock \bibinfo{journal}{\emph{arXiv preprint arXiv:2105.06643}} (\bibinfo{year}{2021}).
\newblock


\bibitem[\protect\citeauthoryear{Guo, Jensen, and Yang}{Guo et~al\mbox{.}}{2014}]%
        {DBLP:journals/sigmod/GuoJ014}
\bibfield{author}{\bibinfo{person}{Chenjuan Guo}, \bibinfo{person}{Christian~S. Jensen}, {and} \bibinfo{person}{Bin Yang}.} \bibinfo{year}{2014}\natexlab{}.
\newblock \showarticletitle{Towards Total Traffic Awareness}.
\newblock \bibinfo{journal}{\emph{{SIGMOD} Record}} \bibinfo{volume}{43}, \bibinfo{number}{3} (\bibinfo{year}{2014}), \bibinfo{pages}{18--23}.
\newblock


\bibitem[\protect\citeauthoryear{Guo, Yang, Andersen, Jensen, and Torp}{Guo et~al\mbox{.}}{2015}]%
        {guo2015ecomark}
\bibfield{author}{\bibinfo{person}{Chenjuan Guo}, \bibinfo{person}{Bin Yang}, \bibinfo{person}{Ove Andersen}, \bibinfo{person}{Christian~S Jensen}, {and} \bibinfo{person}{Kristian Torp}.} \bibinfo{year}{2015}\natexlab{}.
\newblock \showarticletitle{Ecomark 2.0: empowering eco-routing with vehicular environmental models and actual vehicle fuel consumption data}.
\newblock \bibinfo{journal}{\emph{GeoInformatica}}  \bibinfo{volume}{19} (\bibinfo{year}{2015}), \bibinfo{pages}{567--599}.
\newblock


\bibitem[\protect\citeauthoryear{Guo, Yang, Hu, Jensen, and Chen}{Guo et~al\mbox{.}}{2020}]%
        {DBLP:journals/vldb/GuoYHJC20}
\bibfield{author}{\bibinfo{person}{Chenjuan Guo}, \bibinfo{person}{Bin Yang}, \bibinfo{person}{Jilin Hu}, \bibinfo{person}{Christian~S. Jensen}, {and} \bibinfo{person}{Lu Chen}.} \bibinfo{year}{2020}\natexlab{}.
\newblock \showarticletitle{Context-aware, preference-based vehicle routing}.
\newblock \bibinfo{journal}{\emph{The VLDB Journal}} \bibinfo{volume}{29}, \bibinfo{number}{5} (\bibinfo{year}{2020}), \bibinfo{pages}{1149--1170}.
\newblock


\bibitem[\protect\citeauthoryear{Harvey}{Harvey}{1990}]%
        {harvey1990forecasting}
\bibfield{author}{\bibinfo{person}{Andrew~C Harvey}.} \bibinfo{year}{1990}\natexlab{}.
\newblock \showarticletitle{Forecasting, structural time series models and the Kalman filter}.
\newblock  (\bibinfo{year}{1990}).
\newblock


\bibitem[\protect\citeauthoryear{Herzen, L{\"a}ssig, Piazzetta, Neuer, Tafti, Raille, Van~Pottelbergh, Pasieka, Skrodzki, Huguenin, et~al\mbox{.}}{Herzen et~al\mbox{.}}{2022}]%
        {herzen2022darts}
\bibfield{author}{\bibinfo{person}{Julien Herzen}, \bibinfo{person}{Francesco L{\"a}ssig}, \bibinfo{person}{Samuele~Giuliano Piazzetta}, \bibinfo{person}{Thomas Neuer}, \bibinfo{person}{L{\'e}o Tafti}, \bibinfo{person}{Guillaume Raille}, \bibinfo{person}{Tomas Van~Pottelbergh}, \bibinfo{person}{Marek Pasieka}, \bibinfo{person}{Andrzej Skrodzki}, \bibinfo{person}{Nicolas Huguenin}, {et~al\mbox{.}}} \bibinfo{year}{2022}\natexlab{}.
\newblock \showarticletitle{Darts: User-friendly modern machine learning for time series}.
\newblock \bibinfo{journal}{\emph{The Journal of Machine Learning Research}} \bibinfo{volume}{23}, \bibinfo{number}{1} (\bibinfo{year}{2022}), \bibinfo{pages}{5442--5447}.
\newblock


\bibitem[\protect\citeauthoryear{Hu, Guo, Yang, and Jensen}{Hu et~al\mbox{.}}{2019}]%
        {hu2019stochastic}
\bibfield{author}{\bibinfo{person}{Jilin Hu}, \bibinfo{person}{Chenjuan Guo}, \bibinfo{person}{Bin Yang}, {and} \bibinfo{person}{Christian~S Jensen}.} \bibinfo{year}{2019}\natexlab{}.
\newblock \showarticletitle{Stochastic weight completion for road networks using graph convolutional networks}. In \bibinfo{booktitle}{\emph{{ICDE}}}. \bibinfo{pages}{1274--1285}.
\newblock


\bibitem[\protect\citeauthoryear{Hu, Yang, Guo, and Jensen}{Hu et~al\mbox{.}}{2018}]%
        {hu2018risk}
\bibfield{author}{\bibinfo{person}{Jilin Hu}, \bibinfo{person}{Bin Yang}, \bibinfo{person}{Chenjuan Guo}, {and} \bibinfo{person}{Christian~S Jensen}.} \bibinfo{year}{2018}\natexlab{}.
\newblock \showarticletitle{Risk-aware path selection with time-varying, uncertain travel costs: a time series approach}.
\newblock \bibinfo{journal}{\emph{The VLDB Journal}}  \bibinfo{volume}{27} (\bibinfo{year}{2018}), \bibinfo{pages}{179--200}.
\newblock


\bibitem[\protect\citeauthoryear{Hu, Yang, Jensen, and Ma}{Hu et~al\mbox{.}}{2017}]%
        {DBLP:journals/geoinformatica/HuYJM17}
\bibfield{author}{\bibinfo{person}{Jilin Hu}, \bibinfo{person}{Bin Yang}, \bibinfo{person}{Christian~S. Jensen}, {and} \bibinfo{person}{Yu Ma}.} \bibinfo{year}{2017}\natexlab{}.
\newblock \showarticletitle{Enabling time-dependent uncertain eco-weights for road networks}.
\newblock \bibinfo{journal}{\emph{GeoInformatica}} \bibinfo{volume}{21}, \bibinfo{number}{1} (\bibinfo{year}{2017}), \bibinfo{pages}{57--88}.
\newblock


\bibitem[\protect\citeauthoryear{Huang, Yang, Wang, Wang, Zhang, Xu, Chen, and Vazirgiannis}{Huang et~al\mbox{.}}{2022}]%
        {huang2022dgraph}
\bibfield{author}{\bibinfo{person}{Xuanwen Huang}, \bibinfo{person}{Yang Yang}, \bibinfo{person}{Yang Wang}, \bibinfo{person}{Chunping Wang}, \bibinfo{person}{Zhisheng Zhang}, \bibinfo{person}{Jiarong Xu}, \bibinfo{person}{Lei Chen}, {and} \bibinfo{person}{Michalis Vazirgiannis}.} \bibinfo{year}{2022}\natexlab{}.
\newblock \showarticletitle{Dgraph: A large-scale financial dataset for graph anomaly detection}.
\newblock \bibinfo{journal}{\emph{Advances in Neural Information Processing Systems}}  \bibinfo{volume}{35} (\bibinfo{year}{2022}), \bibinfo{pages}{22765--22777}.
\newblock


\bibitem[\protect\citeauthoryear{Hyndman, Koehler, Ord, and Snyder}{Hyndman et~al\mbox{.}}{2008}]%
        {hyndman2008forecasting}
\bibfield{author}{\bibinfo{person}{Rob Hyndman}, \bibinfo{person}{Anne~B Koehler}, \bibinfo{person}{J~Keith Ord}, {and} \bibinfo{person}{Ralph~D Snyder}.} \bibinfo{year}{2008}\natexlab{}.
\newblock \bibinfo{booktitle}{\emph{Forecasting with exponential smoothing: the state space approach}}.
\newblock


\bibitem[\protect\citeauthoryear{Hyndman and Koehler}{Hyndman and Koehler}{2006}]%
        {hyndman2006another}
\bibfield{author}{\bibinfo{person}{Rob~J Hyndman} {and} \bibinfo{person}{Anne~B Koehler}.} \bibinfo{year}{2006}\natexlab{}.
\newblock \showarticletitle{Another look at measures of forecast accuracy}.
\newblock \bibinfo{journal}{\emph{International journal of forecasting}} \bibinfo{volume}{22}, \bibinfo{number}{4} (\bibinfo{year}{2006}), \bibinfo{pages}{679--688}.
\newblock


\bibitem[\protect\citeauthoryear{Ke, Meng, Finley, Wang, Chen, Ma, Ye, and Liu}{Ke et~al\mbox{.}}{2017}]%
        {ke2017lightgbm}
\bibfield{author}{\bibinfo{person}{Guolin Ke}, \bibinfo{person}{Qi Meng}, \bibinfo{person}{Thomas Finley}, \bibinfo{person}{Taifeng Wang}, \bibinfo{person}{Wei Chen}, \bibinfo{person}{Weidong Ma}, \bibinfo{person}{Qiwei Ye}, {and} \bibinfo{person}{Tie-Yan Liu}.} \bibinfo{year}{2017}\natexlab{}.
\newblock \showarticletitle{Lightgbm: A highly efficient gradient boosting decision tree}.
\newblock \bibinfo{journal}{\emph{Advances in Neural Information Processing Systems}}  \bibinfo{volume}{30} (\bibinfo{year}{2017}).
\newblock


\bibitem[\protect\citeauthoryear{Kedem and Fokianos}{Kedem and Fokianos}{2005}]%
        {kedem2005regression}
\bibfield{author}{\bibinfo{person}{Benjamin Kedem} {and} \bibinfo{person}{Konstantinos Fokianos}.} \bibinfo{year}{2005}\natexlab{}.
\newblock \bibinfo{booktitle}{\emph{Regression models for time series analysis}}.
\newblock


\bibitem[\protect\citeauthoryear{Kieu, Yang, Guo, Jensen, Zhao, Huang, and Zheng}{Kieu et~al\mbox{.}}{2022}]%
        {DBLP:conf/icde/KieuYGJZHZ22}
\bibfield{author}{\bibinfo{person}{Tung Kieu}, \bibinfo{person}{Bin Yang}, \bibinfo{person}{Chenjuan Guo}, \bibinfo{person}{Christian~S. Jensen}, \bibinfo{person}{Yan Zhao}, \bibinfo{person}{Feiteng Huang}, {and} \bibinfo{person}{Kai Zheng}.} \bibinfo{year}{2022}\natexlab{}.
\newblock \showarticletitle{Robust and Explainable Autoencoders for Unsupervised Time Series Outlier Detection}. In \bibinfo{booktitle}{\emph{{ICDE}}}. \bibinfo{pages}{3038--3050}.
\newblock


\bibitem[\protect\citeauthoryear{Kim, Kim, Tae, Park, Choi, and Choo}{Kim et~al\mbox{.}}{2021}]%
        {kim2021reversible}
\bibfield{author}{\bibinfo{person}{Taesung Kim}, \bibinfo{person}{Jinhee Kim}, \bibinfo{person}{Yunwon Tae}, \bibinfo{person}{Cheonbok Park}, \bibinfo{person}{Jang-Ho Choi}, {and} \bibinfo{person}{Jaegul Choo}.} \bibinfo{year}{2021}\natexlab{}.
\newblock \showarticletitle{Reversible instance normalization for accurate time-series forecasting against distribution shift}. In \bibinfo{booktitle}{\emph{{ICLR}}}.
\newblock


\bibitem[\protect\citeauthoryear{Lee, Luo, Ngiam, Zhang, Zheng, Chen, Ooi, and Yip}{Lee et~al\mbox{.}}{2017}]%
        {lee2017big}
\bibfield{author}{\bibinfo{person}{Chonho Lee}, \bibinfo{person}{Zhaojing Luo}, \bibinfo{person}{Kee~Yuan Ngiam}, \bibinfo{person}{Meihui Zhang}, \bibinfo{person}{Kaiping Zheng}, \bibinfo{person}{Gang Chen}, \bibinfo{person}{Beng~Chin Ooi}, {and} \bibinfo{person}{Wei Luen~James Yip}.} \bibinfo{year}{2017}\natexlab{}.
\newblock \showarticletitle{Big healthcare data analytics: Challenges and applications}.
\newblock \bibinfo{journal}{\emph{Handbook of large-scale distributed computing in smart healthcare}} (\bibinfo{year}{2017}), \bibinfo{pages}{11--41}.
\newblock


\bibitem[\protect\citeauthoryear{Lee}{Lee}{2017}]%
        {lee2017anomaly}
\bibfield{author}{\bibinfo{person}{Doyup Lee}.} \bibinfo{year}{2017}\natexlab{}.
\newblock \showarticletitle{Anomaly detection in multivariate non-stationary time series for automatic DBMS diagnosis}. In \bibinfo{booktitle}{\emph{{ICMLA}}}. \bibinfo{pages}{412--419}.
\newblock


\bibitem[\protect\citeauthoryear{Li, Lu, Wang, and Dou}{Li et~al\mbox{.}}{2022}]%
        {li2022generative}
\bibfield{author}{\bibinfo{person}{Yan Li}, \bibinfo{person}{Xinjiang Lu}, \bibinfo{person}{Yaqing Wang}, {and} \bibinfo{person}{Dejing Dou}.} \bibinfo{year}{2022}\natexlab{}.
\newblock \showarticletitle{Generative time series forecasting with diffusion, denoise, and disentanglement}.
\newblock \bibinfo{journal}{\emph{Advances in Neural Information Processing Systems}}  \bibinfo{volume}{35} (\bibinfo{year}{2022}), \bibinfo{pages}{23009--23022}.
\newblock


\bibitem[\protect\citeauthoryear{Liang, Shao, Wang, Zhang, Sun, and Xu}{Liang et~al\mbox{.}}{2022}]%
        {liang2022basicts}
\bibfield{author}{\bibinfo{person}{Yubo Liang}, \bibinfo{person}{Zezhi Shao}, \bibinfo{person}{Fei Wang}, \bibinfo{person}{Zhao Zhang}, \bibinfo{person}{Tao Sun}, {and} \bibinfo{person}{Yongjun Xu}.} \bibinfo{year}{2022}\natexlab{}.
\newblock \showarticletitle{BasicTS: An Open Source Fair Multivariate Time Series Prediction Benchmark}. In \bibinfo{booktitle}{\emph{International Symposium on Benchmarking, Measuring and Optimization}}. \bibinfo{pages}{87--101}.
\newblock


\bibitem[\protect\citeauthoryear{Lin, Hu, Guo, Yang, Jensen, Lin, and Wan}{Lin et~al\mbox{.}}{2024}]%
        {lin2024genstl}
\bibfield{author}{\bibinfo{person}{Yan Lin}, \bibinfo{person}{Jilin Hu}, \bibinfo{person}{Shengnan Guo}, \bibinfo{person}{Bin Yang}, \bibinfo{person}{Christian~S Jensen}, \bibinfo{person}{Youfang Lin}, {and} \bibinfo{person}{Huaiyu Wan}.} \bibinfo{year}{2024}\natexlab{}.
\newblock \showarticletitle{GenSTL: General Sparse Trajectory Learning via Auto-regressive Generation of Feature Domains}.
\newblock \bibinfo{journal}{\emph{arXiv preprint arXiv:2402.07232}} (\bibinfo{year}{2024}).
\newblock


\bibitem[\protect\citeauthoryear{Lin, Wan, Guo, Hu, Jensen, and Lin}{Lin et~al\mbox{.}}{2023a}]%
        {lin2023pre}
\bibfield{author}{\bibinfo{person}{Yan Lin}, \bibinfo{person}{Huaiyu Wan}, \bibinfo{person}{Shengnan Guo}, \bibinfo{person}{Jilin Hu}, \bibinfo{person}{Christian~S Jensen}, {and} \bibinfo{person}{Youfang Lin}.} \bibinfo{year}{2023}\natexlab{a}.
\newblock \showarticletitle{Pre-Training General Trajectory Embeddings With Maximum Multi-View Entropy Coding}.
\newblock \bibinfo{journal}{\emph{IEEE Transactions on Knowledge and Data Engineering}} (\bibinfo{year}{2023}).
\newblock


\bibitem[\protect\citeauthoryear{Lin, Wan, Guo, and Lin}{Lin et~al\mbox{.}}{2021}]%
        {lin2021pre}
\bibfield{author}{\bibinfo{person}{Yan Lin}, \bibinfo{person}{Huaiyu Wan}, \bibinfo{person}{Shengnan Guo}, {and} \bibinfo{person}{Youfang Lin}.} \bibinfo{year}{2021}\natexlab{}.
\newblock \showarticletitle{Pre-training context and time aware location embeddings from spatial-temporal trajectories for user next location prediction}. In \bibinfo{booktitle}{\emph{{AAAI}}}, Vol.~\bibinfo{volume}{35}. \bibinfo{pages}{4241--4248}.
\newblock


\bibitem[\protect\citeauthoryear{Lin, Wan, Hu, Guo, Yang, Lin, and Jensen}{Lin et~al\mbox{.}}{2023b}]%
        {lin2023origin}
\bibfield{author}{\bibinfo{person}{Yan Lin}, \bibinfo{person}{Huaiyu Wan}, \bibinfo{person}{Jilin Hu}, \bibinfo{person}{Shengnan Guo}, \bibinfo{person}{Bin Yang}, \bibinfo{person}{Youfang Lin}, {and} \bibinfo{person}{Christian~S Jensen}.} \bibinfo{year}{2023}\natexlab{b}.
\newblock \showarticletitle{Origin-destination travel time oracle for map-based services}.
\newblock \bibinfo{journal}{\emph{Proceedings of the ACM on Management of Data}} \bibinfo{volume}{1}, \bibinfo{number}{3} (\bibinfo{year}{2023}), \bibinfo{pages}{1--27}.
\newblock


\bibitem[\protect\citeauthoryear{Liu, Wu, Wang, and Long}{Liu et~al\mbox{.}}{2022}]%
        {liu2022non}
\bibfield{author}{\bibinfo{person}{Yong Liu}, \bibinfo{person}{Haixu Wu}, \bibinfo{person}{Jianmin Wang}, {and} \bibinfo{person}{Mingsheng Long}.} \bibinfo{year}{2022}\natexlab{}.
\newblock \showarticletitle{Non-stationary transformers: Exploring the stationarity in time series forecasting}.
\newblock \bibinfo{journal}{\emph{Advances in Neural Information Processing Systems}}  \bibinfo{volume}{35} (\bibinfo{year}{2022}), \bibinfo{pages}{9881--9893}.
\newblock


\bibitem[\protect\citeauthoryear{Lu, Cohen, Zhou, and Tian}{Lu et~al\mbox{.}}{2007}]%
        {lu2007feature}
\bibfield{author}{\bibinfo{person}{Yijuan Lu}, \bibinfo{person}{Ira Cohen}, \bibinfo{person}{Xiang~Sean Zhou}, {and} \bibinfo{person}{Qi Tian}.} \bibinfo{year}{2007}\natexlab{}.
\newblock \showarticletitle{Feature selection using principal feature analysis}. In \bibinfo{booktitle}{\emph{{ACM MM}}}. \bibinfo{pages}{301--304}.
\newblock


\bibitem[\protect\citeauthoryear{Lubba, Sethi, Knaute, Schultz, Fulcher, and Jones}{Lubba et~al\mbox{.}}{2019}]%
        {lubba2019catch22}
\bibfield{author}{\bibinfo{person}{Carl~H Lubba}, \bibinfo{person}{Sarab~S Sethi}, \bibinfo{person}{Philip Knaute}, \bibinfo{person}{Simon~R Schultz}, \bibinfo{person}{Ben~D Fulcher}, {and} \bibinfo{person}{Nick~S Jones}.} \bibinfo{year}{2019}\natexlab{}.
\newblock \showarticletitle{catch22: CAnonical Time-series CHaracteristics: Selected through highly comparative time-series analysis}.
\newblock \bibinfo{journal}{\emph{Data Mining and Knowledge Discovery}} \bibinfo{volume}{33}, \bibinfo{number}{6} (\bibinfo{year}{2019}), \bibinfo{pages}{1821--1852}.
\newblock


\bibitem[\protect\citeauthoryear{Makridakis and Hibon}{Makridakis and Hibon}{2000}]%
        {makridakis2000m3}
\bibfield{author}{\bibinfo{person}{Spyros Makridakis} {and} \bibinfo{person}{Michele Hibon}.} \bibinfo{year}{2000}\natexlab{}.
\newblock \showarticletitle{The M3-Competition: results, conclusions and implications}.
\newblock \bibinfo{journal}{\emph{International journal of forecasting}} \bibinfo{volume}{16}, \bibinfo{number}{4} (\bibinfo{year}{2000}), \bibinfo{pages}{451--476}.
\newblock


\bibitem[\protect\citeauthoryear{Makridakis, Spiliotis, and Assimakopoulos}{Makridakis et~al\mbox{.}}{2018}]%
        {makridakis2018m4}
\bibfield{author}{\bibinfo{person}{Spyros Makridakis}, \bibinfo{person}{Evangelos Spiliotis}, {and} \bibinfo{person}{Vassilios Assimakopoulos}.} \bibinfo{year}{2018}\natexlab{}.
\newblock \showarticletitle{The M4 Competition: Results, findings, conclusion and way forward}.
\newblock \bibinfo{journal}{\emph{International Journal of Forecasting}} \bibinfo{volume}{34}, \bibinfo{number}{4} (\bibinfo{year}{2018}), \bibinfo{pages}{802--808}.
\newblock


\bibitem[\protect\citeauthoryear{McCracken and Ng}{McCracken and Ng}{2016}]%
        {mccracken2016fred}
\bibfield{author}{\bibinfo{person}{Michael~W McCracken} {and} \bibinfo{person}{Serena Ng}.} \bibinfo{year}{2016}\natexlab{}.
\newblock \showarticletitle{FRED-MD: A monthly database for macroeconomic research}.
\newblock \bibinfo{journal}{\emph{Journal of Business \& Economic Statistics}} \bibinfo{volume}{34}, \bibinfo{number}{4} (\bibinfo{year}{2016}), \bibinfo{pages}{574--589}.
\newblock


\bibitem[\protect\citeauthoryear{Mei, He, Harley, Habetler, and Qu}{Mei et~al\mbox{.}}{2014}]%
        {mei2014random}
\bibfield{author}{\bibinfo{person}{Jie Mei}, \bibinfo{person}{Dawei He}, \bibinfo{person}{Ronald Harley}, \bibinfo{person}{Thomas Habetler}, {and} \bibinfo{person}{Guannan Qu}.} \bibinfo{year}{2014}\natexlab{}.
\newblock \showarticletitle{A random forest method for real-time price forecasting in New York electricity market}. In \bibinfo{booktitle}{\emph{2014 IEEE PES General Meeting| Conference \& Exposition}}. \bibinfo{pages}{1--5}.
\newblock


\bibitem[\protect\citeauthoryear{Miao, Zhao, Guo, Yang, Kai, Huang, Xie, and Jensen}{Miao et~al\mbox{.}}{2024}]%
        {haoicde24}
\bibfield{author}{\bibinfo{person}{Hao Miao}, \bibinfo{person}{Yan Zhao}, \bibinfo{person}{Chenjuan Guo}, \bibinfo{person}{Bin Yang}, \bibinfo{person}{Zheng Kai}, \bibinfo{person}{Feiteng Huang}, \bibinfo{person}{Jiandong Xie}, {and} \bibinfo{person}{Christian~S. Jensen}.} \bibinfo{year}{2024}\natexlab{}.
\newblock \showarticletitle{A Unified Replay-based Continuous Learning Framework for Spatio-Temporal Prediction on Streaming Data}.
\newblock \bibinfo{journal}{\emph{{ICDE}}} (\bibinfo{year}{2024}).
\newblock


\bibitem[\protect\citeauthoryear{Miao, Wu, Wang, Gao, Mao, and Yin}{Miao et~al\mbox{.}}{2021}]%
        {miao2021generative}
\bibfield{author}{\bibinfo{person}{Xiaoye Miao}, \bibinfo{person}{Yangyang Wu}, \bibinfo{person}{Jun Wang}, \bibinfo{person}{Yunjun Gao}, \bibinfo{person}{Xudong Mao}, {and} \bibinfo{person}{Jianwei Yin}.} \bibinfo{year}{2021}\natexlab{}.
\newblock \showarticletitle{Generative semi-supervised learning for multivariate time series imputation}. In \bibinfo{booktitle}{\emph{{AAAI}}}, Vol.~\bibinfo{volume}{35}. \bibinfo{pages}{8983--8991}.
\newblock


\bibitem[\protect\citeauthoryear{Mo, Pang, and Liu}{Mo et~al\mbox{.}}{2022}]%
        {mo2022ths}
\bibfield{author}{\bibinfo{person}{Xian Mo}, \bibinfo{person}{Jun Pang}, {and} \bibinfo{person}{Zhiming Liu}.} \bibinfo{year}{2022}\natexlab{}.
\newblock \showarticletitle{THS-GWNN: a deep learning framework for temporal network link prediction}.
\newblock \bibinfo{journal}{\emph{Frontiers of Computer Science}} \bibinfo{volume}{16}, \bibinfo{number}{2} (\bibinfo{year}{2022}), \bibinfo{pages}{162304}.
\newblock


\bibitem[\protect\citeauthoryear{Nason}{Nason}{2006}]%
        {nason2006stationary}
\bibfield{author}{\bibinfo{person}{Guy~P Nason}.} \bibinfo{year}{2006}\natexlab{}.
\newblock \showarticletitle{Stationary and non-stationary time series}.
\newblock  (\bibinfo{year}{2006}).
\newblock


\bibitem[\protect\citeauthoryear{Nie, Nguyen, Sinthong, and Kalagnanam}{Nie et~al\mbox{.}}{2022}]%
        {nie2022time}
\bibfield{author}{\bibinfo{person}{Yuqi Nie}, \bibinfo{person}{Nam~H Nguyen}, \bibinfo{person}{Phanwadee Sinthong}, {and} \bibinfo{person}{Jayant Kalagnanam}.} \bibinfo{year}{2022}\natexlab{}.
\newblock \showarticletitle{A time series is worth 64 words: Long-term forecasting with transformers}.
\newblock \bibinfo{journal}{\emph{arXiv preprint arXiv:2211.14730}} (\bibinfo{year}{2022}).
\newblock


\bibitem[\protect\citeauthoryear{O'Grady}{O'Grady}{1982}]%
        {o1982measures}
\bibfield{author}{\bibinfo{person}{Kevin~E O'Grady}.} \bibinfo{year}{1982}\natexlab{}.
\newblock \showarticletitle{Measures of explained variance: Cautions and limitations}.
\newblock \bibinfo{journal}{\emph{Psychological Bulletin}} \bibinfo{volume}{92}, \bibinfo{number}{3} (\bibinfo{year}{1982}), \bibinfo{pages}{766}.
\newblock


\bibitem[\protect\citeauthoryear{Oreshkin, Carpov, Chapados, and Bengio}{Oreshkin et~al\mbox{.}}{2019}]%
        {oreshkin2019n}
\bibfield{author}{\bibinfo{person}{Boris~N Oreshkin}, \bibinfo{person}{Dmitri Carpov}, \bibinfo{person}{Nicolas Chapados}, {and} \bibinfo{person}{Yoshua Bengio}.} \bibinfo{year}{2019}\natexlab{}.
\newblock \showarticletitle{N-BEATS: Neural basis expansion analysis for interpretable time series forecasting}.
\newblock \bibinfo{journal}{\emph{arXiv preprint arXiv:1905.10437}} (\bibinfo{year}{2019}).
\newblock


\bibitem[\protect\citeauthoryear{Pan, Wang, Zhang, Yang, Cheng, Chen, Guo, Wen, Tian, Dou, et~al\mbox{.}}{Pan et~al\mbox{.}}{2023}]%
        {pan2023magicscaler}
\bibfield{author}{\bibinfo{person}{Zhicheng Pan}, \bibinfo{person}{Yihang Wang}, \bibinfo{person}{Yingying Zhang}, \bibinfo{person}{Sean~Bin Yang}, \bibinfo{person}{Yunyao Cheng}, \bibinfo{person}{Peng Chen}, \bibinfo{person}{Chenjuan Guo}, \bibinfo{person}{Qingsong Wen}, \bibinfo{person}{Xiduo Tian}, \bibinfo{person}{Yunliang Dou}, {et~al\mbox{.}}} \bibinfo{year}{2023}\natexlab{}.
\newblock \showarticletitle{Magicscaler: Uncertainty-aware, predictive autoscaling}.
\newblock \bibinfo{journal}{\emph{Proc. {VLDB} Endow.}} \bibinfo{volume}{16}, \bibinfo{number}{12} (\bibinfo{year}{2023}), \bibinfo{pages}{3808--3821}.
\newblock


\bibitem[\protect\citeauthoryear{Paszke, Gross, Massa, Lerer, Bradbury, Chanan, Killeen, Lin, Gimelshein, Antiga, et~al\mbox{.}}{Paszke et~al\mbox{.}}{2019}]%
        {paszke2019pytorch}
\bibfield{author}{\bibinfo{person}{Adam Paszke}, \bibinfo{person}{Sam Gross}, \bibinfo{person}{Francisco Massa}, \bibinfo{person}{Adam Lerer}, \bibinfo{person}{James Bradbury}, \bibinfo{person}{Gregory Chanan}, \bibinfo{person}{Trevor Killeen}, \bibinfo{person}{Zeming Lin}, \bibinfo{person}{Natalia Gimelshein}, \bibinfo{person}{Luca Antiga}, {et~al\mbox{.}}} \bibinfo{year}{2019}\natexlab{}.
\newblock \showarticletitle{Pytorch: An imperative style, high-performance deep learning library}.
\newblock \bibinfo{journal}{\emph{Advances in Neural Information Processing Systems}}  \bibinfo{volume}{32} (\bibinfo{year}{2019}).
\newblock


\bibitem[\protect\citeauthoryear{Pedersen, Yang, and Jensen}{Pedersen et~al\mbox{.}}{2020}]%
        {DBLP:journals/pvldb/PedersenYJ20}
\bibfield{author}{\bibinfo{person}{Simon~Aagaard Pedersen}, \bibinfo{person}{Bin Yang}, {and} \bibinfo{person}{Christian~S. Jensen}.} \bibinfo{year}{2020}\natexlab{}.
\newblock \showarticletitle{Anytime Stochastic Routing with Hybrid Learning}.
\newblock \bibinfo{journal}{\emph{Proc. {VLDB} Endow.}} \bibinfo{volume}{13}, \bibinfo{number}{9} (\bibinfo{year}{2020}), \bibinfo{pages}{1555--1567}.
\newblock


\bibitem[\protect\citeauthoryear{Qi, Hu, Guo, Huang, Zhou, Xu, Fu, and Zhou}{Qi et~al\mbox{.}}{2023}]%
        {qi2023high}
\bibfield{author}{\bibinfo{person}{Xuecheng Qi}, \bibinfo{person}{Huiqi Hu}, \bibinfo{person}{Jinwei Guo}, \bibinfo{person}{Chenchen Huang}, \bibinfo{person}{Xuan Zhou}, \bibinfo{person}{Ning Xu}, \bibinfo{person}{Yu Fu}, {and} \bibinfo{person}{Aoying Zhou}.} \bibinfo{year}{2023}\natexlab{}.
\newblock \showarticletitle{High-availability in-memory key-value store using RDMA and Optane DCPMM}.
\newblock \bibinfo{journal}{\emph{Frontiers of Computer Science}} \bibinfo{volume}{17}, \bibinfo{number}{1} (\bibinfo{year}{2023}), \bibinfo{pages}{171603}.
\newblock


\bibitem[\protect\citeauthoryear{Qiao, Pham, Cao, Le, Suganthan, Jiang, and Savitha}{Qiao et~al\mbox{.}}{2024}]%
        {qiao2024class}
\bibfield{author}{\bibinfo{person}{Zhongzheng Qiao}, \bibinfo{person}{Quang Pham}, \bibinfo{person}{Zhen Cao}, \bibinfo{person}{Hoang~H Le}, \bibinfo{person}{PN Suganthan}, \bibinfo{person}{Xudong Jiang}, {and} \bibinfo{person}{Ramasamy Savitha}.} \bibinfo{year}{2024}\natexlab{}.
\newblock \showarticletitle{Class-incremental Learning for Time Series: Benchmark and Evaluation}.
\newblock \bibinfo{journal}{\emph{arXiv preprint arXiv:2402.12035}} (\bibinfo{year}{2024}).
\newblock


\bibitem[\protect\citeauthoryear{Salinas, Flunkert, Gasthaus, and Januschowski}{Salinas et~al\mbox{.}}{2020}]%
        {salinas2020deepar}
\bibfield{author}{\bibinfo{person}{David Salinas}, \bibinfo{person}{Valentin Flunkert}, \bibinfo{person}{Jan Gasthaus}, {and} \bibinfo{person}{Tim Januschowski}.} \bibinfo{year}{2020}\natexlab{}.
\newblock \showarticletitle{DeepAR: Probabilistic forecasting with autoregressive recurrent networks}.
\newblock \bibinfo{journal}{\emph{International Journal of Forecasting}} \bibinfo{volume}{36}, \bibinfo{number}{3} (\bibinfo{year}{2020}), \bibinfo{pages}{1181--1191}.
\newblock


\bibitem[\protect\citeauthoryear{Sezer, Gudelek, and Ozbayoglu}{Sezer et~al\mbox{.}}{2020}]%
        {sezer2020financial}
\bibfield{author}{\bibinfo{person}{Omer~Berat Sezer}, \bibinfo{person}{Mehmet~Ugur Gudelek}, {and} \bibinfo{person}{Ahmet~Murat Ozbayoglu}.} \bibinfo{year}{2020}\natexlab{}.
\newblock \showarticletitle{Financial time series forecasting with deep learning: A systematic literature review: 2005--2019}.
\newblock \bibinfo{journal}{\emph{Applied soft computing}}  \bibinfo{volume}{90} (\bibinfo{year}{2020}), \bibinfo{pages}{106181}.
\newblock


\bibitem[\protect\citeauthoryear{Shao, Wang, Xu, Wei, Yu, Zhang, Yao, Jin, Cao, Cong, et~al\mbox{.}}{Shao et~al\mbox{.}}{2023}]%
        {shao2023exploring}
\bibfield{author}{\bibinfo{person}{Zezhi Shao}, \bibinfo{person}{Fei Wang}, \bibinfo{person}{Yongjun Xu}, \bibinfo{person}{Wei Wei}, \bibinfo{person}{Chengqing Yu}, \bibinfo{person}{Zhao Zhang}, \bibinfo{person}{Di Yao}, \bibinfo{person}{Guangyin Jin}, \bibinfo{person}{Xin Cao}, \bibinfo{person}{Gao Cong}, {et~al\mbox{.}}} \bibinfo{year}{2023}\natexlab{}.
\newblock \showarticletitle{Exploring Progress in Multivariate Time Series Forecasting: Comprehensive Benchmarking and Heterogeneity Analysis}.
\newblock \bibinfo{journal}{\emph{arXiv preprint arXiv:2310.06119}} (\bibinfo{year}{2023}).
\newblock


\bibitem[\protect\citeauthoryear{Suilin}{Suilin}{2017}]%
        {msmape}
\bibfield{author}{\bibinfo{person}{A Suilin}.} \bibinfo{year}{2017}\natexlab{}.
\newblock \bibinfo{booktitle}{\emph{kaggle-web-traffic}}.
\newblock
\urldef\tempurl%
\url{https://github.com/Arturus/kaggle-web-traffic}
\showURL{%
\tempurl}


\bibitem[\protect\citeauthoryear{Sun, Ning, Shen, and Nie}{Sun et~al\mbox{.}}{2023}]%
        {sun2023graph}
\bibfield{author}{\bibinfo{person}{Chenchen Sun}, \bibinfo{person}{Yan Ning}, \bibinfo{person}{Derong Shen}, {and} \bibinfo{person}{Tiezheng Nie}.} \bibinfo{year}{2023}\natexlab{}.
\newblock \showarticletitle{Graph Neural Network-Based Short-Term Load Forecasting with Temporal Convolution}.
\newblock \bibinfo{journal}{\emph{Data Science and Engineering}} (\bibinfo{year}{2023}), \bibinfo{pages}{1--20}.
\newblock


\bibitem[\protect\citeauthoryear{Tan, Bergmeir, Petitjean, and Webb}{Tan et~al\mbox{.}}{2020}]%
        {tanmonash}
\bibfield{author}{\bibinfo{person}{Chang~Wei Tan}, \bibinfo{person}{Christoph Bergmeir}, \bibinfo{person}{Fran{\c{c}}ois Petitjean}, {and} \bibinfo{person}{Geoffrey~I. Webb}.} \bibinfo{year}{2020}\natexlab{}.
\newblock \showarticletitle{Monash University, UEA, {UCR} Time Series Regression Archive}.
\newblock \bibinfo{journal}{\emph{arXiv preprint arXiv:2006.10996}} (\bibinfo{year}{2020}).
\newblock


\bibitem[\protect\citeauthoryear{Toda and Phillips}{Toda and Phillips}{1994}]%
        {toda1994vector}
\bibfield{author}{\bibinfo{person}{Hiro~Y Toda} {and} \bibinfo{person}{Peter~CB Phillips}.} \bibinfo{year}{1994}\natexlab{}.
\newblock \showarticletitle{Vector autoregression and causality: a theoretical overview and simulation study}.
\newblock \bibinfo{journal}{\emph{Econometric reviews}} \bibinfo{volume}{13}, \bibinfo{number}{2} (\bibinfo{year}{1994}), \bibinfo{pages}{259--285}.
\newblock


\bibitem[\protect\citeauthoryear{Tran, Nguyen, and Shahabi}{Tran et~al\mbox{.}}{2019}]%
        {tran2019representation}
\bibfield{author}{\bibinfo{person}{Luan Tran}, \bibinfo{person}{Manh Nguyen}, {and} \bibinfo{person}{Cyrus Shahabi}.} \bibinfo{year}{2019}\natexlab{}.
\newblock \showarticletitle{Representation learning for early sepsis prediction}. In \bibinfo{booktitle}{\emph{2019 Computing in Cardiology (CinC)}}. \bibinfo{pages}{1--4}.
\newblock


\bibitem[\protect\citeauthoryear{Trindade}{Trindade}{2015}]%
        {misc_electricityloaddiagrams20112014_321}
\bibfield{author}{\bibinfo{person}{Artur Trindade}.} \bibinfo{year}{2015}\natexlab{}.
\newblock \bibinfo{title}{{ElectricityLoadDiagrams20112014}}.
\newblock \bibinfo{howpublished}{UCI Machine Learning Repository}.
\newblock
\newblock
\shownote{{DOI}: https://doi.org/10.24432/C58C86.}


\bibitem[\protect\citeauthoryear{Wan, Li, Wang, Chen, Gao, Jiang, and Pu}{Wan et~al\mbox{.}}{2022}]%
        {wan2022mttpre}
\bibfield{author}{\bibinfo{person}{Feng Wan}, \bibinfo{person}{Linsen Li}, \bibinfo{person}{Ke Wang}, \bibinfo{person}{Lu Chen}, \bibinfo{person}{Yunjun Gao}, \bibinfo{person}{Weihao Jiang}, {and} \bibinfo{person}{Shiliang Pu}.} \bibinfo{year}{2022}\natexlab{}.
\newblock \showarticletitle{MTTPRE: a multi-scale spatial-temporal model for travel time prediction}. In \bibinfo{booktitle}{\emph{{SIGSPATIAL}}}. \bibinfo{pages}{1--10}.
\newblock


\bibitem[\protect\citeauthoryear{Wang, Peng, Huang, Wang, Chen, and Xiao}{Wang et~al\mbox{.}}{2022}]%
        {wang2022micn}
\bibfield{author}{\bibinfo{person}{Huiqiang Wang}, \bibinfo{person}{Jian Peng}, \bibinfo{person}{Feihu Huang}, \bibinfo{person}{Jince Wang}, \bibinfo{person}{Junhui Chen}, {and} \bibinfo{person}{Yifei Xiao}.} \bibinfo{year}{2022}\natexlab{}.
\newblock \showarticletitle{Micn: Multi-scale local and global context modeling for long-term series forecasting}. In \bibinfo{booktitle}{\emph{{ICLR}}}.
\newblock


\bibitem[\protect\citeauthoryear{Wang, Li, Wang, Liu, Chen, Chen, Liu, Wu, Li, and Gao}{Wang et~al\mbox{.}}{2023}]%
        {wang2023real}
\bibfield{author}{\bibinfo{person}{Jiaqi Wang}, \bibinfo{person}{Tianyi Li}, \bibinfo{person}{Anni Wang}, \bibinfo{person}{Xiaoze Liu}, \bibinfo{person}{Lu Chen}, \bibinfo{person}{Jie Chen}, \bibinfo{person}{Jianye Liu}, \bibinfo{person}{Junyang Wu}, \bibinfo{person}{Feifei Li}, {and} \bibinfo{person}{Yunjun Gao}.} \bibinfo{year}{2023}\natexlab{}.
\newblock \showarticletitle{Real-time Workload Pattern Analysis for Large-scale Cloud Databases}.
\newblock \bibinfo{journal}{\emph{arXiv preprint arXiv:2307.02626}} (\bibinfo{year}{2023}).
\newblock


\bibitem[\protect\citeauthoryear{Wei, Li, Huang, Chen, and He}{Wei et~al\mbox{.}}{2022}]%
        {wei2022cancer}
\bibfield{author}{\bibinfo{person}{Kaimin Wei}, \bibinfo{person}{Tianqi Li}, \bibinfo{person}{Feiran Huang}, \bibinfo{person}{Jinpeng Chen}, {and} \bibinfo{person}{Zefan He}.} \bibinfo{year}{2022}\natexlab{}.
\newblock \showarticletitle{Cancer classification with data augmentation based on generative adversarial networks}.
\newblock \bibinfo{journal}{\emph{Frontiers of Computer Science}}  \bibinfo{volume}{16} (\bibinfo{year}{2022}), \bibinfo{pages}{1--11}.
\newblock


\bibitem[\protect\citeauthoryear{Wu, Hu, Liu, Zhou, Wang, and Long}{Wu et~al\mbox{.}}{2022}]%
        {wu2022timesnet}
\bibfield{author}{\bibinfo{person}{Haixu Wu}, \bibinfo{person}{Tengge Hu}, \bibinfo{person}{Yong Liu}, \bibinfo{person}{Hang Zhou}, \bibinfo{person}{Jianmin Wang}, {and} \bibinfo{person}{Mingsheng Long}.} \bibinfo{year}{2022}\natexlab{}.
\newblock \showarticletitle{Timesnet: Temporal 2d-variation modeling for general time series analysis}.
\newblock \bibinfo{journal}{\emph{arXiv preprint arXiv:2210.02186}} (\bibinfo{year}{2022}).
\newblock


\bibitem[\protect\citeauthoryear{Wu, Xu, Wang, and Long}{Wu et~al\mbox{.}}{2021a}]%
        {wu2021autoformer}
\bibfield{author}{\bibinfo{person}{Haixu Wu}, \bibinfo{person}{Jiehui Xu}, \bibinfo{person}{Jianmin Wang}, {and} \bibinfo{person}{Mingsheng Long}.} \bibinfo{year}{2021}\natexlab{a}.
\newblock \showarticletitle{Autoformer: Decomposition transformers with auto-correlation for long-term series forecasting}.
\newblock \bibinfo{journal}{\emph{Advances in Neural Information Processing Systems}}  \bibinfo{volume}{34} (\bibinfo{year}{2021}), \bibinfo{pages}{22419--22430}.
\newblock


\bibitem[\protect\citeauthoryear{Wu, Zhang, Guo, He, Yang, and Jensen}{Wu et~al\mbox{.}}{2021b}]%
        {wu2021autocts}
\bibfield{author}{\bibinfo{person}{Xinle Wu}, \bibinfo{person}{Dalin Zhang}, \bibinfo{person}{Chenjuan Guo}, \bibinfo{person}{Chaoyang He}, \bibinfo{person}{Bin Yang}, {and} \bibinfo{person}{Christian~S Jensen}.} \bibinfo{year}{2021}\natexlab{b}.
\newblock \showarticletitle{AutoCTS: Automated correlated time series forecasting}.
\newblock \bibinfo{journal}{\emph{Proc. {VLDB} Endow.}} \bibinfo{volume}{15}, \bibinfo{number}{4} (\bibinfo{year}{2021}), \bibinfo{pages}{971--983}.
\newblock


\bibitem[\protect\citeauthoryear{Wu, Zhang, Zhang, Guo, Yang, and Jensen}{Wu et~al\mbox{.}}{2023}]%
        {wu2023autocts+}
\bibfield{author}{\bibinfo{person}{Xinle Wu}, \bibinfo{person}{Dalin Zhang}, \bibinfo{person}{Miao Zhang}, \bibinfo{person}{Chenjuan Guo}, \bibinfo{person}{Bin Yang}, {and} \bibinfo{person}{Christian~S Jensen}.} \bibinfo{year}{2023}\natexlab{}.
\newblock \showarticletitle{AutoCTS+: Joint Neural Architecture and Hyperparameter Search for Correlated Time Series Forecasting}.
\newblock \bibinfo{journal}{\emph{Proceedings of the ACM on Management of Data}} \bibinfo{volume}{1}, \bibinfo{number}{1} (\bibinfo{year}{2023}), \bibinfo{pages}{1--26}.
\newblock


\bibitem[\protect\citeauthoryear{Xu, Chen, Gong, Liu, Yu, and Nie}{Xu et~al\mbox{.}}{2023}]%
        {xu2023tme}
\bibfield{author}{\bibinfo{person}{Ronghui Xu}, \bibinfo{person}{Meng Chen}, \bibinfo{person}{Yongshun Gong}, \bibinfo{person}{Yang Liu}, \bibinfo{person}{Xiaohui Yu}, {and} \bibinfo{person}{Liqiang Nie}.} \bibinfo{year}{2023}\natexlab{}.
\newblock \showarticletitle{TME: Tree-guided Multi-task Embedding Learning towards Semantic Venue Annotation}.
\newblock \bibinfo{journal}{\emph{ACM Transactions on Information Systems}} \bibinfo{volume}{41}, \bibinfo{number}{4} (\bibinfo{year}{2023}), \bibinfo{pages}{1--24}.
\newblock


\bibitem[\protect\citeauthoryear{Yang, Guo, Hu, Tang, and Yang}{Yang et~al\mbox{.}}{2021}]%
        {DBLP:conf/ijcai/YangGHT021}
\bibfield{author}{\bibinfo{person}{Sean~Bin Yang}, \bibinfo{person}{Chenjuan Guo}, \bibinfo{person}{Jilin Hu}, \bibinfo{person}{Jian Tang}, {and} \bibinfo{person}{Bin Yang}.} \bibinfo{year}{2021}\natexlab{}.
\newblock \showarticletitle{Unsupervised Path Representation Learning with Curriculum Negative Sampling}. In \bibinfo{booktitle}{\emph{{IJCAI}}}. \bibinfo{pages}{3286--3292}.
\newblock


\bibitem[\protect\citeauthoryear{Yang, Hu, Guo, Yang, and Jensen}{Yang et~al\mbox{.}}{2023}]%
        {yang2023lightpath}
\bibfield{author}{\bibinfo{person}{Sean~Bin Yang}, \bibinfo{person}{Jilin Hu}, \bibinfo{person}{Chenjuan Guo}, \bibinfo{person}{Bin Yang}, {and} \bibinfo{person}{Christian~S Jensen}.} \bibinfo{year}{2023}\natexlab{}.
\newblock \showarticletitle{Lightpath: Lightweight and scalable path representation learning}. In \bibinfo{booktitle}{\emph{{SIGKDD}}}. \bibinfo{pages}{2999--3010}.
\newblock


\bibitem[\protect\citeauthoryear{Yao, Li, Jie, Jie, Li, Chen, Wang, Li, and Gao}{Yao et~al\mbox{.}}{2023}]%
        {yao2023simplets}
\bibfield{author}{\bibinfo{person}{Yuanyuan Yao}, \bibinfo{person}{Dimeng Li}, \bibinfo{person}{Hailiang Jie}, \bibinfo{person}{Hailiang Jie}, \bibinfo{person}{Tianyi Li}, \bibinfo{person}{Jie Chen}, \bibinfo{person}{Jiaqi Wang}, \bibinfo{person}{Feifei Li}, {and} \bibinfo{person}{Yunjun Gao}.} \bibinfo{year}{2023}\natexlab{}.
\newblock \showarticletitle{SimpleTS: An efficient and universal model selection framework for time series forecasting}.
\newblock \bibinfo{journal}{\emph{Proc. {VLDB} Endow.}} \bibinfo{volume}{16}, \bibinfo{number}{12} (\bibinfo{year}{2023}), \bibinfo{pages}{3741--3753}.
\newblock


\bibitem[\protect\citeauthoryear{Yu, Hu, Zhou, Guo, Yang, and Li}{Yu et~al\mbox{.}}{2023}]%
        {yu2023cgf}
\bibfield{author}{\bibinfo{person}{Haomin Yu}, \bibinfo{person}{Jilin Hu}, \bibinfo{person}{Xinyuan Zhou}, \bibinfo{person}{Chenjuan Guo}, \bibinfo{person}{Bin Yang}, {and} \bibinfo{person}{Qingyong Li}.} \bibinfo{year}{2023}\natexlab{}.
\newblock \showarticletitle{CGF: A Category Guidance Based PM$_{2.5}$ Sequence Forecasting Training Framework}.
\newblock \bibinfo{journal}{\emph{IEEE Transactions on Knowledge and Data Engineering}} (\bibinfo{year}{2023}).
\newblock


\bibitem[\protect\citeauthoryear{Zeng, Chen, Zhang, and Xu}{Zeng et~al\mbox{.}}{2023}]%
        {zeng2023transformers}
\bibfield{author}{\bibinfo{person}{Ailing Zeng}, \bibinfo{person}{Muxi Chen}, \bibinfo{person}{Lei Zhang}, {and} \bibinfo{person}{Qiang Xu}.} \bibinfo{year}{2023}\natexlab{}.
\newblock \showarticletitle{Are transformers effective for time series forecasting?}. In \bibinfo{booktitle}{\emph{{AAAI}}}, Vol.~\bibinfo{volume}{37}. \bibinfo{pages}{11121--11128}.
\newblock


\bibitem[\protect\citeauthoryear{Zhang, Bian, Qu, Tuo, and Wang}{Zhang et~al\mbox{.}}{2021}]%
        {zhang2021time}
\bibfield{author}{\bibinfo{person}{Lingyu Zhang}, \bibinfo{person}{Wenjie Bian}, \bibinfo{person}{Wenyi Qu}, \bibinfo{person}{Liheng Tuo}, {and} \bibinfo{person}{Yunhai Wang}.} \bibinfo{year}{2021}\natexlab{}.
\newblock \showarticletitle{Time series forecast of sales volume based on XGBoost}. In \bibinfo{booktitle}{\emph{Journal of Physics: Conference Series}}, Vol.~\bibinfo{volume}{1873}. \bibinfo{pages}{012067}.
\newblock


\bibitem[\protect\citeauthoryear{Zhang, Guo, Dong, He, Xu, and Chen}{Zhang et~al\mbox{.}}{2017}]%
        {zhang2017cautionary}
\bibfield{author}{\bibinfo{person}{Shuyi Zhang}, \bibinfo{person}{Bin Guo}, \bibinfo{person}{Anlan Dong}, \bibinfo{person}{Jing He}, \bibinfo{person}{Ziping Xu}, {and} \bibinfo{person}{Song~Xi Chen}.} \bibinfo{year}{2017}\natexlab{}.
\newblock \showarticletitle{Cautionary tales on air-quality improvement in Beijing}.
\newblock \bibinfo{journal}{\emph{Proceedings of the Royal Society A: Mathematical, Physical and Engineering Sciences}} \bibinfo{volume}{473}, \bibinfo{number}{2205} (\bibinfo{year}{2017}), \bibinfo{pages}{20170457}.
\newblock


\bibitem[\protect\citeauthoryear{Zhang and Yan}{Zhang and Yan}{2022}]%
        {zhang2022crossformer}
\bibfield{author}{\bibinfo{person}{Yunhao Zhang} {and} \bibinfo{person}{Junchi Yan}.} \bibinfo{year}{2022}\natexlab{}.
\newblock \showarticletitle{Crossformer: Transformer utilizing cross-dimension dependency for multivariate time series forecasting}. In \bibinfo{booktitle}{\emph{{ICLR}}}.
\newblock


\bibitem[\protect\citeauthoryear{Zhao, Guo, Cheng, Han, Zhang, and Yang}{Zhao et~al\mbox{.}}{2023}]%
        {zhao2023multiple}
\bibfield{author}{\bibinfo{person}{Kai Zhao}, \bibinfo{person}{Chenjuan Guo}, \bibinfo{person}{Yunyao Cheng}, \bibinfo{person}{Peng Han}, \bibinfo{person}{Miao Zhang}, {and} \bibinfo{person}{Bin Yang}.} \bibinfo{year}{2023}\natexlab{}.
\newblock \showarticletitle{Multiple time series forecasting with dynamic graph modeling}.
\newblock \bibinfo{journal}{\emph{Proc. {VLDB} Endow.}} \bibinfo{volume}{17}, \bibinfo{number}{4} (\bibinfo{year}{2023}), \bibinfo{pages}{753--765}.
\newblock


\bibitem[\protect\citeauthoryear{Zhao, Chen, Deng, Kieu, Guo, Yang, Zheng, and Jensen}{Zhao et~al\mbox{.}}{2022}]%
        {zhao2022outlier}
\bibfield{author}{\bibinfo{person}{Yan Zhao}, \bibinfo{person}{Xuanhao Chen}, \bibinfo{person}{Liwei Deng}, \bibinfo{person}{Tung Kieu}, \bibinfo{person}{Chenjuan Guo}, \bibinfo{person}{Bin Yang}, \bibinfo{person}{Kai Zheng}, {and} \bibinfo{person}{Christian~S Jensen}.} \bibinfo{year}{2022}\natexlab{}.
\newblock \showarticletitle{Outlier detection for streaming task assignment in crowdsourcing}. In \bibinfo{booktitle}{\emph{{WWW}}}. \bibinfo{pages}{1933--1943}.
\newblock


\bibitem[\protect\citeauthoryear{Zhou, Zhang, Peng, Zhang, Li, Xiong, and Zhang}{Zhou et~al\mbox{.}}{2021}]%
        {zhou2021informer}
\bibfield{author}{\bibinfo{person}{Haoyi Zhou}, \bibinfo{person}{Shanghang Zhang}, \bibinfo{person}{Jieqi Peng}, \bibinfo{person}{Shuai Zhang}, \bibinfo{person}{Jianxin Li}, \bibinfo{person}{Hui Xiong}, {and} \bibinfo{person}{Wancai Zhang}.} \bibinfo{year}{2021}\natexlab{}.
\newblock \showarticletitle{Informer: Beyond efficient transformer for long sequence time-series forecasting}. In \bibinfo{booktitle}{\emph{{AAAI}}}, Vol.~\bibinfo{volume}{35}. \bibinfo{pages}{11106--11115}.
\newblock


\bibitem[\protect\citeauthoryear{Zhou, Ma, Wen, Sun, Yao, Yin, Jin, et~al\mbox{.}}{Zhou et~al\mbox{.}}{2022a}]%
        {zhou2022FiLM}
\bibfield{author}{\bibinfo{person}{Tian Zhou}, \bibinfo{person}{Ziqing Ma}, \bibinfo{person}{Qingsong Wen}, \bibinfo{person}{Liang Sun}, \bibinfo{person}{Tao Yao}, \bibinfo{person}{Wotao Yin}, \bibinfo{person}{Rong Jin}, {et~al\mbox{.}}} \bibinfo{year}{2022}\natexlab{a}.
\newblock \showarticletitle{Film: Frequency improved legendre memory model for long-term time series forecasting}.
\newblock \bibinfo{journal}{\emph{Advances in Neural Information Processing Systems}}  \bibinfo{volume}{35} (\bibinfo{year}{2022}), \bibinfo{pages}{12677--12690}.
\newblock


\bibitem[\protect\citeauthoryear{Zhou, Ma, Wen, Wang, Sun, and Jin}{Zhou et~al\mbox{.}}{2022b}]%
        {zhou2022fedformer}
\bibfield{author}{\bibinfo{person}{Tian Zhou}, \bibinfo{person}{Ziqing Ma}, \bibinfo{person}{Qingsong Wen}, \bibinfo{person}{Xue Wang}, \bibinfo{person}{Liang Sun}, {and} \bibinfo{person}{Rong Jin}.} \bibinfo{year}{2022}\natexlab{b}.
\newblock \showarticletitle{Fedformer: Frequency enhanced decomposed transformer for long-term series forecasting}. In \bibinfo{booktitle}{\emph{{ICML}}}. \bibinfo{pages}{27268--27286}.
\newblock


\end{thebibliography}
\end{document}